博士論文

**Analysis and Modeling of Driver Behavior with Integrated Feedback of Visual and Haptic Information Under Shared Control**
(シェアードコントロールを行っている際の視覚と力覚情報を統合してフィードバックするドライバの行動の解析とモデリング)

王 正

# Contents

# Acknowledgements

It is a great pleasure to express here my great thanks to all who made my PhD work possible. I sincerely thank my supervisor, Dr. Kimihiko Nakano, Associate Professor at The University of Tokyo, for his patient guidance, continuous support and enduring encouragement. He always expands my knowledge and gives professional advice on my research work through the four years. His enthusiasm to cutting-edge research and attention to details will continue to influence my research career in the future.

I greatly thank Professor Shigehiko Kaneko, Professor Yoshihiro Suda, Dr. Yuji Yamakawa, and Dr. Takuji Narumi. As referees of my dissertation defense committee, all of their valuable comments and constructive suggestions helped me to complement my research and greatly improve the dissertation. I learned a great deal from the inspiring discussions with all of them.

Coming from the field of Mechanical Engineering, the work in the field of Human-Machine Interaction in Japan presented a big challenge and a valuable experience for me, and I would like to thank all of you who supported me on my way. I would like to thank Dr. Rencheng Zheng, the project research associate of K. Nakano Laboratory. He supported me in my experiments and academic paper writings, and helped me so much not only in my research work but also in my daily life. I appreciate Dr. Tsutomu Kaizuka, the research associate and Dr. Keisuke Shimono, the project researcher of K. Nakano Laboratory, who gave me much advice about preparing experimental materials and using experimental devices. To our secretary Ms. Atsuko Hasegawa, thanks for all support in my research and daily life, and for helping me out with all the forms for business travel, monetary compensation, and so on.

I also would like to express my gratitude to Dr. Dongxu Su, Dr. Yunshun Zhang, Bo Yang, Antonin Joly, and all the other members of K. Nakano Laboratory. They gave me their help with their kindness in research discussions and daily life sharing. I hope we will work together and achieve wonderful outcomes in the future.

I am grateful to my beloved wife, Hefei Guan, as this whole work could not have been possible without her consistent support and deep love. She was my high school sweetheart thirteen years ago and she has been supporting me since then. She makes my life so delightful and so meaningful and she keeps inspiring me to be a better person. I would like to thank her for everything she did for me, including recently pregnancy with our baby.

Finally, I am deeply indebted to my parents for their unceasing encouragement and support.

<div style="text-align: right;">
July 30, 2018<br>
WANG Zheng
</div>

# Abstract

The development of mass-produced vehicles from last century allowed vehicle to be the ultimate consumer product due to affordability that gave great mobility. However, it makes the traffic accidents to be unavoidable. Although there have been many improvements in driving safety due to passive and active safety systems in recent decades, the number of traffic accidents is still very high.

To improve driving safety and comfort, driver-automation shared control, which combines the benefit of vehicular automation and human driver during a driving task, has been investigated. Shared control with a haptic guidance steering system allows both the human and the automation to exert torques on a steering wheel. Haptic guidance steering has been reported as a promising way to reduce the workload of drivers while keeping them in the control loop of the driving task, and meanwhile to improve the lane following performance. However, the haptic guidance system has its drawback. When facing a critical event, confusion may occurs to the driver on trusting the system or not. What's more, the haptic guidance system may induce conflicts and increased workload to the driver if the system is misused and the guidance torques cannot be relied on. Thus, it is reasonable to believe that supplementary haptic information should be provided to the driver on an as-needed basis.

However, there is not much agreement on how to design and how to evaluate a haptic guidance system on an as-needed basis, in order to improve both driving safety and comfort.

To solve this problem, we proposed that the need of haptic information provided by the haptic guidance system could be based on the reliability of visual information perceived by a driver, and the system could be evaluated by its performance compensation on visual perception. The aim of the thesis is to perform an analysis of the effect of haptic information provided by the guidance system on driver behavior in cases of normal and degraded visual information. Through analysis and modeling of driver behavior, we aim to provide theoretical and experimental knowledge of driver behavior when driving with a haptic guidance system for lane following tasks. Therefore, this thesis focuses on the design and evaluation of a haptic guidance system by analysis and modeling of driver behavior with integrated feedback of visual and haptic information.

In this thesis, three driving simulator experimental studies were presented. Experimental study I focused on testing driver behavior under different degrees of haptic guidance in the condition of normal visual information. The 15 participants drove five trials: no haptic guidance (Manual), haptic guidance with a normal feedback gain (HG-normal), haptic guidance with a strong feedback gain (HG-strong), and two kinds of automated driving. The driver behaviors, including steering behavior, lane following performance, gaze behavior and subjective evaluation, were analyzed. From the results, we found that the lane following performance increased or remained similar when driving with the proposed haptic guidance system, and meanwhile driver steering effort was reduced. However, the great reduction in driver steering effort caused by strong haptic guidance had a downside effect, which was the feeling of frustration. In addition, the implementation of haptic guidance system did not significantly influence the driver visual attention on the look-ahead point along curves, which indicates that the haptic guidance system was reliable and did not induce the increase of driver visual demand.

Experimental study II focused on testing driver behavior when driving with different degrees of haptic guidance in the condition of degraded visual information caused by visual occlusion from road ahead. The 12 participants drove twelve trials designed by combining three degrees of haptic guidance (HG): HG none (or manual driving), HG normal, and HG strong, with four scenarios of visual feedback (VF): VF whole, VF near, VF mid, and VF far. From the results, we found that the decrement of lane following performance caused by visual occlusion from road ahead could be compensated by the haptic information provided by the guidance system, and strong haptic guidance was more effective than normal haptic guidance.

Experimental study III focused on testing driver behavior influenced by the haptic guidance system in the condition of degraded visual information caused by declined visual attention under fatigue driving. The 12 participants drove two trials: a treatment session that implemented the haptic guidance steering system after a prolonged driving on monotonous roads, and a control session without the system. From the results, we found that the decrement of lane following performance on straight lanes caused by declined visual attention under fatigue driving could be compensated by haptic information provided by the guidance system. On the other hand, the active torque on the steering wheel stimulated the fatigued driver, and according to the result of visual behavior, the driver visual attention was increased.

To address the objective evaluation of the effect of haptic guidance on driver behavior, the development of a parameterized driver model with consideration of both visual and haptic information is required. Moreover, in order to understand the driver behavior obtained in the experiments, and furthermore, to predict driver behavior, we performed the identification and validation of the proposed driver model, and conducted numerical analysis of driver behavior influenced by degraded visual information and the haptic guidance system. The results showed that the lane following performance was decreased in the conditions of low visibility and declined visual attention caused by fatigue, and the haptic guidance system was effective in improving the lane following performance. It suggests the potential of using the proposed driver model and numerical simulations for designing and evaluating the haptic guidance system.

To sum up, the thesis presents contributions made to the evaluation and design of a haptic guidance system on improving driving performance in cases of normal and degraded visual information, which are based on behavior experiments, modeling and numerical simulations. The effect of shared control on driver behavior in cases of normal and degraded visual information has been successfully evaluated experimentally and numerically. The evaluation results indicate that the proposed haptic guidance system is capable of providing reliable haptic information, and is effective on improving lane following performance in the conditions of visual occlusion from road ahead and declined visual attention under fatigue driving. Moreover, the appropriate degree of haptic guidance is highly related to the reliability of visual information perceived by the driver, which suggests that designing the haptic guidance system based on the reliability of visual information would allow for greater driver acceptance. Furthermore, the parameterized driver model, which considers the integrated feedback of visual and haptic information, is capable of predicting driver behavior under shared control, and has the potential of being used for designing and evaluating the haptic guidance system.

# Nomenclature

| | |
|---|---|
| $a_1$ | Constant gain for lateral error at the near point in driver model |
| $a_2$ | Constant gain for integral of lateral error at the near point in driver model |
| $a_3$ | Constant gain for derivative of lateral error at the near point in driver model |
| $a_4$ | Constant gain for yaw error at the far point in driver model |
| $a'_1$ | Constant gain for lateral error at the near point in haptic model |
| $a'_2$ | Constant gain for derivative of lateral error at the near point in haptic model |
| $a'_3$ | Constant gain for yaw error at the far point in haptic model |
| $a'_4$ | Constant gain for derivative of yaw error at the far point in haptic model |
| $B_s$ | Steering wheel system damping |
| $E_t$ | The sum of pneumatic trail and castor trail |
| $e_y$ | Lateral error at the near point in driver model |
| $e_\theta$ | Yaw error at the far point in driver model |
| $e'_y$ | Lateral error at the near point in haptic model |
| $e'_\theta$ | Yaw error at the far point in haptic model |
| $I$ | Vehicle yaw moment inertia |
| $J_s$ | Steering wheel system inertia |
| $K_{aln}$ | Aligning coefficient |
| $K_d$ | Target steering angle to torque gain |
| $K_f$ | Cornering stiffness of front tire |
| $K_{hf}$ | Neuromuscular reaction gain for haptic feedback |
| $K_{nms}$ | Neuromuscular reflex gain |
| $K_r$ | Cornering stiffness of rear tire |
| $K_s$ | Spring constant converted to the kingpin |
| $K_t$ | Transmission ratio between steering wheel and tire angle |
| $K_1$ | Constant gain for haptic guidance torque |
| $l_f$ | Longitudinal position of front wheels from vehicle center of gravity |
| $l_r$ | Longitudinal position of rear wheels from vehicle center of gravity |
| $m$ | Vehicle mass |
| $r$ | Yaw rate |
| $T_a$ | Aligning torque |
| $T_d$ | Driver input torque |
| $T_h$ | Haptic guidance torque |
| $t_{nms}$ | Neuromuscular time constant |
| $t_p$ | Processing time delay |
| $t_n$ | Look ahead time at near point in driver model |
| $t_f$ | Look ahead time at far point in driver model |
| $t'_n$ | Look ahead time at near point in haptic model |
| $t'_f$ | Look ahead time at far point in haptic model |
| $v$ | Vehicle speed |
| $x_1$ | 1st state variable |
| $x_2$ | 2nd state variable |
| $x_3$ | 3rd state variable |
| $\beta$ | Side slip angle |
| $\varphi$ | Steering wheel angle |
| $\varphi'$ | Target steering wheel angle by driver |
| $\delta$ | Front wheel steer angle |

# Chapter 1

# Introduction

# 1 Introduction

Traffic accidents have been a main cause of human death over the last couple of decades. In order to reduce the number of traffic accidents, it is necessary to make a continuous progress on designing driver assistance systems. A recent direction is designing assistance systems that are capable of actively and intelligently assisting the drivers.

This thesis presents contributions made to the design and evaluation of a haptic guidance system on improving driving performance for lane following tasks in cases of normal and degraded visual information, which are based on behavior experiments, modeling and numerical simulations.

This Chapter firstly introduces the background and motivation of this work, and then the related previous research is explained. After that, the current problem of designing and evaluating the haptic guidance system is stated, and it is followed by the research challenges to solve the problem and the objectives of this work. The last part is the overview of the thesis.

## 1.1 Background

The development of mass-produced vehicles from last century allowed vehicle to be the ultimate consumer product due to affordability that gave great mobility, but it also makes the traffic accidents to be unavoidable. Although there have been many improvements in driving safety due to safety systems, the number of traffic accidents is still remarkable.

Over the past years, a plenty of safety systems have been invented by auto companies. Generally, the safety systems can be classified into two categories: passive safety systems and active safety systems. Passive safety systems help to reduce the negative influence of an accident as they activate during or after the accident takes place. For example, crumple zones, airbags and seat-belts have been widely made by manufactures in order to increase passive safety. In contrast, active safety systems include a number of safety functions which reduce the chances of an accident in the first place. Thus, the manufactures always employ active safety systems in order to avoid accidents. For example, anti-lock braking systems, head up displays, and electronic stability control have been widely employed to increase active safety.

In the last decades, due to the continuous progress in the development of automated driving, more and more research has focused on Advanced Driver Assistance Systems (ADAS). They have been designed to improve both driving safety and comfort. Among these systems, adaptive cruise control (ACC) system [1] and lane keeping assistance (LKA) system [2] have been well developed to enhance the longitudinal and lateral driving performance.

Recently, to further improve driving safety and comfort, driver-automation shared control, which combines the benefit of vehicular automation and human driver during a driving task, has been investigated [3]. Haptic guidance steering control is one kind of driver-automation shared control [4, 5], and a haptic guidance steering system enables the driver and the automation system simultaneously control the steering wheel through a haptic interface, as shown in Figure 1.1. In such case, the driver is influenced not only by the visual information from road ahead, but also by the haptic information from the steering wheel. Moreover, the authority of driver is always higher than that of the haptic guidance system, which means the driver can always overrule the steering wheel when there is a conflict.

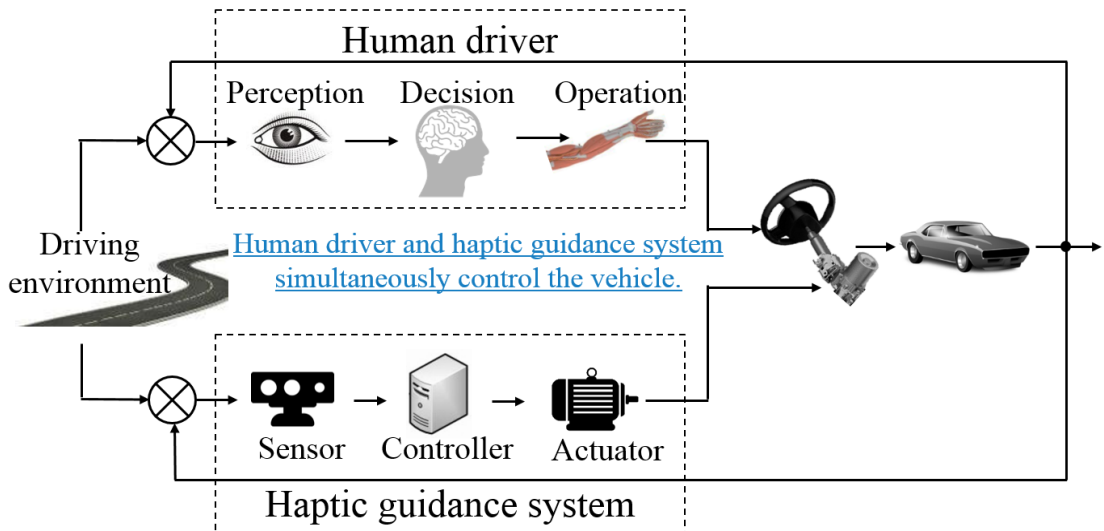

Figure 1.1 Diagram of one kind of driver-automation shared control through a haptic interface on a steering wheel.

## 1.2 Motivation

Better understanding of the driver behavior based on measurements and modeling is expected to lead to better design of the assistance system, which contributes to the improvement of driving safety and comfort. In the current state, the understanding of driver behavior under driver-automation shared control through a haptic interface is still scant.

## 1.3 Previous research

### 1.3.1 Human-automation interaction

Interaction is a kind of action that occurs as two or more objects have an effect upon one another. Human-automation interaction is the circumstance in which human: (a) specify to the automation, the task goals and constrains, and trade-offs between them; (b) control the automation to start or stop or modify the automatic task execution; and (c) receive from the automation information, energy, physical objects, or substances [6]. Examples of human-automation interaction could be people adjusting the controls on their cameras and setting the safe distance on forward collision warning system in their automobiles.

Generally, the designing of human-automation interaction need to follow two-parts of guideline. On one hand, the operator should have a widely understanding of the automation system. On the other hand, there should be knowledge of operator embedded in the automation system.

Accordingly, in order to decrease the disagreement between an automation system and human operator, it is necessary to figure out the human behavior according to the measurements and modeling [7].

In order to satisfy the design guidelines for human-automation interaction, normally the appropriate level of automation [8] needs to be determined. The level of automation might be designed to be constant. This is because it has been shown that when the level of automation keeps the same, less automation failures are detected in previous studies. Alternatively, the automation system is capable of deciding when to change its level of automation according to the needs.

Complex human-machine systems are an increasing presence in many aspects of our daily life, which makes it important to develop methods on shared control between an automation machine and a human operator.

### 1.3.2 Shared control

Shared control is broadly defined as the situation where human and machine carry out a task simultaneously. Human operators interact with automated entities such as robots and artificial intelligence to work together to achieve a common goal. Shared control is distinct from manual control and full automation. The idea behind shared control is to keep the human operator in the direct manual control loop while providing the continuous support of an automation system[7].

Shared control has been mainly distinguished into two forms: extension and relief [9]. In the form of extension, the application of shared control with machine could lead to a better performance than the best human could do. The form of relief refers to that machines help human operator to decrease the workload by analyzing the information and situation. For example, the system can detect the distance between the preceding vehicle and the ego vehicle, and helps the operator to keep a safe distance with the preceding vehicle by offering a force feedback on the gas pedal [10, 11]. In addition, decision supports and other types of mental activity can also be accomplished by shared control [12].

According to the utility of shared control system, the current study mainly focuses on three areas: automotive, robotic surgery assistance, and brain-machine interfaces [13]. In terms of automotive domain, a shared control system has been employed on a gas pedal to exert force feedback to drivers in order to keep a safe distance with the preceding vehicle [10]. In the haptic shared control for lateral tasks, a haptic steering wheel has been applied to help keeping the vehicle within the lane on straight roads and along curves [14]. Shared control system has also been applied in different driving maneuvers, such as evasive maneuvers [15], lane changing maneuvers[16], and merging or cut-in maneuvers [17]. In terms of robot-assisted surgery domain, shared control has its benefit, because it is barely possible to perform a surgery without the assistance of machine due to the high risk. In the scientific experiments as well as clinic applications, the robotic support systems in surgery have been developed with different automation levels of shared control [18]. The robotic assistance system used in tele-operated cardiac surgery enhances the stability of operation without communication with the operator, and this system is considered as a type of 'input-mixing' shared control [7]. In terms of brain-machine interface (BMI) domain, the applications of shared control aim to support patients who suffer from motor impairments. The patients can accomplish some of the daily activities by controlling assistant robotic devices through thoughts via BMIs [19, 20]. Furthermore, the shared control is expected to be a promising way to compensate the uncertainties related to BMIs. By combining the BMI with a shared control system, the performance of operating a device has been improved in terms of speed and safety [21, 22].

Shared control is currently still a quite novel definition; however, it is becoming popular to improve the communication between human beings and intelligent machines.

### 1.3.3 Haptic interface

It has been found that, to achieve a better performance of human-automation interaction, there are several requirements, including: (a) human remains in control all the time and can shift between different automation levels smoothly. (b) human receives continuous feedback from automation. (c) human always has interaction with the automation. (d) performance can be increased and human workload can be reduced [7]. The above four requirements can be achieved by a shared control between the operator and automation through haptic interface. Shared control through haptic interface (or haptic shared control) allows both the human and the automation to exert forces on a control interface, of which its output remains the direct input to the controlled system [23]. A famous haptics interface would be the direct interaction regarding the physical control of the human-horse system [3]. A horseback rider controls the forward, backward, sideways and rotational movement of the horse with a combination of continuous and discrete inputs of the hands on the reins, pressure with the legs, seat movement, and weight shift.

Haptic shared control has been applied in the areas of vehicle control [4, 24-27], UAV control [28-30], robotic surgery [31-33] and remote operation [34, 35]. In terms of haptic shared control between driver and automation, the steering wheel is both grasped by the driver and motorized by the automation system. The motion of the steering wheel is then a response to the sum of forces acting from the human grasp and from the automatic control motor. Negotiation of authority is also possible: upon sensing the intentions of the automatic controller through his grip on the steering wheel, the driver can judge them either reasonable or inappropriate, and can choose to yield to those intentions or to override them [24]. Moreover, to ensure that the actions of an automatic controller are accepted by the driver, it is advantageous to base it on a driver model [36].

### 1.3.4 Haptic guidance steering

Haptic guidance steering is one kind of haptic shared control between driver and automation. As for haptic guidance steering, the authority of the driver is always higher than that of the automation system, which means the driver can always overrule the steering wheel when there is a conflict. Thus, a haptic guidance steering system is designed to guide the driver to steer on the target trajectory with assistant torques on the steering wheel.

When driving with a haptic guidance system, ideally, the drive can comfortably rely on the force feedback and conduct a driving with more safety. It has been found that haptic guidance could assist drivers to perform appropriate actions for curve negotiation by producing both the recommended direction and magnitude of the suitable steering operation [14]. Haptic guidance systems have also been developed to continuously and smoothly support lane-changing maneuvers, while maintaining the benefits of the lane-keeping system [16]. For emergency obstacle avoidance, an active steering system is designed to assist drivers to maneuver the steering wheel promptly and steadily [37]. Haptic guidance steering also leads to reduced workload as demonstrated by faster reaction times and better secondary task performance [25].

Regarding the automation authority, it is crucial to determine an optimal degree of haptic guidance to mitigate the conflicts and achieve high quality of cooperation [14, 27]. With higher degree of haptic guidance, the driver can achieve lane following with less physical effort [4]; however, when the haptic guidance does not match the individual driver's steering intention, more intense conflicts occur between the driver's input torque and the haptic guidance torque during a lane following task [38].

Some other negative results of haptic guidance steering have been presented, including an increased number of collisions because the system does not guide the driver in the case of encountering an obstacle within the lane on the road [39]. In such case, it demands an evasive maneuver by the driver, and the driver will have to counteract the haptic guidance torque if the system still keeps the vehicle centered within the lane [25]. Moreover, it is also possible that the haptic guidance system provides inaccurate information in other certain circumstances. In terms of a curve negotiation, the system might misestimate the radius of a curve, resulting in inappropriate steering torques [39]. In the above cases, the driver even has to spend more workload compared to manual driving or fully automated driving.

**1.4 Problem statement**

Apparently, the haptic guidance system has its drawback. When facing a critical event, confusion may occurs to the driver on trusting the system or not. What's more, the haptic guidance system may induce conflicts and increased workload to the driver if the system is misused and the guidance torque cannot be relied on. Thus, it is reasonable to believe that supplementary haptic information should be provided by the haptic guidance system on an as-needed basis.

However, there is not much agreement on how to design and how to evaluate a haptic guidance system on an as-needed basis in order to improve both driving safety and comfort.

To solve this problem, we proposed that the need of haptic information provided by the haptic guidance system could be based on the reliability of visual information perceived by a driver, and the system could be evaluated by its performance compensation on visual perception. The goal of the thesis is, therefore, to perform an analysis of the effect of haptic information provided by the guidance system on driver behavior in cases of normal and degraded visual information. To achieve this goal, the thesis focuses on the design and evaluation of a haptic guidance system by analysis and modeling of driver behavior with integrated feedback of visual and haptic information.

**1.5 Research challenges**

- Haptic guidance system design challenges

The design of a haptic guidance system includes several challenges. First, the variables that are used to calculate the direction and amplitude of the haptic guidance torque need to be determined, and a wrong mapping would lead to a mismatch between driver intention and haptic gudiance provided by the system. In addition, the amplitude of an appropriate haptic guidance torque needs to be determined. This is because a relatively low amplitude of guidance torque can hardly be perceived by the driver, and a relatively high amplitude of guidance torque is likely to cause intrusive feelings. Moreover, choices of sensors and actuators are needed for the system prototype design, and as for our prototype, this has been done by JTEKT Corporation, Japan.

- Haptic guidance system evaluation challenges

The designed haptic guidance system needs to be evaluated by experiments and numerical analysis in order to qualify its effect on driver behavior. One of the evaluation challenges is to determine the proper indexes for evaluating driver lane following performance and driver workload. Another evaluation challenge is to design a driving simulator experiment that allows for data recording and analysis, and meanwhile, the driver behavior in real-world could be mimicked. Moreover, trade-offs between subjective and objective evaluations on the haptic guidance system need to be addressed.

- Fundamental analysis challenges

Understanding of driver behavior based on measurements and modeling is crucial to the design and evaluation of a haptic guidance system. However, in the current state, the study on driver behavior based on integrated feedback of visual and haptic information for lane following tasks is scant. To this end, the analysis on the effect of haptic guidance system on driver behavior is performed, and the analysis mainly consists of theory, experiments and modeling.

**1.6 Objective**

The aim of this thesis is to perform an analysis of the effect of haptic information provided by the guidance system on driver behavior in cases of normal and degraded visual information. The analysis is aimed to provide theoretical, experimental and modeling knowledge of driver behavior when driving with a haptic guidance system for lane following tasks. This knowledge should also be helpful in possible further improvements of the haptic guidance system design.

The main objectives of this thesis are as follows:
(a) To design a haptic guidance system being capable of providing reliable haptic information to the driver in lane following tasks.

(b) To evaluate the effect of the haptic guidance system on driver behavior in cases of normal and degraded visual information by experimental studies.

(c) To model driver behavior with integrated feedback of visual and haptic information, and to evaluate the effect of the haptic guidance system on driver behavior by numerical simulations.

**1.7 Overview**

The schematic overview of the thesis is shown in Figure 1.2.

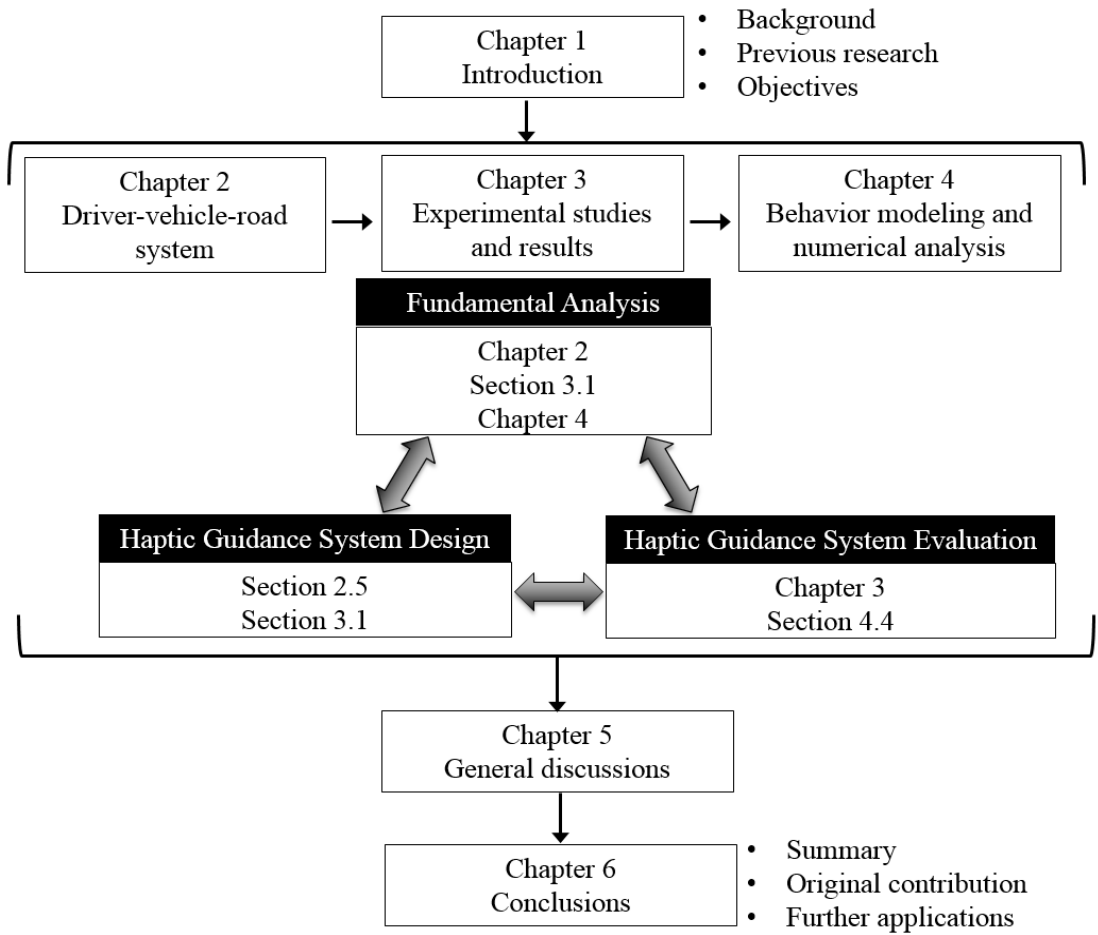

Figure 1.2 Schematic overview of the thesis. The thesis mainly consists of three research areas: fundamental analysis, haptic guidance system design and haptic guidance system evaluation. New knowledge achieved from one of the research areas will impact the other areas. Related chapters and sections are also shown in the research areas.

The thesis consists of 6 chapters:

In Chapter 1, an overview is executed about the developing techniques to improve safety and comfort in driving tasks. Haptic guidance steering control is comprehensively explained covering the introduction of human-automation interaction, shared control, and haptic interface. After that, the current problem of designing and evaluating haptic guidance system is stated, and it is followed by research challenges to solve the problem and the objective of the thesis.

In Chapter 2, the fundamental theory is described which helps in the design of the haptic guidance system, and also in the understanding of driver steering behavior when driving with the haptic guidance system. The general structure of a driver-vehicle-road system is firstly presented, and it is followed by the description of each subsystem: driver visual system, driver neuromuscular system, haptic guidance system, steering column system, and vehicle dynamics system.

Chapter 3 includes three experimental studies on evaluation of the haptic guidance system in cases of normal and degraded visual information. Experimental study I focuses on the effect of different degrees of haptic guidance on driver gaze behavior, lane following performance and driver workload. Experimental study II focuses on the effect of the haptic guidance system on driver behavior when driving with degraded visual information caused by visual occlusions from road ahead. Experimental study III focuses on the effect of the haptic guidance system on driver behavior when driving with degraded visual information caused by declined visual attention under fatigue.

In Chapter 4, firstly, the driver model parameters are identified based on the measured data from experimental study I. Secondly, model validation test is performed. After that, numerical analysis of driver behavior with normal and degraded visual information is conducted, and the effect of the haptic guidance system on the lane following performance is investigated based on integrated feedback of visual and haptic information.

In Chapter 5, general discussions about the observed results from three experimental studies and the numerical analysis are provided. The discussions are divided into two parts. The first part addresses the effect of degraded visual information on driver behavior, and the conditions of degraded visual information include visual occlusion from road ahead and declined visual attention under fatigue driving. The second part addresses the effect of the haptic guidance system on driver behavior in cases of degraded visual information.

In Chapter 6, the conclusions are drawn mainly about contributions and significance of the thesis work. The outlook of further applications based on the current findings is also addressed.

# Chapter 2

# Driver-Vehicle-Road System

# 2 Driver-Vehicle-Road System

## 2.1 Introduction

The driver-vehicle-road system is a broad research topic considering the complicated driving situations and numerous driver and environmental factors in a driving task. In this thesis, the modeling and analysis of a driver-vehicle-road system have been focused on a steering task for lane following purpose.

Figure 2.1 shows a general structure of a driver-vehicle-road system in the case of driving with a haptic guidance system. The driver-vehicle-road system consists of five subsystems, namely: driver, haptic guidance system, steering system, vehicle, and road path. The interconnection between the subsystems are also illustrated, and the variables in the driver-vehicle-road model are shown in Table 2.1. $T_h$ is the haptic guidance torque provided by the haptic guidance system, and a manual driving condition corresponds to $T_h = 0$. It is important to address the driver behavior based on integrated feedback of visual and haptic information in this chapter.

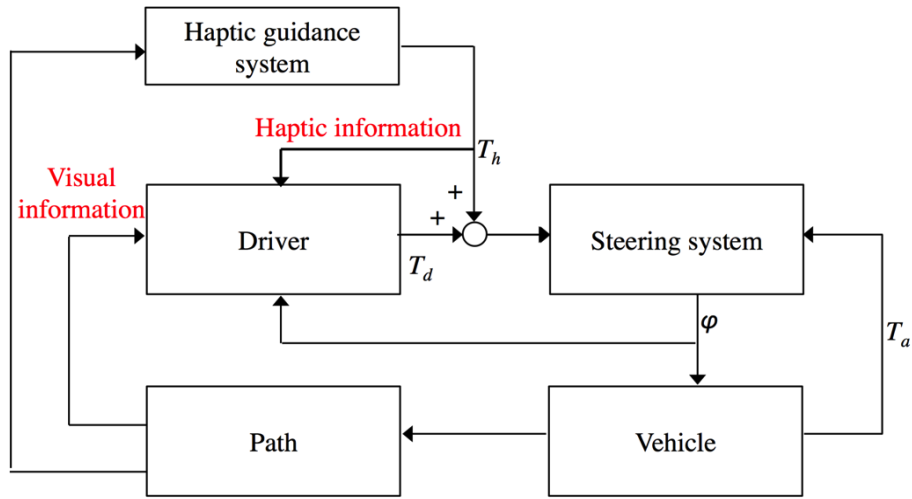

Figure 2.1 General structure of driver-vehicle-road system.

Table 2.1 System variables of driver-vehicle-road system.

|  | Definition |
| --- | --- |
| $\varphi$ | Steering wheel angle |
| $T_a$ | Aligning torque |
| $T_h$ | Haptic guidance torque |
| $T_d$ | Driver input torque |

Modeling is an important method within the development of vehicles to gain qualification criteria for the interaction of driver, vehicle, and road. Though there are numerous model approaches that describe the various aspects of vehicle behavior with high accuracy, the modeling of driver behavior is not far developed, especially in the case of driving with a haptic guidance system. In such case, the driver is expected to integrate the feedback of visual and haptic information to perform an appropriate maneuver in a lane following task. This chapter firstly introduces the background of human motion control which is necessary for understanding driver control behavior, and then the details for the subsystems of driver-vehicle-road system are discussed in the following parts.

## 2.2 Human motion control

When human beings move around in the environment, they are able to effectively interact with the environment and make responses according to different environments. The control system for the movement of human beings is quite complex, and is able to change in order to fit variable conditions. The human motion control system can be in accordance with a robot or an automation system: a linkage (skeleton of human), actuators (muscles of human), sensors (sensory receptions of human) and a controller (central nervous system of human) that is connected to the actuators and sensors by wires (nerves of human).

### 2.2.1 Motion control systems

The central nervous system can be divided into two parts: brain and spinal cords. It receives different types of sensory information including vision, haptics, and so on, and integrates them together. After integration, the central nervous system makes the decision and sends command to the muscles.

The muscles connected to the skeletons through the tendon generate and exert forces. In the central nervous system, there is a group of neurons called motor neuron. The motor neurons collect neural information from other areas of brain and spinal cords, integrate them and send commands to the muscle fibers. The muscle fibers and the controlling motor neuron are the main components of a motor unit, and thousands of motors units form into muscle. A motor neuron controls the contraction and relaxation of a muscle through muscle fibers. As a result, muscles generate and exert force in our daily life and they are controlled by brain.

Human sensory reception means by which humans react to changes in external and internal environments. Ancient philosophers called the human senses "the windows of the soul," and Aristotle described at least five senses: sight, hearing, smell, taste, and touch. The sensory information of this study comes from visual information from road ahead and haptic information from the guidance system.

### 2.2.2 Cybernetic approach

Cybernetics is defined as the scientific study of control and communication in the animal and the machine by Norbert Wiener [40]. It is an approach by which people explore regulatory systems, including their structures, constrains and possibilities. Nowadays, the cybernetics also implies that people use technology for the control of any systems. For example, the controlling and communications between human beings and machines could be a part of cybernetics.

The cybernetic approach has been developed to understand and mathematically model how humans control vehicles and devices [41-43]. The power of cybernetics is evident from the seminal crossover model [44], which captures the systematic adaptation of the human controller to the dynamics of the controlled vehicle, to achieve good feedback performance and robustness, which are largely invariant with the controlled system. By revealing such key invariants and providing a means for predicting manual control performance, classical cybernetics theory has accelerated many innovations in human–machine control system design [45-48].

In this thesis, a cybernetic driver model based on integrated feedback of visual and haptic information has been proposed. The driver model consists of two subsystems: a driver visual system which deals with the processing of the visual information from road ahead, and a driver neuromuscular system which deals with steering movement.

**2.3 Driver visual system**

The inputs to the driver visual system could be the error signals provided by the motion and the vision kinematics tracked by the driver. The output of the driver visual system could be the steering command to the driver neuromuscular system.

**2.3.1 Previous research on driver visual system**

With a visual system, a driver could sense the vehicle attitude, i.e., yaw angle, as well as the lateral displacement. The driver inside the vehicle is assumed to look forward to $L$ m ahead of the vehicle and estimate the deviation of the vehicle lateral displacement relative to the target course. The estimated lateral displacement is a lateral component of the vehicle traveling distance in a time of $L/v$ ($v$ is the driving speed) at the current attitude. The driver makes the feedback control based on this deviation, as shown in Figure 2.2.

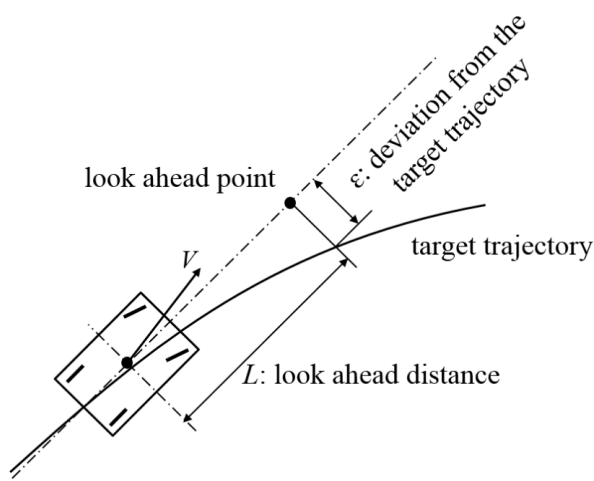

Figure 2.2 Course error at $L$ (m) ahead point, adapted from [49].

The human controller to follow the target trajectory could be studied by transfer function, and the proportional-integral-derivative (PID) controller is a commonly used one, which is given by:

$$H(s) = h(\tau_D s + 1 + \frac{1}{\tau_I s})e^{-\tau_L s} \qquad (2.1)$$

Firstly, a human operator will have a processing time delay to make an action under a given stimulus. This is represented by $e^{-\tau_L s}$, and the time delay is expressed by $\tau_L$. It has been found that the existence of $\tau_L$ is a basic cause for the vehicle motion to become unstable. Moreover, reduced attentiveness induces longer driver processing time delay [50]. In this thesis, a driving condition of declined visual attention under fatigue could be represented by a longer processing time delay in the driver model.

The control action that the human operator can do at ease and with the least workload is the kind of action that gives an output signal proportional to the input signal, in other words, proportional action. This is represented by the proportional constant, *h*. The human operator is also capable of the control action that predicts a change in the input. Here the output signal is normally proportional to the input rate or the differential value of the input. In other words, this is called the derivative control action. This is represented by the derivative time, $\tau_D$. The human operator is also capable of making integral action where the output signal is proportional to the integrated value of the input signal. This shows that the human operator is able to relocate point when it is deviated away from the original. This is represented by the integral time, $\tau_I$. Moreover, the human control action without much workload is considered to be the proportional action, with weak derivative action and integral action. Especially, the increase of derivative action could give much workload to the human operator [49].

In addition to the above-mentioned look-ahead model, a two-point visual model has been proposed for more stable steering. Generally, a driver steering a car on a twisting road has two distinct tasks: to match the road curvature, and to keep a proper distance from the lane edges. If the driver makes the control action by sensing the yaw angle, rather than the lateral displacement, it is expected that the driver will be able to control the vehicle more easily, even without the derivative action. The previous research agrees that successful steering requires the driver to monitor the angular deviation of the road from the vehicle's present heading at some 'preview' distance ahead, typically about 1 s into the future. The more recent study suggests that good driving performance requires that both a distant and a near region of the road are visible [51, 52]. As shown in Figure 2.3, the distant region, "segment a", is used to estimate road curvature, the near region, "segment c", is used to provide position-in-lane feedback, and the mid-region, "segment b", is between the near region and the distant region. This thesis addresses the analysis of driver behavior in the case of degraded visual information caused by visual occlusion from road ahead , and the road ahead is segmented according to [51].

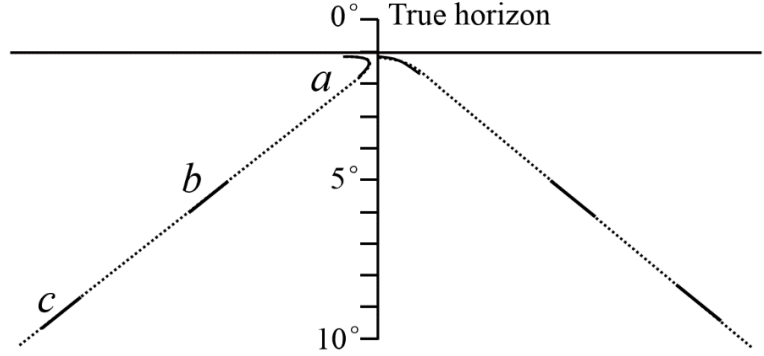

Figure 2.3 Visual feedback from road ahead segmented by a, b, and c, adapted from [51].

Regarding the two distinct tasks for curve negotiation, a driver steering behavior model that consists of two levels has been proposed: the guidance level involving the perception of the instantaneous and future course of the forcing function provided by the forward view of the road, and the response to it in an anticipatory open-loop control mode; and the stabilization level whereby any occurring deviations from the forcing function are compensated for in a closed-loop control mode [53]. After looking into the control algorithm, a two-point visual control model of steering has been proposed [54]. The two-point visual control model uses both a 'near point' to maintain a central lane position and a 'far point' to account for the upcoming roadway. Rather than a single variable $\theta$, a variable $\theta_n$ representing the change in visual direction to the near point and a variable $\theta_f$ representing the change in visual direction to the far point have been proposed, as follows:

$$\varphi = k_f \theta_f + k_n \theta_n + k_I \int \theta_n \, dt \qquad (2.2)$$

where $k_f$, $k_n$, $k_I$ are constant gains for $\theta_f$, $\theta_n$, $\int \theta_n \, dt$, respectively, and $\varphi$ represents the target steering wheel angle.

This two-point visual control model has been used to mimic driver visual behavior when designing driver assistance systems [46, 47, 55, 56]. The advantage of the two-point visual model is that the far point is used for maintaining stability and near point helps to keep the vehicle centered within the lane.

**2.3.2 Two-point driver visual model**

The two-point driver visual model used in this thesis is presented in Figure 2.4. The coupling between the vehicle model and the driver model is achieved via the road environment. The figure shows the geometric relationship between the vehicle, the target trajectory, and a two-point driver visual model. The centerline of the lane is set as the target trajectory. This driver visual model relies on both near and far points of road ahead: a near point to maintain a central lane position, and a far point to account for the upcoming road curvature. The lateral error, $e_y$, is defined as the distance between the vehicle and target trajectory at the near point. The yaw error, $e_\theta$, is defined as the angle between the movement direction of the vehicle and target trajectory at the far point. In this thesis, the default look-ahead time for the near point is 0.3 seconds, and the default look-ahead time for the far point is 1.0 second [51, 52]. A PID controller dealing with $e_y$ at the near point, and a proportion controller dealing with $e_\theta$ at the far point, have been proposed. The block diagram of the transfer function for the driver visual system is shown in Figure 2.5, and the variables and parameters in the visual system are shown in Table 2.2.

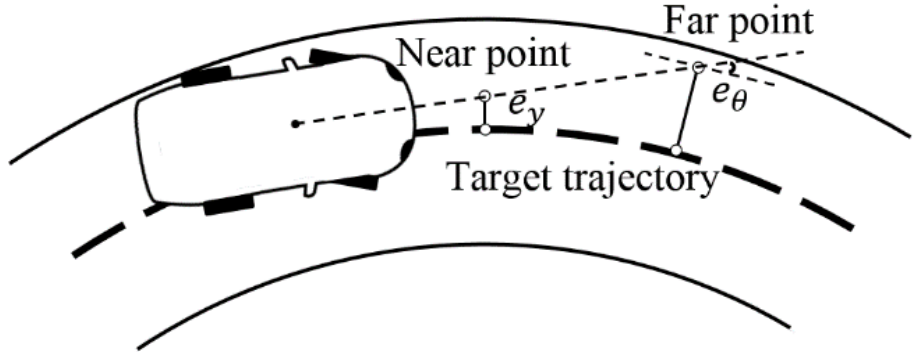

Figure 2.4 Schematic of two-point driver visual model.

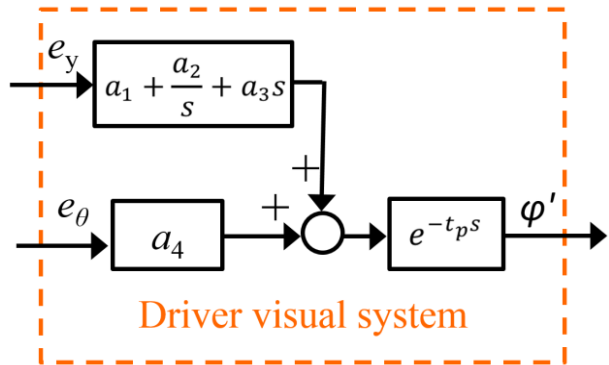

Figure 2.5 Block diagram of transfer function for driver visual system.

Table 2.2 Variables and parameters of driver visual system.

|  | Definition |
| --- | --- |
| $a_1$ | Constant gain for $e_y$ |
| $a_2$ | Constant gain for integral of $e_y$ |
| $a_3$ | Constant gain for derivative of $e_y$ |
| $a_4$ | Constant gain for $e_\theta$ |
| $t_f$ | Look ahead time at far point in driver model |
| $t_n$ | Look ahead time at near point in driver model |
| $e_y$ | Lateral error at the near point in driver model |
| $e_\theta$ | Yaw error at the far point in driver model |
| $t_p$ | Processing time delay |

In the proposed driver visual system, it is assumed that constant gain for derivative action $a_3$ equals to 0 when driving with normal visual information in a smooth driving task [54], as the increase of derivative action could give much workload to the human driver [49].

Furthermore, the effect of degraded visual information has been considered in the two-point driver visual model. As mentioned in Section 2.3.1, the declined visual attention under fatigue driving could be represented by a longer processing time delay. According to [47], a normal processing time delay should be between 0.01 s and 0.3 s; thus, a declined visual attention yields a processing time delay longer than 0.3 s. The effect of visual occlusion from the far segment of road ahead on driver behavior is also considered in the driver model. Low visibility, or only a near segment of road ahead being visible, is quite common in real life driving situation, as dense fog or raining would cause visual occlusion from far segment of the road ahead [57, 58]. In such case, it is assumed that far point is not available and the driver applies a PID control action to minimize the lateral error at the near point; moreover, it is assumed that the driver could increase the constant gain for derivative action $a_3$ to deal with the low visibility, especially when negotiating a curve.

**2.4 Driver Neuromuscular system**

The neuromuscular system includes all the muscles in the body and the nerves serving them. Every movement the body makes requires communication between the brain and the muscles. A suitable driver model needs to incorporate a neuromuscular system.

**2.4.1 Previous research on driver neuromuscular system**

The driver steering behavior is highly related to driver visual perception. The neuromuscular system generates the steering wheel angle demanded by the path-following controller with the eyes leading, as shown in Figure 2.6. The model demonstrates that reflex action improves the steering angle control and thus increase the path-following accuracy. The force producing fibers of the muscles are activated by $\alpha$-motoneuron in the spine. The $\alpha$-motoneuron can be signaled in two ways: directly from the brain and by feedback from spindles in the muscle. The spindles sense displacement and velocity of the muscle and feedback this information to the $\alpha$-motoneuron to maintain muscle length at a value commanded by the brain via the $\gamma$-motoneuron. Neuromuscular reflex continuously acts to minimize the difference between the actual steering wheel angle and the target steering wheel angle [59]. It has been found that the neuromuscular reflex can be simplified as a constant gain, which rejects any disturbance on the steering wheel that is caused by an external torque [47]. In addition, neuromuscular time constant needs to be considered for describing the time response of a driver's arms for a steering maneuver [46]. In this thesis, the neuromuscular system should also be able to represent driver feedback to the haptic information proved by the steering guidance system. In other words, the neuromuscular system should address the integrated feedback of visual and haptic information.

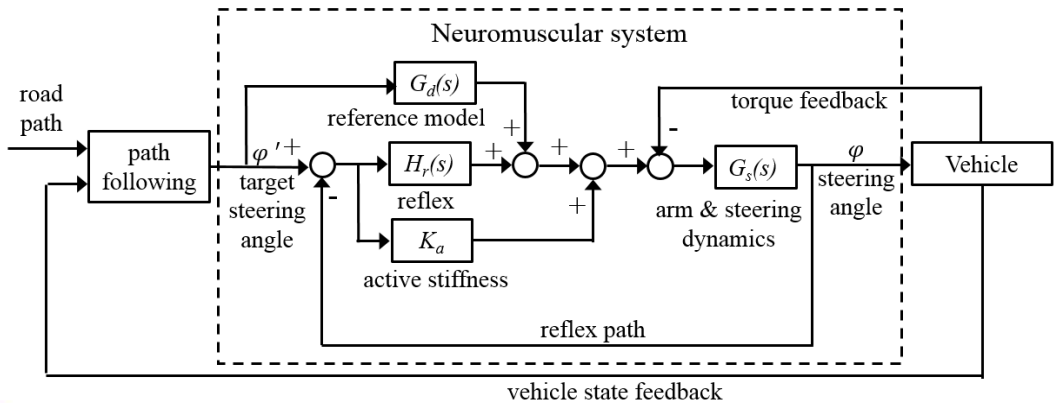

Figure 2.6 Linear driver steering model with neuromuscular dynamics, adapted from [59].

## 2.4.2 Previous research on human integration of multiple sensory information

Human beings receive and integrate multiple sensory information from the surroundings, in order to understand the world and interact with it. It has been shown in perceptual decision studies that human beings can receive information from single modality [60-62] as well as across different modalities [63, 64]. For example, when estimating the size [65] and shape [66] of an object, different modalities (visual and haptic cues) work together. In addition, the sensory information and motor information can also be integrated to control a task. For example, the location of hands and body postures can be controlled by the integration of multiple information [67-69]. In most cases, human beings make estimation by maximum likelihood or minimum variance when integrating cues in a near-optimal manner [70].

Human perceptual judgements rely upon visual information and haptic information in a weighted fashion, in which the more reliable information is given a higher weight [63]. In addition, the compliance of a material can be conveyed through mechanical interactions in a virtual environment and perceived through both visual and haptic cues. The model subsumes optimal fusion but provides valid predictions also if the weights are not optimal [71]. In terms of human postural control and task performing, human's sensorimotor system also tends to integrate multiple senses of visual and haptic [72-74]. Previous experimental results, where body sway is evoked by manipulation of individual and combined sensory cues, appear to be consistent with an essentially linear model [75, 76]. One possibility to combine the multiple senses is in an essentially linear manner by adding a 'sensory weighting factor' [68].

Inspired by that the human is shown to integrate multi-sensory information based on each reliability, it is hypothesized that the driver estimates from visual and haptic information that are linearly integrated with each kind of information estimate based on its reliability. In addition, it is hypothesized that, due to the integrated feedback of haptic and visual information, the lane following performance could be improved by haptic guidance when visual information is degraded.

## 2.4.3 Proposed driver neuromuscular model

As shown in Figure 2.7, the driver neuromuscular system considers driver's interaction with the haptic guidance system by using neuromuscular reaction gain for haptic feedback $K_{hf}$, and thus integrates the feedback of visual and haptic information via $K_d$ and $K_{hf}$. The parameters of driver neuromuscular system are shown in Table 2.3.

The target steering wheel angle, $\varphi'$, is converted into the driver input torque $T_d$ by means of the neuromuscular system. $K_d$ represents that the neuromuscular system provides a steering torque proportional to the target steering wheel angle. $K_{nms}$ represents the neuromuscular reflex that rejects any disturbance on the steering wheel that is caused by an external torque [47]. $K_{nms}$ continuously acts to minimize the difference between the actual steering wheel angle $\varphi$ and the target steering wheel angle $\varphi'$. $t_{nms}$ approximately describes the time response of a driver's arms for a steering maneuver [59].

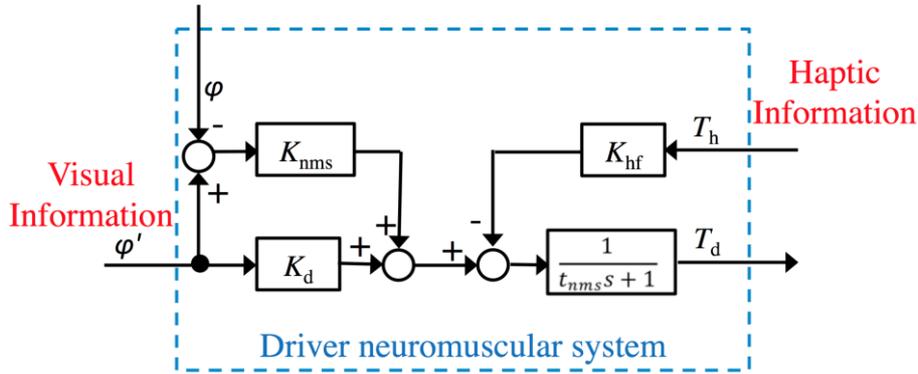

Figure 2.7 Block diagram of transfer function for driver neuromuscular system.

Table 2.3 System parameters of driver neuromuscular system.

|  | **Definition** |
|---|---|
| $K_d$ | Angle to torque gain |
| $K_{hf}$ | Neuromuscular reaction gain for haptic feedback |
| $K_{nms}$ | Neuromuscular reflex gain |
| $t_{nms}$ | Neuromuscular time constant |

It should be noticed that the neuromuscular reaction gain for haptic feedback, $K_{hf}$, is expected to describe driver interaction and reliance on the haptic guidance steering. When $K_{hf}$ equals to 1, it indicates that the driver does not rely on the haptic guidance steering. This is because the variation of $T_h$ is included in the driver input torque $T_d$, as shown in Figure 2.7, and it almost does not influence the sum torque of $T_h$ and $T_d$ into the steering system, as shown in Figure 2.1. Consequently, the driver steering performance will be barely influenced by $T_h$. As a comparison, when $K_{hf}$ equals to 0, it indicates that the driver fully relies on the haptic guidance steering. This is because the variation of $T_h$ is not included in the driver input torque $T_d$, as shown in Figure 2.7, and it remarkably influences the sum torque of $T_h$ and $T_d$ into the steering system, as shown in Figure 2.1. Considering this, the range of $K_{hf}$ from 0 to 1 is expected to represent the degree of the driver's reliance on haptic guidance steering. In addition, the value of $K_d$ is expected to be related to the value of $K_{hf}$.

**2.5 Haptic guidance steering system**

The haptic guidance system enables both the driver and haptic guidance to contribute to the steering input. The advantages of the haptic guidance control over manual driving have been demonstrated. However, the beneficial effects of haptic guidance control might be accompanied by downside effects. When the haptic guidance does not match the individual driver's steering intention, a conflict occurs between the driver's torque and haptic guidance torque. In order to mitigate the conflicts and achieve high quality of cooperation between the driver intention and the haptic guidance system, a driver model needs to be mimicked when designing the haptic guidance model.

**2.5.1 Previous research on haptic guidance model**

A predictive guidance method with a look-ahead interval has been used in the design of haptic guidance system. The method assumes that the lateral error between the vehicle and the target trajectory, and the yaw error can be measured. To minimize the lateral error and yaw error, proportional–derivative (PD) theory has been applied to design the controller [77]. In addition, according to [25], a haptic guidance system divides the steering task into two problems: generating a desired path that would return a stray vehicle to the road centerline and turning the steering wheel to follow that desired path. The path planning employs a geometric approach based on knowledge of the vehicle's position and orientation relative to the nearby road geometry. This approach follows the predictive driver model, using the notion of "aim point" ahead of the vehicle on the centerline of the road. This aim point is located by finding the closest point to the vehicle on the road and then looking forward 10 m along the road. Another predictive guidance method estimates the vehicle's position at the look ahead point by $P = vt$, where $v$ is the vehicle's current velocity. When the predicted point is outside of the path, a haptic guidance torque was applied based on the desired vehicle heading change [5]. The yaw error is the angle between the current vehicle heading and the line from the vehicle to the target point, $T$, the point on the path closest to $P$, as shown in Figure 2.8.

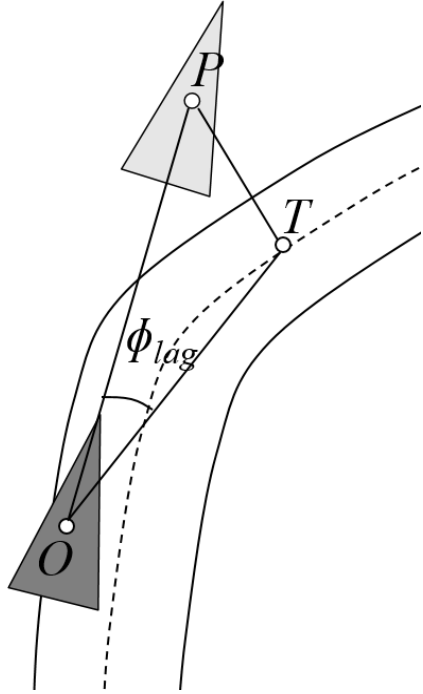

Figure 2.8 Components of the look-ahead guidance method, adapted from [5].

Apart from continuous haptic guidance torque, the bandwidth haptic guidance system has been designed to guide the drivers when the predicted lateral error exceeds a threshold [78]. The haptic guidance torque is also based on predicted lateral error to the lane center, but there will be no haptic guidance when the predicted lateral error is within the threshold.

The above methods for designing haptic guidance model do not quite mimic the driver behavior from the viewpoint of a two-point visual model. The advantage of a two-point visual model is that the far point is used for maintaining stability and near point helps to keep the vehicle centered within the lane. In this thesis, the proposed haptic guidance model is based on a two-point visual model.

**2.5.2 Proposed haptic guidance model**

The proposed haptic guidance model is based on the two-point visual model, as shown in Figure 2.4; only $e_y$ is replaced by $e'_y$, and $e_\theta$ is replaced by $e'_\theta$ in the haptic guidance model, which is for distinguishing the driver model and haptic guidance model. The target trajectory was previously designed as the centerline of the lane in the driving scenario. As the increase of derivative action could give much workload to the human driver, the haptic guidance system applies a PD control theory to take the derivative action in place of a human driver, in order to help keep the vehicle accurately following the target trajectory. The PD control theory is used to dealing with $e'_y$ at the near point as well as $e'_\theta$ at the far point. Accordingly, the haptic guidance torque, $T_h$, is expressed as

$$T_h = K_1(a'_1 e'_y + a'_2 \dot{e}'_y + a'_3 e'_\theta + a'_4 \dot{e}'_\theta) \tag{2.3}$$

where $a'_1$, $a'_2$, $a'_3$, and $a'_4$ are constant gains for $e'_y$, $\dot{e}'_y$, $e'_\theta$, and $\dot{e}'_\theta$, respectively; $K_1$ is a constant gain for the overall haptic guidance torque. The values of $a'_1$, $a'_2$, $a'_3$, and $a'_4$ are normally decided through a trial-and-error method. The method is generally divided into two steps. The first step is to find rough solution of automated driving with aggressive tuning of $a'_1$ and $a'_3$, when $K_1$ equals to 1; during the tuning, $a'_2$ is proportional to $a'_1$, and $a'_4$ is proportional to $a'_3$ [77]. The second step is to find desired solution with small tweaks in order to achieve good driving performance. In addition, the magnitude of haptic guidance torque is limited to 5 N·m by program, so that the drivers can choose to overrule the haptic guidance system at any time by exerting more torques on the steering wheel. The block diagram of transfer function for the haptic guidance system is shown in Figure 2.9, and the variables and parameters of the haptic guidance system are shown in Table 2.4.

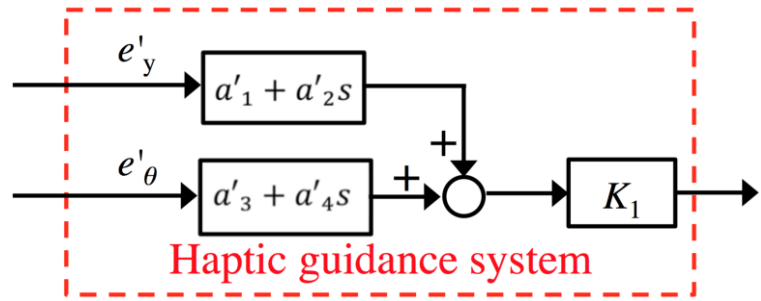

Figure 2.9 Block diagram of transfer function for haptic guidance system.

Table 2.4 Variables and parameters of haptic guidance system.

|  | **Definition** |
| --- | --- |
| $a'_1$ | Constant gain for $e'_y$ |
| $a'_2$ | Constant gain for derivative of $e'_y$ |
| $a'_3$ | Constant gain for $e'_\theta$ |
| $a'_4$ | Constant gain for derivative of $e'_\theta$ |
| $e'_y$ | Lateral error at the near point in haptic guidance model |
| $e'_\theta$ | Yaw error at the far point in haptic guidance model |
| $K_1$ | Constant gain for haptic guidance torque |
| $t'_f$ | Look ahead time at far point in haptic guidance model |
| $t'_n$ | Look ahead time at near point in haptic guidance model |

**2.6 Steering column dynamics**

An automotive steering column is a device intended primarily for connecting the steering wheel to the steering mechanism. Considering that the steering wheel is jointly actuated by the driver input torque, haptic guidance torque, and aligning torque, the steering wheel system dynamics

can be expressed as

$$J_s \ddot{\varphi} + B_s \dot{\varphi} = T_d + T_h + T_a \tag{2.4}$$

where $J_s$, and $B_s$ are the steering system inertia and damping, respectively. In addition, there is a transmission ratio between the steering wheel angle and front wheel steer angle, which is given by

$$\delta = K_t \varphi \tag{2.5}$$

The block diagram of transfer function for the steering column dynamics is shown in Figure 2.10. The output of the steering system, front wheel steer angle $\delta$, is the input of the vehicle dynamics system.

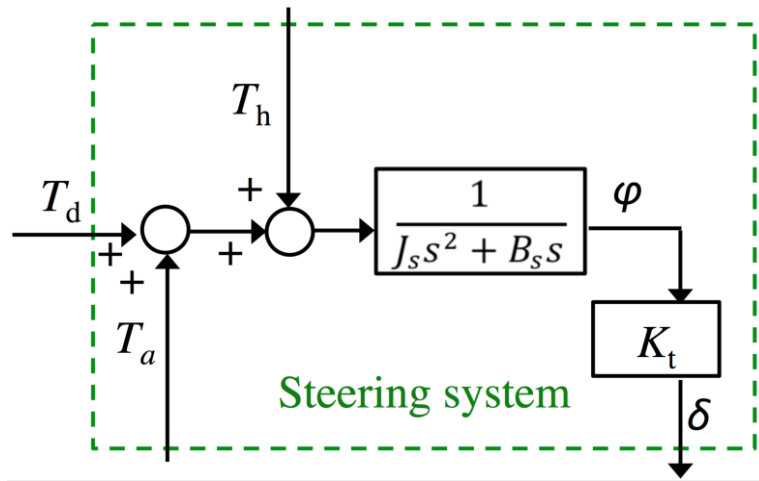

Figure 2.10 Block diagram of transfer function for the steering column system.

**2.7 Vehicle dynamics**

If a vehicle travels at constant speed, and without roll, the vehicle vertical height can be neglected and only the lateral and yawing motions need to be considered. The vehicle is represented as a rigid body projected to the ground. The lateral motion and yaw motion of the vehicle will generate slip angles at the four tires. A lateral force will be produced at the tire in response to the side-slip angle. If the left and right tire side-slip angles are equal, the steer angle is small and there is a negligible roll motion, it is suitable to consider the left and right tires of the front and rear wheels to be concentrated at the intersecting point of the vehicle x-axis with the front and rear axles as shown in Figure 2.11. In this way, a four-wheeled vehicle could be transformed to an equivalent two-wheeled vehicle or bicycle model, which makes the analysis of vehicle motion simpler. The variables and parameters of equivalent bicycle model are shown in Table 2.5.

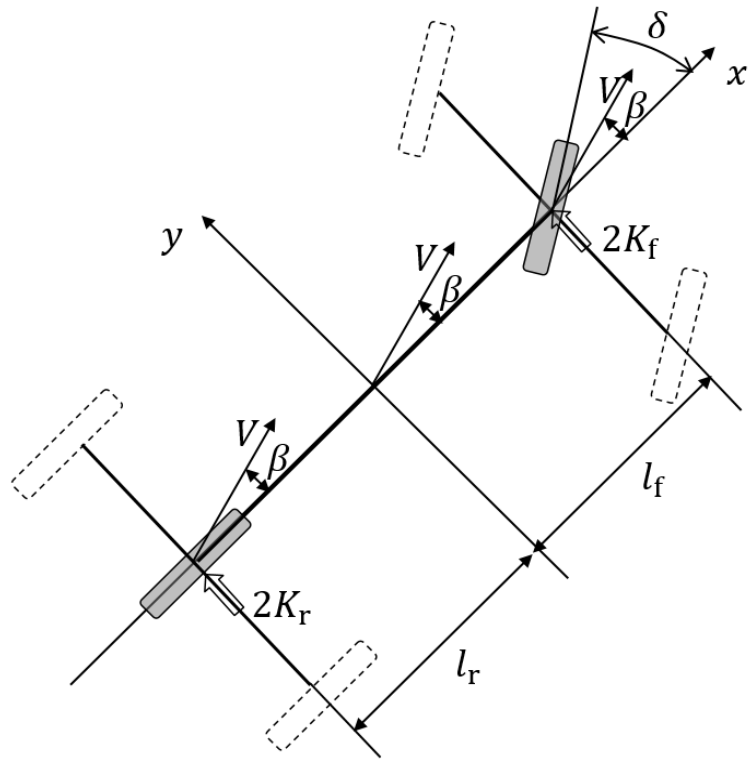

Figure 2.11 Equivalent bicycle model, adapted from [79].

Table 2.5 Variables and parameters of vehicle system.

|   | **Definition** |
|---|---|
| $E_t$ | The sum of pneumatic trail and castor trail |
| $I$ | Vehicle yaw moment inertia |
| $K_f$ | Cornering stiffness of front tire |
| $K_r$ | Cornering stiffness of rear tire |
| $K_s$ | Spring constant converted to the kingpin |
| $l_f$ | Longitudinal position of front wheels from vehicle center of gravity |
| $l_r$ | Longitudinal position of rear wheels from vehicle center of gravity |
| $m$ | Vehicle mass |
| $r$ | Yaw rate |
| $v$ | Vehicle speed |
| $\beta$ | Side slip angle |
| $\delta$ | Front wheel steer angle |

The vehicle model is based on the linearized bicycle model. The linearized vehicle dynamics to describe the vehicle plane motion are given as follows [79]:

$$mv\frac{d\beta}{dt} + 2(K_f + K_r)\beta + \left\{mv + \frac{2}{v}(l_f K_f - l_r K_r)\right\}r = 2K_f\delta \qquad (2.6)$$

$$2(l_f K_f - l_r K_r)\beta + I\frac{dr}{dt} + \frac{2(l_f^2 K_f + l_r^2 K_r)}{v}r = 2l_f K_f\delta \qquad (2.7)$$

The vehicle dynamics interact with the steering column dynamics via the aligning torque, which is given by [79]:

$$T_a = K_{aln}\left(\beta + \frac{l_f r}{v} - \delta\right) \qquad (2.8)$$

$$K_{aln} = 2E_t K_f K_t \left(\frac{1}{1 + 2E_t K_f/K_s}\right) \qquad (2.9)$$

The inputs and outputs of the vehicle dynamics system are shown in Figure. 2.12.

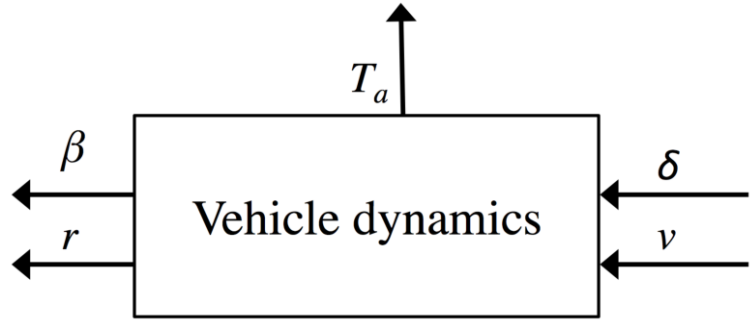

Figure 2.12 Inputs and outputs of the vehicle dynamics system.

## 2.8 Summary

This chapter comprehensively describes the linear model of the driver-vehicle-road system used in the thesis. A diagram of the whole system is shown in Figure 2.13.

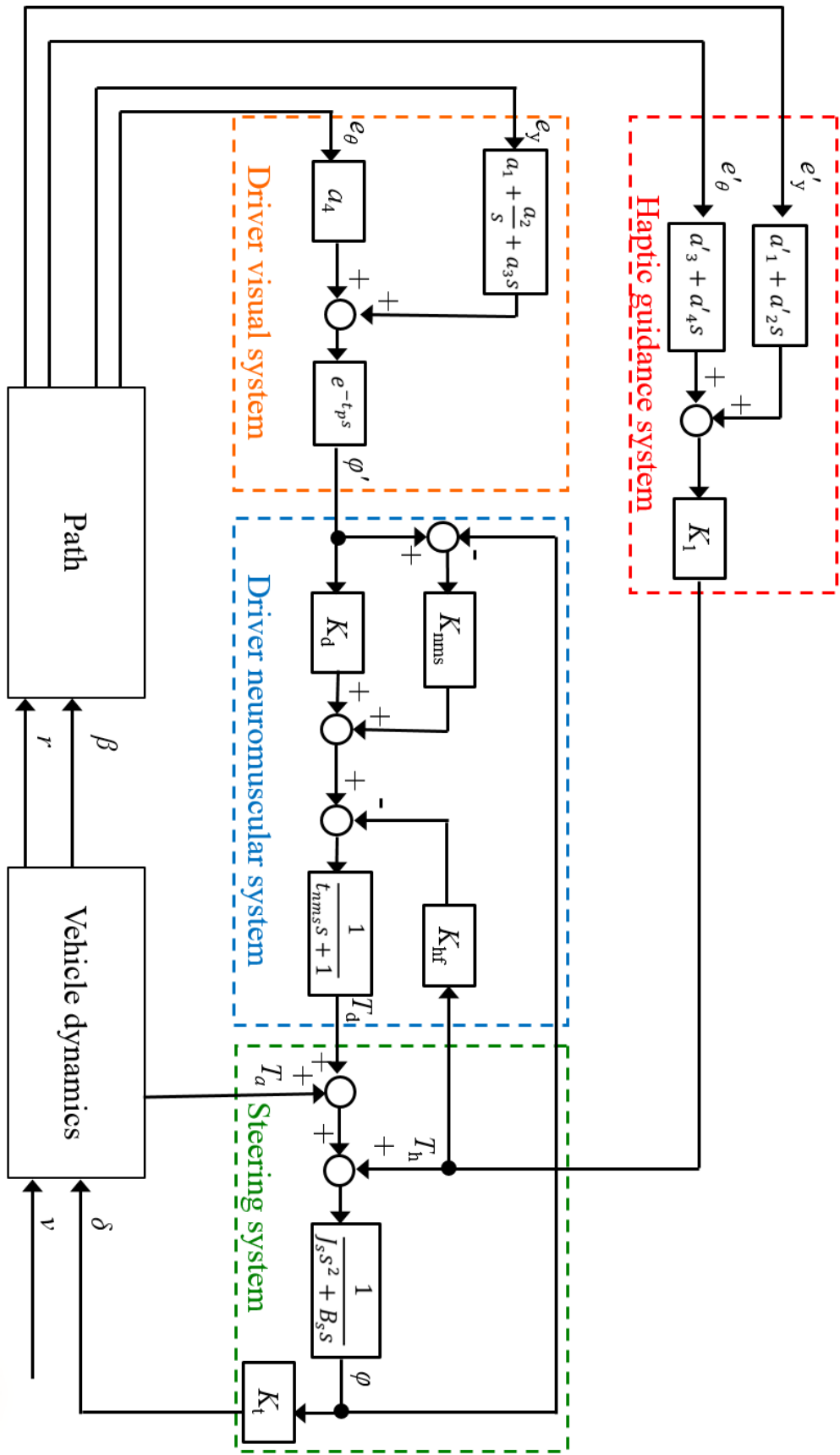

Figure 2.13 The linear model of the driver-vehicle-road system.

The concluding remarks of the proposed linear model of the driver-vehicle-road system are as follows:

First, the model of driver visual system addresses different cases of degraded visual information, including visual occlusion from the far segment of road ahead and declined visual attention under fatigue. As a result, it is expected to investigate the driver behavior in cases of degrade visual information by numerical simulations.

Second, the model of driver neuromuscular system considers integrated feedback of visual and haptic information via target steering angle to torque gain $K_d$, and neuromuscular reaction gain for haptic feedback $K_{\text{hf}}$. As a result, it is expected to investigate the effect of haptic guidance on driver behavior by numerical simulations.

Third, the model of the haptic guidance system is based on a two-point visual model which mimics the human driver, and applies a PD controller to compensate the limitation of the driver's derivative control action. It is hypothesized that the proposed haptic guidance system is capable of improving driver lane following performance in the cases of degraded visual information. This hypothesis will be tested by experimental and numerical studies.

In the next chapter, the effectiveness of the proposed haptic guidance system will be evaluated by driving simulator experimental studies in cases of normal and degraded visual information, and the driver behavior will be analyzed.

*Chapter 3*

*Experimental Studies and Results*

# 3 Experimental Studies and Results

As proposed in Section 1.4, the need of haptic information provided by the haptic guidance system could be based on the reliability of visual information perceived by a driver. To be specific, the supplementary haptic information should be able to compensate the performance decrement caused by degraded visual information, and should not induce additional workloads when visual information is reliable. To evaluate the effect of the haptic guidance system on driver behavior in cases of normal and degraded visual information, three experimental studies were conducted and described in this chapter. Experimental study I addressed the driver behavior in the case of normal visual information. It focused on the effect of different degrees of haptic guidance on driver gaze behavior, lane following performance, workload and driver reaction to a critical event. Experimental studies II and III addressed the driver behavior in the cases of degraded visual information. According to the driver model introduced in Section 2.3, driver perception of visual information is mainly influenced by look-ahead points and processing time delay. Visual occlusion from road ahead and longer processing time delay would induce decrement of driver performance. In experimental study II, the driving conditions of visual occlusion from road ahead were conducted, and the effect of haptic guidance on driver behavior was evaluated. In experimental study III, longer processing time delay was caused by fatigue driving condition in which driver visual attention was declined, and the effect of haptic guidance on driver behavior was evaluated.

**3.1 Experimental study I: Effect of different degrees of haptic guidance on driver behavior**

**3.1.1 Introduction**

Recent developments of driver support systems have shown promising applications for using continuous force feedback on the steering wheel to assist drivers with lane following tasks [4, 5, 25]. The haptic guidance system enables both the driver input torque and haptic guidance torque to contribute to the steering maneuver [14]. To evaluate the effect of haptic guidance system on driver behavior, driver steering behavior including steering effort and steering wheel movement, lane following performance, and subjective feedback have been widely investigated [4, 14, 27]. However, studies on driver visual behavior influenced by the haptic guidance system have been limited.

The driver's visual behavior, which is crucial for a steering task in manual driving, is assumed to be influenced by haptic information from active torques provided by the haptic guidance steering system. Furthermore, different degrees of haptic guidance would have different influence on driver visual behavior. Driver eye movements and gaze patterns in manual driving have been studied for a couple of decades [52, 54]. Eye movements direct steering movements, and the steering performance is significantly affected by the correlation coefficient and relative timing of eye–steering coordination [80]. In the frequency domain, the relationship between gaze and steering movements can also be observed in the lower frequency range [81]. When the optimal correlation is disturbed, the driving performance is seriously impaired [82]. Accordingly, driver gaze movements and their correlation to steering movements have been used to indicate driver distraction and drunk driving [83, 84]. Hence, driver visual behavior is highly related to driving conditions. The question here is what the driver visual behavior would be like when the haptic guidance system is applied.

Therefore, the goal of this experiment is to test the proposed haptic guidance system by investigating its effect on lane following performance, driver workload, and especially on driver visual behavior. Driver visual behavior will be measured by percent road center which indicates driver visual attention on the road ahead. It is hypothesized that the proposed haptic guidance is capable of providing reliable feedback to the driver, and as a result, driver visual attention on road ahead might be lower when driving with the haptic guidance system. Moreover, the lane following performance will be increased, and driver workload will be reduced. In addition, different degrees of haptic guidance will be compared in order to find an appropriate degree. It is hypothesized that relatively lower degree of haptic guidance might be better as visual information is reliable in the case of driving with normal visual information. Furthermore, the measured data in this experiment are used for driver model identification in Section 4.2 to test the proposed driver model in which integrated feedback of visual and haptic information is addressed.

### 3.1.2 Experimental design

A driving simulator experiment was conducted with 15 participants recruited from the University of Tokyo. The experiment was approved by the Office for Life Science Research Ethics and Safety, the University of Tokyo (No. 14-113). The details of the experimental design, including participants, apparatus, experiment conditions, experiment scenario and experiment protocol, are presented below.

#### 3.1.2.1 Participants

Fifteen healthy males were recruited to participate in the experiment. Their age ranged from 21 to 54 (mean = 26.6 years old, SD = 8.7), and all had a valid Japanese driver's license for at least one year (mean = 5.4 years, SD = 7.2). In response to the question of driving frequency, three participants reported to drive once a week, and others less than once a week. They all had normal or corrected-to-normal vision when performing the driving tasks in the experiment. Each participant received monetary compensation for the involvement in the experiment.

#### 3.1.2.2 Apparatus

The experiment was conducted in a high-fidelity driving simulator (Mitsubishi Precision Co., Ltd., Japan), as shown in Figure 3.1. The driving simulator consisted of brake and accelerator pedals, an electric steering system, an instrument dashboard, and two rear-view mirrors. In addition, a 140° field-of-view of driving scene was visualized by three projectors. To emulate the feeling of on-road driving, self-aligning torque was generated by the electric steering system, two stereos were used to provide the engine sound, and other vehicles appeared at various intervals. Raw data of driving performance were recorded in the host computer at a sampling rate of 120 Hz.

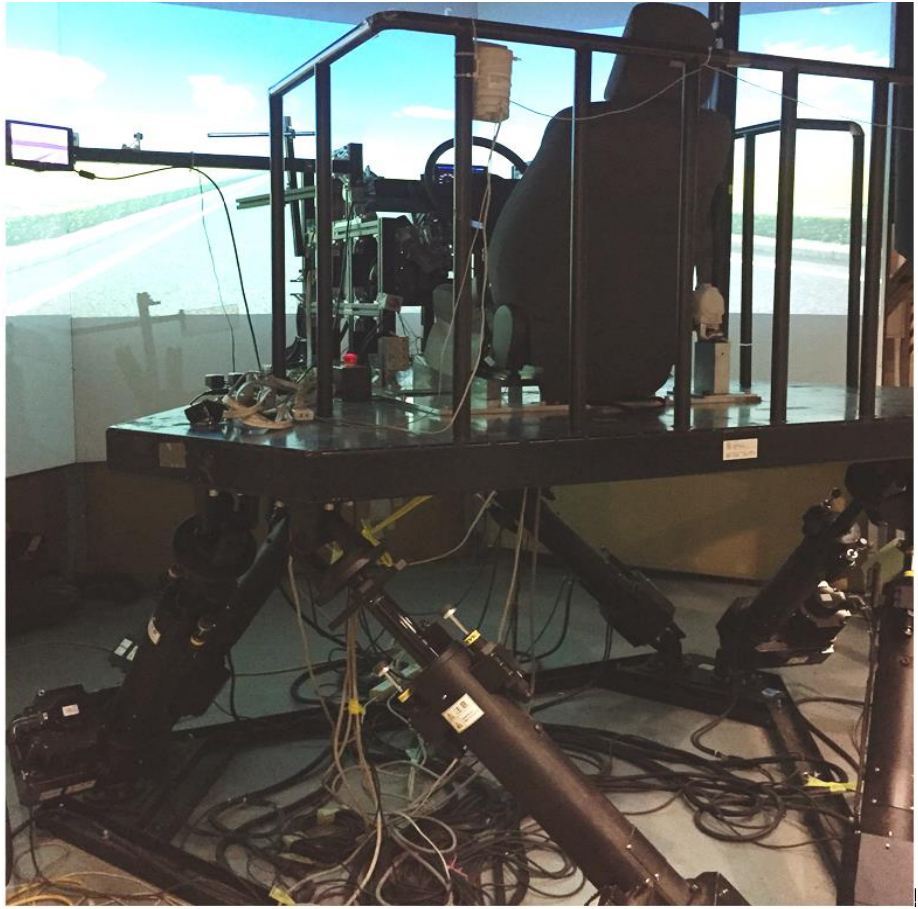

Figure 3.1 Driving simulator used in the experiment.

In the driving simulator, an electric steering system was connected to the host computer through a controller area network communication. As shown in Figure 3.2, the torque sensor was used to measure the driver input torque, and the sensor resolution was 0.005 N·m. The angular position sensor was used to measure driver steering angle, and the sensor resolution was 0.1 degrees. The motor was used to provide haptic guidance torque to the driver. It should be noticed that the measured torque was the approximation of driver input as the $J\ddot{\varphi}$ was negligible during a smooth driving, which is given by

$$T_m = T_d - J\ddot{\varphi} \approx T_d \qquad (3.1)$$

where $T_m$ is the measured input torque by the torque sensor, $T_d$ is the driver input torque, $J$ is the inertia of the steering wheel, and $\varphi$ is the steering wheel angle.

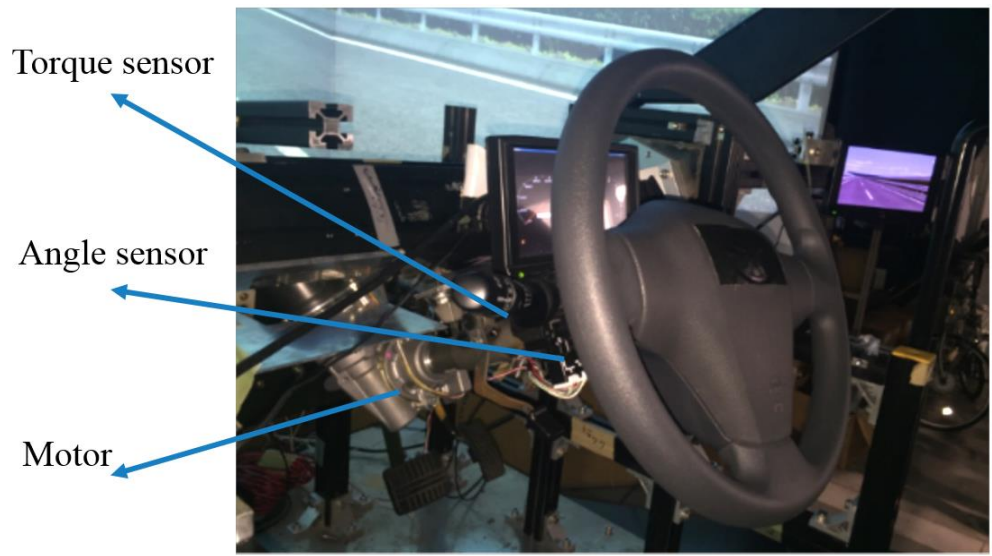

Figure 3.2 Electrical steering system used in the driving simulator.

An eye-gaze tracking system (Smart Eye Pro, Smart Eye AB, Sweden) was used to measure the eye movements of the participants, as shown in Figure 3.3. The system comprised two infrared flashes and three cameras, and it did not cause any physical burden to the participants during the measurements. The cameras were adjusted at three locations in front of the participants to ensure a high-quality measurement. The camera calibration ensured an average difference of less than 0.2 pixels by projecting a chessboard in front of the three cameras. Gaze accuracy, which included standard deviation and bias, was ensured to be less than 0.5 degrees in the gaze calibration. Original data were recorded at a sampling rate of 60 Hz.

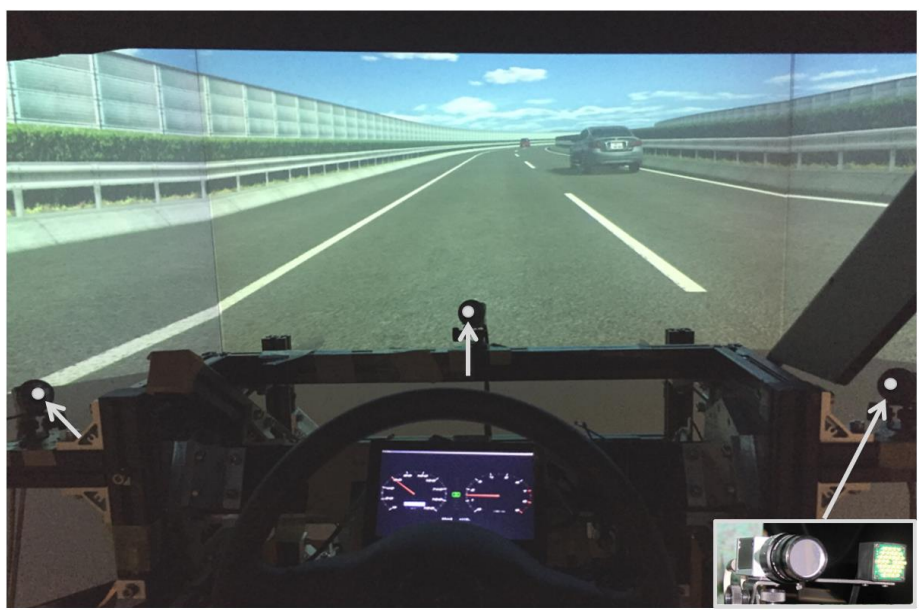

Figure 3.3 Driving environment and Smart Eye Pro system.

### 3.1.2.3 Experiment conditions

Each participant drove five trials that correspond to five driving conditions. Two different degrees of haptic guidance were addressed: haptic guidance with a normal feedback gain (HG-normal), and haptic guidance with a double feedback gain (HG-strong). Manual driving without haptic guidance was conducted as a comparison. In addition, two kinds of automated lane tracking were conducted: Hands-standby and Hands-free. To partially counterbalance the trial order within the participants, Latin squares were used in the order design to mitigate the order effect, as shown in Table 3.1.

Table 3.1 Partially counterbalanced design of trial order by Latin squares.

| Participants | 1st trial | 2nd trial | 3rd trial | 4th trial | 5th trial |
|---|---|---|---|---|---|
| No.1 | 1 | 2 | 5 | 3 | 4 |
| No.2 | 2 | 3 | 1 | 4 | 5 |
| No.3 | 3 | 4 | 2 | 5 | 1 |
| No.4 | 4 | 5 | 3 | 1 | 2 |
| No.5 | 5 | 1 | 4 | 2 | 3 |
| No.6 | 4 | 3 | 5 | 2 | 1 |
| No.7 | 5 | 4 | 1 | 3 | 2 |
| No.8 | 1 | 5 | 2 | 4 | 3 |
| No.9 | 2 | 1 | 3 | 5 | 4 |
| No.10 | 3 | 2 | 4 | 1 | 5 |
| No.11 | 1 | 2 | 5 | 3 | 4 |
| No.12 | 2 | 3 | 1 | 4 | 5 |
| No.13 | 3 | 4 | 2 | 5 | 1 |
| No.14 | 4 | 5 | 3 | 1 | 2 |
| No.15 | 5 | 1 | 4 | 2 | 3 |

1: Manual; 2: HG-normal; 3: HG-strong; 4: Hands-standby; 5: Hands-free

It should be noticed that the two conditions of automated driving are not so much related to the main purpose of this thesis, as no haptic information was provided to the drivers in those two conditions. Thus, the results in the conditions of Hands-standby and Hands-free are not shown in this thesis, which leaves three driving conditions corresponding to three degrees of haptic guidance: Manual, HG-normal, and HG-strong, when presenting the results.

Different degrees of haptic guidance was determined by the value of $K_1$ (see Equation 2.3 in Section 2.5.2): The haptic guidance with a double feedback gain (HG-strong, $K_1 = 0.5$) was half of the gain set for automated lane tracking ($K_1 = 1.0$), and the haptic guidance with a normal feedback gain (HG-normal, $K_1 = 0.25$) was half of the gain set for HG-strong. The values of $a'_1$, $a'_3$, and $a'_4$ were decided as 1.9, 38, and 1.9, respectively, through a trial-and-error method. The method was generally divided into two steps. The first step was to find rough solution of automated driving with aggressive tuning. The second step was to find desired solution with small tweaks in order to achieve good performance.

**3.1.2.4 Experiment scenario**

The driving environment was a two-lane expressway road with 3.6 m width lanes and an emergency lane on the left, as shown in Figure 3.3. Lane markings were solid lines and dashed lines. Other vehicles appeared at various intervals in the right lane, and some of them changed to the left lane after passing the ego vehicle. The driving speed of the ego vehicle was fixed at 60 km/h; thus, the participants did not need to operate the accelerator and brake pedals unless it was necessary. The reason of fixing the driving speed was that gaze behavior and steering performance are highly related with speed choice. By removing the variability of speed, the assessment of gaze behavior and steering performance would be less difficult.

The driving course consisted of straight roads and curves, as shown in Figure 3.4. The length of each curve was 314 m, and the radius was either 200 or 300 m. The total length of the driving course was 11.6 km, and the main section was 10 km. After the main section, there was a critical event section in which the participants needed to deal with a critical event.

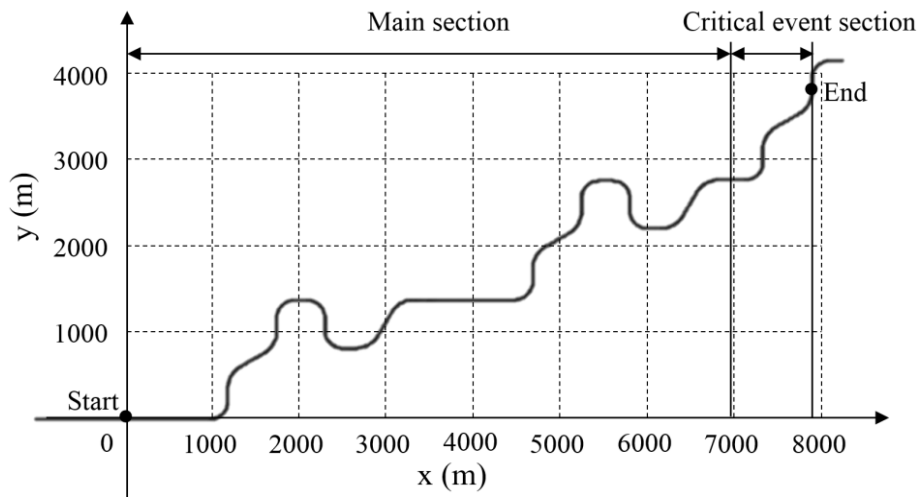

Figure 3.4 Driving course in the experiment.

As shown in Figure 3.5, when the critical event occurred, a lead vehicle stopped suddenly with a deceleration rate of 7 m/s$^2$ and with an identical distance of 100 m ahead of the ego vehicle in the different driving trials. It should be noted that different lead vehicles ran 100 m ahead of the ego vehicle at various intervals in the whole driving course. Moreover, the look of the lead vehicle and the place where the critical event occurred in the driving course varied in the different trials, so that the participants were not able to anticipate the critical event.

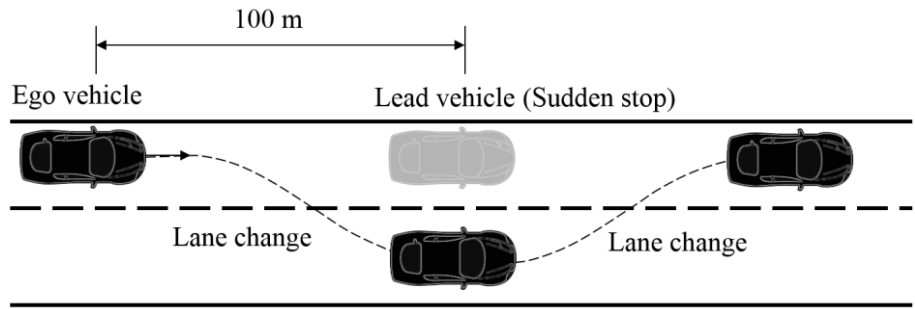

Figure 3.5 Schematic representation of the critical event and lane change maneuver to avoid forward collision.

There were no warnings when the critical event occurred. The participants were requested to detect the event by themselves, then to make a lane change maneuver to avoid forward collision with the stopped vehicle, and then to return to the left lane after passing. In the conditions of HG-normal and HG-strong, the haptic guidance steering system was not capable of guiding the drivers to perform a lane change, and instead, it kept working for lane following assistance purpose. Thus, the drivers needed to overrule the haptic guidance torque, and HG-strong required more driver effort to be overruled than HG-normal did. The haptic guidance steering system would stop providing active torques automatically when the ego vehicle entered the right lane, and the remaining task after the critical event was manual driving.

### 3.1.2.5 Experiment protocol

One day before the experiment, the participants were requested to avoid staying up late and were recommended to have a good night's sleep for approximately 7 h. In addition, caffeine and alcohol consumption was not allowed during the 12 h prior to the experiment.

On the experiment day, first, the participants signed a consent form, which explained the procedure of the experiment. The participants were naive to the purpose of the experiment. The participants then filled out a questionnaire about their personal demographic information and driving experience. After that, they got into the driving simulator and adjusted their seat to achieve a normal driving position. The participants were requested to grab the steering wheel in a "ten-and-two" position and to keep the vehicle on the centerline of the lane as accurately as possible while driving. The participants were requested to always keep driving along the left lane, and the participants were also asked to follow Japanese traffic rules all the time. In order to mitigate the learning effect, the participants drove five practice trials, which corresponded to the five driving conditions, so that they can familiarize themselves with the driving simulator, haptic guidance steering system, road track, and the critical event. Each practice trial lasted for 2 min.

After the practice trials, the participants were allowed to rest for 5 min. After the rest, the formal experimental session started. In this session, the participants drove 5 trials, and the order of trials was based on the partially counterbalanced design. After each trial, the participants were asked to complete a questionnaire, and to rest for 10 min which helped to mitigate the tired feeling caused by the driving task in the simulator. The entire experiment took about 170 min per participant.

**3.1.3 Data analysis**

Experimental data from one participant were ignored due to poor quality of gaze signal, leaving a sample of 14 participants when analyzing the data and showing the results. Initial eye movement data were collected in the Smart Eye Pro system, and initial driving performance data were collected in the driving simulator. Further signal processing steps were conducted in the MATLAB, and are comprehensively presented in the following parts. In addition, statistical analysis on the post-processed data is explained.

**3.1.3.1 Divided sections of driving path**

As shown in Figure 3.4, the driving course consists of some straight roads and curves. Driver behavior has been found quite different among driving on the straight lane, on the curve, approaching the curve, and leaving the curve. Thus, one curve negotiation process has been divided into five sections, as shown in Figure 3.6: the straight lane before a curve, approaching the curve, along the curve, leaving the curve, and the straight lane after the curve. There is a junction point between the straight lane and the curve. The driving course around the junction point refers to approaching the curve or leaving the curve. As shown in Figure 3.7, the solid line illustrates the steering wheel angle of one participant when negotiating the curve in the condition of manual driving. In terms of an ideal situation shown as dotted lines in the figure, the steering wheel angle remains zero on the straight lane before the curve, increases linearly when approaching the curve, stays constant along the curve, decreases linearly when leaving the curve, and remains zero on the straight lane after the curve.

The driver behavior in this experiment was investigated in terms of on straight lanes and along curves. The data before and after the 50 m of the junction point were removed, leaving the data representing "on the straight lane (S1 and S5)" and "along the curve (S3)". As there were multiple straight lanes and curves, the presented result was obtained by the mean value of the measured variable across all the straight lanes or curves in the main section of the driving course (see Figure 3.4).

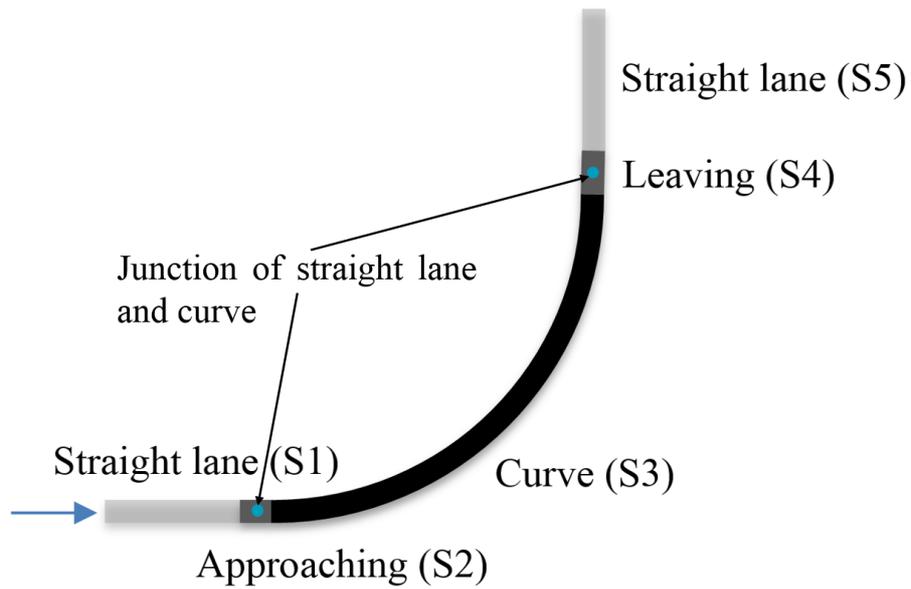

Figure 3.6 Divided sections of one curve negotiation.

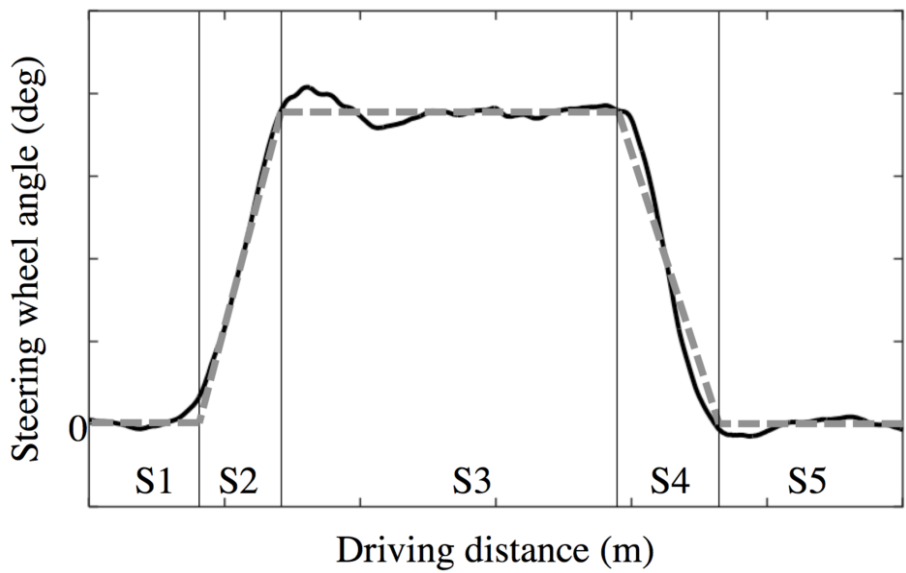

Figure 3.7 Steering wheel angle during curve negotiation. The curve negotiation is divided into five sections by vertical grey lines: straight lane before the curve (S1), approaching the curve (S2), along the curve (S3), leaving the curve (S4), and straight lane after the curve (S5).

**3.1.3.2 Visual behavior measurement**

Driver visual behavior is crucial for a steering task in natural driving. The initial eye movement data were collected in the Smart Eye Pro system. Gaze data from blinks and saccades were removed, leaving fixations, which represented pursuit eye movements. Further signal processing steps and plots were conducted in MATLAB.

A heat map is a graphical representation of data where the individual values contained in a matrix are represented as colors. The distribution of gaze direction along a curve was observed from one typical participant using a heat map, as shown in Figure 3.8. The heat map uses red color to indicate a high frequent gaze direction and blue color to indicate a low frequent gaze direction. It can be observed that the high frequent gaze direction is concentrated on a look-ahead point for curve negotiation. The low frequent gaze directions are somewhere far beyond the look-ahead point and somewhere on the adjacent lane.

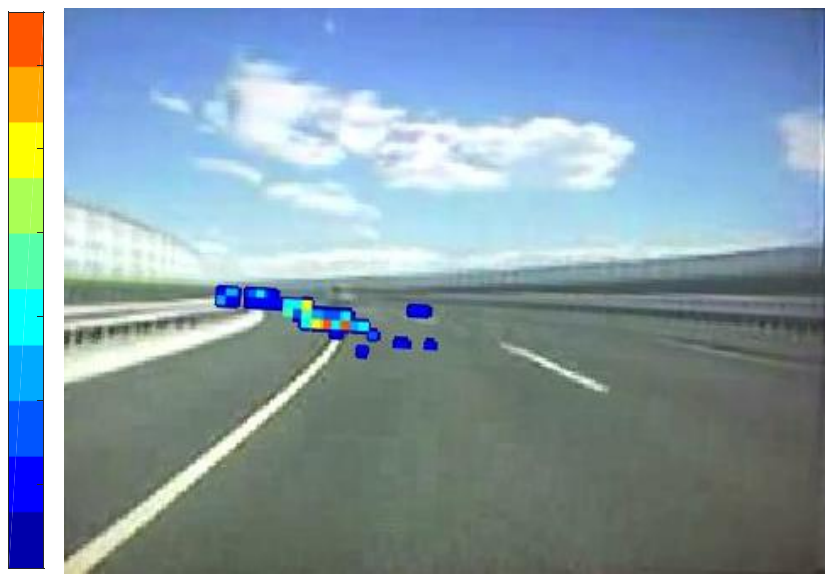

Figure 3.8 Distribution of gaze direction along a curve given by a heat map. The heat map uses red color to indicate a high frequent gaze direction and blue color to indicate a low frequent gaze direction.

The heat map gives a direct illustration of the gaze distribution on the road ahead. To quantitatively analyze driver gaze distribution, percent road center (PRC) was calculated as another measurement of driver visual behavior. PRC has been widely used to analyze the degree of drivers' gaze dispersion during a driving task [85]. PRC is defined as the proportion of gaze data points falling within the road center area. The road center area is defined as a circular area with a radius of 6° centered on the road center point, as shown by the black circle in Figure 3.9. The road center point is determined as the participant's most frequent gaze direction along the curve [85]. Corresponding with Figure 3.8, the most frequent gaze direction would be the red point in the heat map. The scatter plot of gaze points, as shown in Figure 3.9, represents the gaze directions. A higher value of PRC indicates a lower degree of gaze dispersion or a higher degree of gaze concentration.

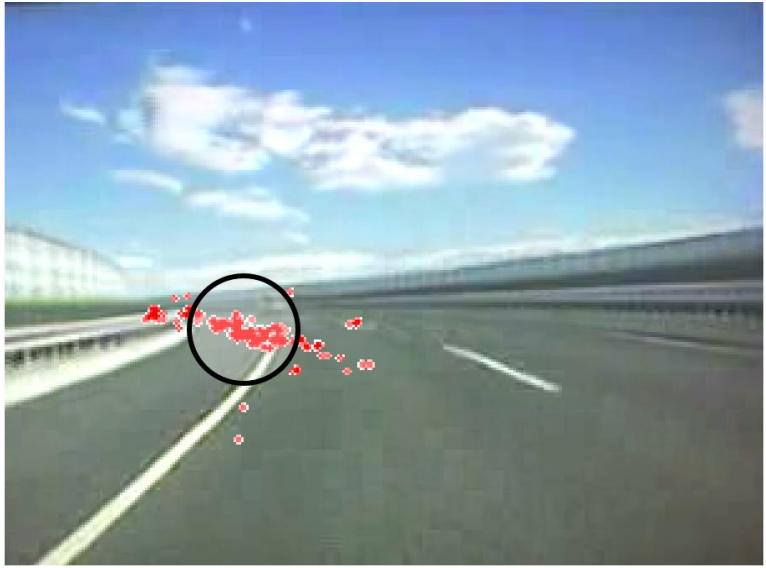

Figure 3.9 Scatter plot of gaze direction on a curve. The black circle represents the road center area centered on the participant's most frequent gaze direction.

### 3.1.3.3 Lane following performance measurement

As mentioned in Section 3.1.3.1, the driver behavior in this experiment was investigated in terms of on straight lanes and along curves. The lane following performance on straight lanes was measured by standard deviation of lane position. The lane following performance along curves was measure by time-to-lane crossing and steering wheel reversal rate.

- Standard deviation of lane position (SDLP)

SDLP is calculated by

$$\text{SDLP} = \sqrt{\frac{1}{N-1}\sum_{i=1}^{N}(x_i - \mu)^2} \ , \tag{3.2}$$

where $x_i$ is the lateral position of vehicle, $N$ is the number of samples taken in each part of driving course, and $\mu$ is the mean lateral position of vehicle [86], as shown in Figure 3.10 (a). Vehicle trajectory with a higher value of SDLP is shown in Figure 3.10 (a), and with a lower value of SDLP is shown in Figure 3.10 (b). SDLP increases when lateral driving control effort decreases or steering operation is unstable. By comparing these two vehicle trajectories, it can be observed that a decreased SDLP indicates a higher lane following performance or a decreased lane departure risk.

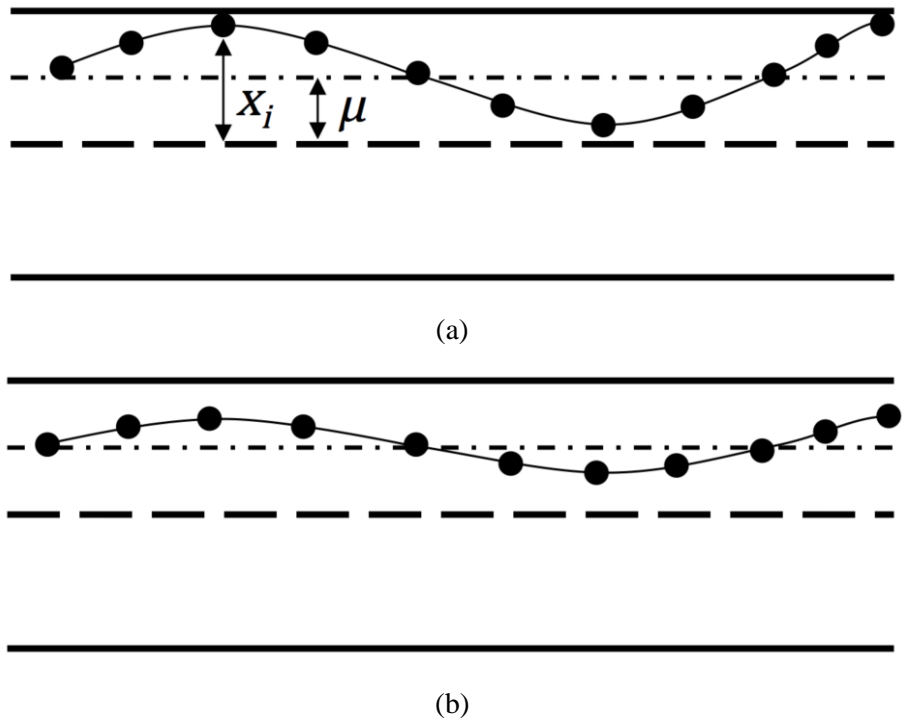

(a)

(b)

Figure 3.10 Diagram of standard deviation of lane position. Black dots represent the positions of the vehicle. (a) A higher value of SDLP (b) A lower value of SDLP.

- Time-to-lane crossing (TLC)

TLC was obtained by evaluating the discrete points along the predictive path of the vehicle center of gravity, until a point lay outside of the road boundary, as shown in Figure 3.11. The road boundaries were represented by the discrete points of two polynomial equations. When calculating the predictive vehicle path, it is assumed that no further steering intervention would be taken by the driver [87]. When the distance between the points on the predictive path and the lane boundary was less than a threshold, the lane crossing point was obtained, and the threshold was set to 0.1 m in this study. TLC was calculated by the following equation,

$$\text{TLC} = \frac{\text{DLC}}{v} \qquad (3.3)$$

where DLC is the distance to lane crossing along the projected vehicle path, and $v$ is the driving speed.

In order to calculate the mean value of TLC among all the curves in the driving course, the minimum 10% of all TLC was averaged [57]. A higher value of TLC indicates a higher lane following performance or a decreased lane departure risk.

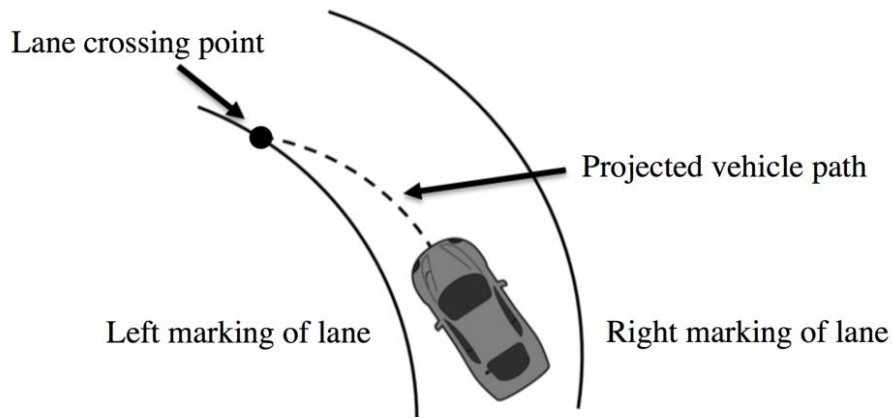

Figure 3.11 Diagram of time-to-lane crossing.

- Steering wheel reversal rate (SWRR)

SWRR is defined as the number of changes in steering wheel direction per minute. The quantity of steering reversals measured is the number of times the direction of steering wheel movement is reversed through a finite angle α, as shown in Figure 3.12. When calculating SWRR, all reversals in one period with the steering wheel angle signal that are greater than α are counted. In this thesis, α was set to 3°, which presents large corrections of steering maneuver. It has been found that drivers tend to make more large corrections instead of small corrections when they have a lower lane following performance, or a higher lane departure risk [88].

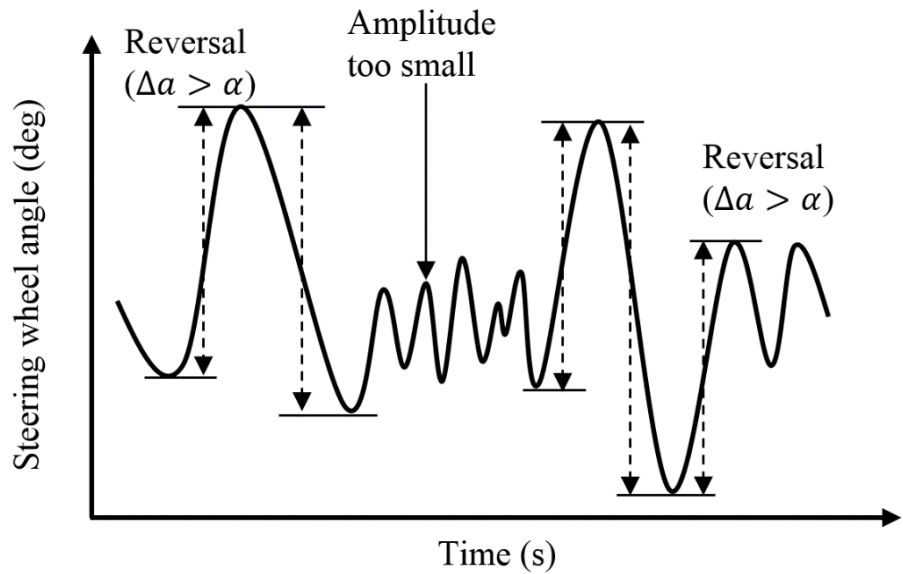

Figure 3.12 Diagram of steering reversals, adapted from [89].

### 3.1.3.4 Reaction to the critical event

When the critical event occurred, the participants needed to make a quick response. The participants were requested to detect the event by themselves, then to make a lane change maneuver to avoid forward collision with the stopped vehicle. To investigate the participants' reaction to the critical event, the vehicle trajectory during the lane change maneuver were analyzed.

### 3.1.3.5 Subjective evaluation

A subjective evaluation is typically based on subjective feedback from the participants as opposed to objective, measured feedback that has been introduced above.

In this experiment, the perceived workload of driving task was evaluated by subjective feedback. The perceived workload of driving task was rated by using the NASA-Task load index (NASA-TLX) [90]. The participants were asked to use the NASA-TLX to assess their workload at the end of each driving trial in the experiment. The index consisted of six items. In the first step, each item of the index was investigated separately to obtain the scale score. The six items are as follows:

(a) Mental Demand: How much mental and perceptual activity was required? Was the task easy or demanding, simple or complex?
(b) Physical Demand: How much physical activity was required? Was the task easy or demanding, slack or strenuous?
(c) Temporal Demand: How much time pressure did you feel due to the pace at which the tasks or task elements occurred? Was the pace slow or rapid?
(d) Performance: How successful were you in performing the task? How satisfied were you with your performance?
(e) Effort: How hard did you have to work (mentally and physically) to accomplish your level of performance?
(f) Frustration: How irritated, stressed, and annoyed versus content, relaxed, and complacent did you feel during the task?

The items are rated for each task within a 100-points range with 5-point steps. An example of scale score for Mental Demand is shown in Figure 3.13.

In the second step, individual weighting of the items was explored to obtain the weighted score. The weighted score was multiplied by the scale score for each item, and then the overall task load score was obtained.

In addition to an English version of NASA-TLX, a Japanese version of NASA-TLX [91] was also provided to the participant during the experiment.

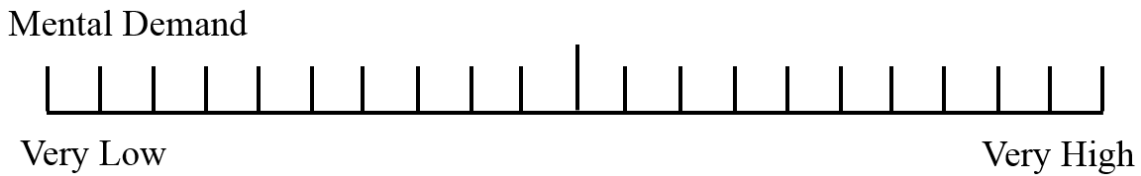

Figure 3.13 An example of scale score for Mental Demand in NASA TLX.

**3.1.3.6 Statistical analysis**

In statistical hypothesis testing, a result has statistical significance when it is very unlikely to have occurred given the null hypothesis. In terms of comparing three means (Manual, HG-normal, and HG-strong) for statistical significance in this experiment, the repeated measures analysis of variance (ANOVA) was used.

ANOVA is a statistical technique used to analyze the differences among group means (usually three or more means). It has been widely used in terms of practical problems, and it is less powerful than multiple two-sample t-test due to less type I error. Rejecting the null hypothesis when it is true is known as a type I error. The level at which a result is significant is known as the type I error rate, denoted by $\alpha$ or called the alpha level. Normally the significance level is set to 0.05 (or 5%). If the calculated $p$-value is less than 0.05, the null hypothesis that there is no difference between the means is rejected and it is concluded that a significant difference does exist.

A repeated measures ANOVA is referred to as a within-subjects ANOVA. A within-subjects design corresponds to an experiment in which the same group of subjects serves in more than one treatment. In this experiment, the same group of subjects participated all the driving conditions. The significance level when performing repeated measures ANOVA was set to 0.05 in this thesis.

Mauchly's test was conducted prior to the repeated measures ANOVA for validation purpose. This is because repeated measures ANOVA is particularly susceptible to the violation of the assumption of sphericity. Sphericity is the condition where the variances of the differences between all combinations of related groups are equal. Violation of sphericity is when the variances of the differences between all combinations of related groups are not equal. The repeated measures ANOVA needs to be corrected according to the degree of sphericity violation. Three kinds of corrections that are widely used include lower-bound estimate, Greenhouse-Geisser correction and Huynh-Feldt correction. These corrections rely on estimating sphericity. In this thesis, Greenhouse-Geisser correction, which is more conservative than Huynh-Feldt correction and less conservative than lower-bound estimate, is applied.

After repeated measures ANOVA, pairwise comparisons were used to show simple effects, and the significance level was set to 0.05. A common pairwise comparison method is called Fishers's least significance difference (LSD) method. Fisher's LSD method controls $\alpha$-level error rate for each pairwise comparison but family error rate was not controller in post ANOVA analysis. The pairwise comparison result by Fisher's least significance would be the same with a paired sample t-test comparison (with-in subjects design). In addition, Fisher's LSD is the least conservative or most powerful pairwise comparison method. In contrast, the post hoc Bonferroni pairwise comparison controls the family error rate by using $\alpha/g$ significance level, in which g is the number of pairwise comparisons. Accordingly, the post hoc Bonferroni pairwise comparison is more conservative or less powerful compared to Fisher's LSD method. In addition, the calculated *p*-value by Bonferroni method would be g times the *p*-value calculated by using Fisher's LSD method. In this thesis, Bonferroni method was used for pairwise comparison and the g was three in this experiment as there are three driving conditions that are compared.

**3.1.4 Results**

Results are presented for driver visual behavior, lane following performance, steering effort, and subjective evaluation.

**3.1.4.1 Visual behavior**

Driver visual behavior was measured by the heat map of gaze distribution on road ahead and the percent road center.

- Heat map of gaze distribution

Figure 3.14 shows an example of the distribution of gaze direction along a curve from an individual participant. The heat maps use red color to indicate a high frequent gaze direction and blue color to indicate a low frequent gaze direction. It can be observed that the driver's gaze directions are more centralized at one look-ahead point in all three conditions of Manual, HG-normal, and HG-strong. In addition, the low frequent gaze directions are somewhere far beyond the look-ahead point, and to some degree somewhere before and after the look-ahead point; this phenomenon is also quite similar among the three driving conditions.

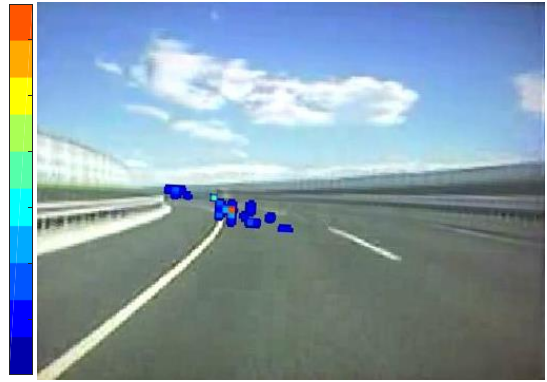

(a)

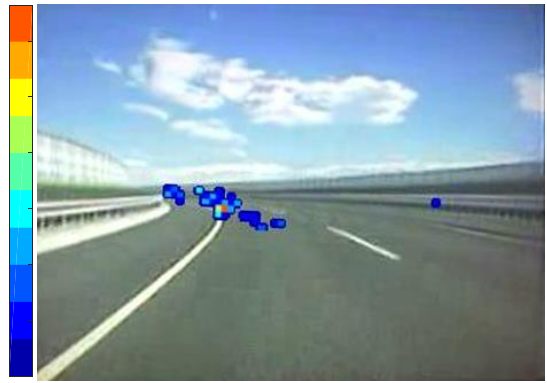

(b)

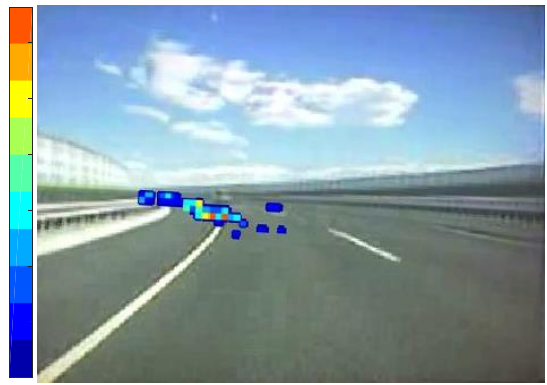

(c)

Figure 3.14 Comparison of gaze distribution among three driving conditions: (a) Manual, (b) HG-normal, (c) HG-strong.

- Percent road center (PRC)

To statistically study the distribution of gaze direction along curves, the mean value of PRC was calculated. PRC has been widely used to analyze the degree of drivers' gaze dispersion during a driving task. A higher value of PRC indicates a lower degree of gaze dispersion. The result of PRC along curves is shown in Figure 3.15. The mean value of PRC along curves was not significantly different among the three driving conditions ($F(2,26) = 2.021$, $p = 0.153$). Post hoc pairwise comparison results are shown in Table 3.2. From pairwise comparison, it can be observed that there was no significant difference between each pair of the three driving conditions. It indicates that the haptic guidance system did not significantly influence the driver visual attention on the look-ahead point along curves.

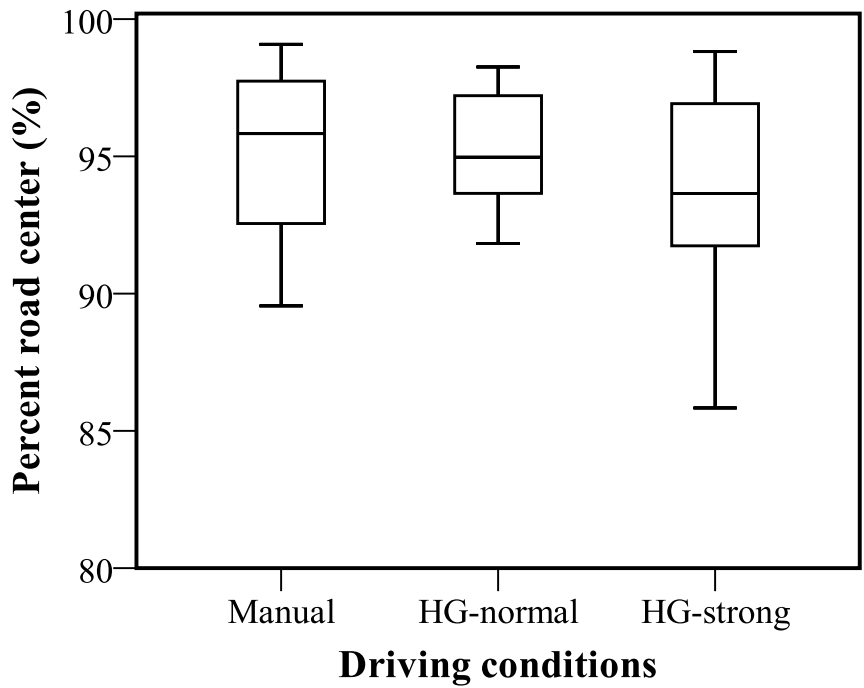

Figure 3.15 Percent road center along curves. A rectangle represents the middle 50% of a set of data. A horizontal line drawn through a rectangle corresponds to the median value of a set of data. An upper bar indicates the maximum value of a set of data, excluding outliers. A lower bar represents the minimum value of a set of data, excluding outliers.

Table 3.2 Post hoc test of percent road center.

|           | Manual | HG-normal | HG-strong |
|-----------|--------|-----------|-----------|
| Manual    | -      | -         | -         |
| HG-normal | 1.000  | -         | -         |
| HG-strong | 0.366  | 0.569     | -         |

**3.1.4.2 Lane following performance**

Lane following performance was measured by calculating SDLP on straight lanes, and also by calculating TLC and SWRR along curves.

- Standard deviation of lane position (SDLP)

The result of SDLP on straight lanes is shown in Figure 3.16. SDLP was significantly different among the driving conditions of Manual, HG-normal, and HG-strong ($F(2,26) = 4.559, p = 0.020$). Post hoc pairwise comparison results are shown in Table 3.3. From pairwise comparison, it can be seen that SDLP in the condition of HG-strong was significantly lower than Manual. There was no significant difference between Manual and HG-normal, and between HG-normal and HG-strong. The tendency of decreasing SDLP indicates that the haptic guidance system was capable of proving reliable haptic information on straight lanes, and the lane following performance was improved. Moreover, the strong haptic guidance was more effective than normal haptic guidance.

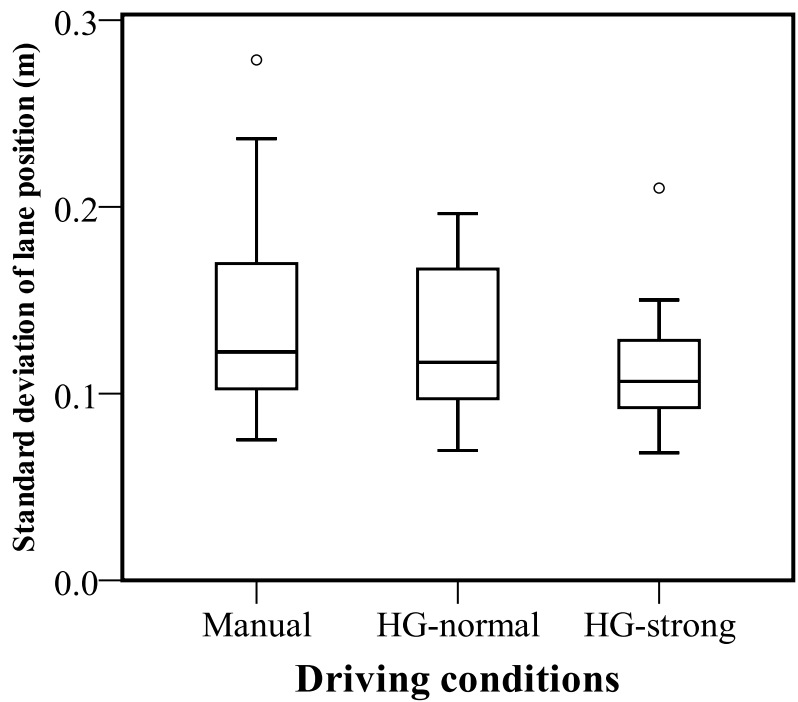

Figure 3.16 Standard deviation of lane position on straight lanes. A rectangle represents the middle 50% of a set of data. A horizontal line drawn through a rectangle corresponds to the median value of a set of data. An upper bar indicates the maximum value of a set of data, excluding outliers. A lower bar represents the minimum value of a set of data, excluding outliers. Outliers which are calculated as 1.5-3x the interquartile range, are plotted as individual circles.

Table 3.3 Post hoc test of standard deviation of lane position.

|  | Manual | HG-normal | HG-strong |
|---|---|---|---|
| Manual | - | - | - |
| HG-normal | 0.506 | - | - |
| HG-strong | 0.040* | 0.269 | - |

*: $p < 0.05$

- Time-to-lane crossing (TLC)

The result of TLC along curves is shown in Figure 3.17. TLC was not significantly different among the driving conditions of Manual, HG-normal, and HG-strong ($F(2,26) = 0.865$, $p = 0.433$). Post hoc pairwise comparison results are shown in Table 3.4. It indicates that HG-strong and HG-normal did not significantly improve lane following performance along curves as the participants' lane following performance was already satisfied in the condition of manual driving in the experiment.

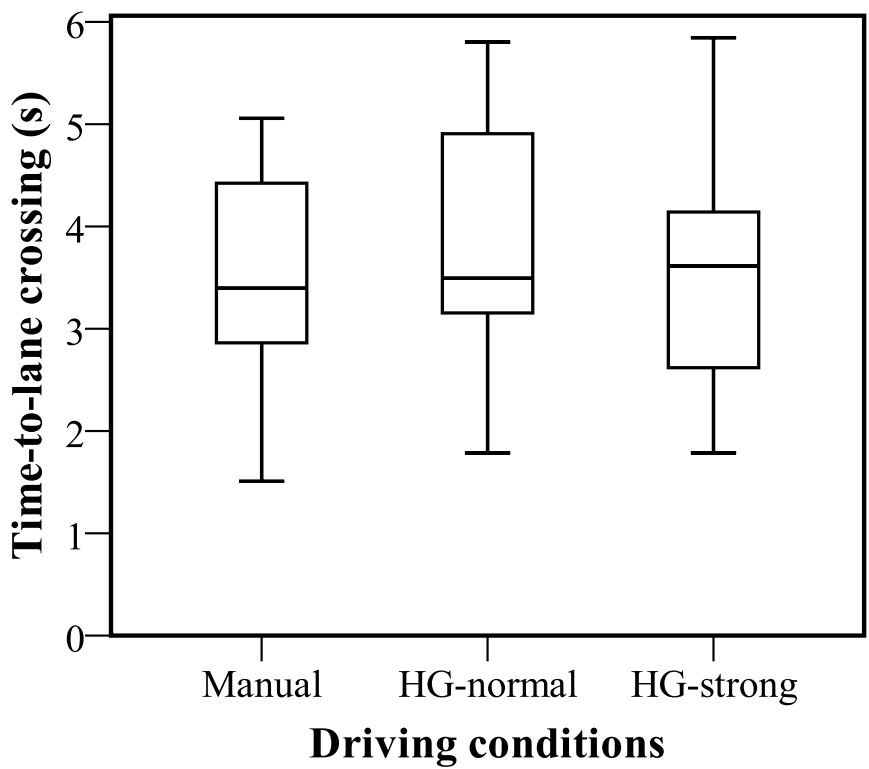

Figure 3.17 Time-to-lane crossing along curves. A rectangle represents the middle 50% of a set of data. A horizontal line drawn through a rectangle corresponds to the median value of a set of data. An upper bar indicates the maximum value of a set of data, excluding outliers. A lower bar represents the minimum value of a set of data, excluding outliers.

Table 3.4 Post hoc test of time-to-lane crossing.

|  | Manual | HG-normal | HG-strong |
| --- | --- | --- | --- |
| Manual | - | - | - |
| HG-normal | 0.174 | - | - |
| HG-strong | 1.000 | 1.000 | - |

- Steering wheel reversal rate (SWRR)

The result of SWRR along curves is shown in Figure 3.18. SWRR was not significantly different among the driving conditions of Manual, HG-normal, and HG-strong ($F(2,26) = 0.749$, $p = 0.483$). Post hoc pairwise comparison results are shown in Table 3.5. This result is in accordance with the result of TLC along curves. It indicates that HG-strong and HG-normal did not significantly improve driver steering performance along curves because the participants' steering performance was already satisfying in the condition of manual driving in the experiment.

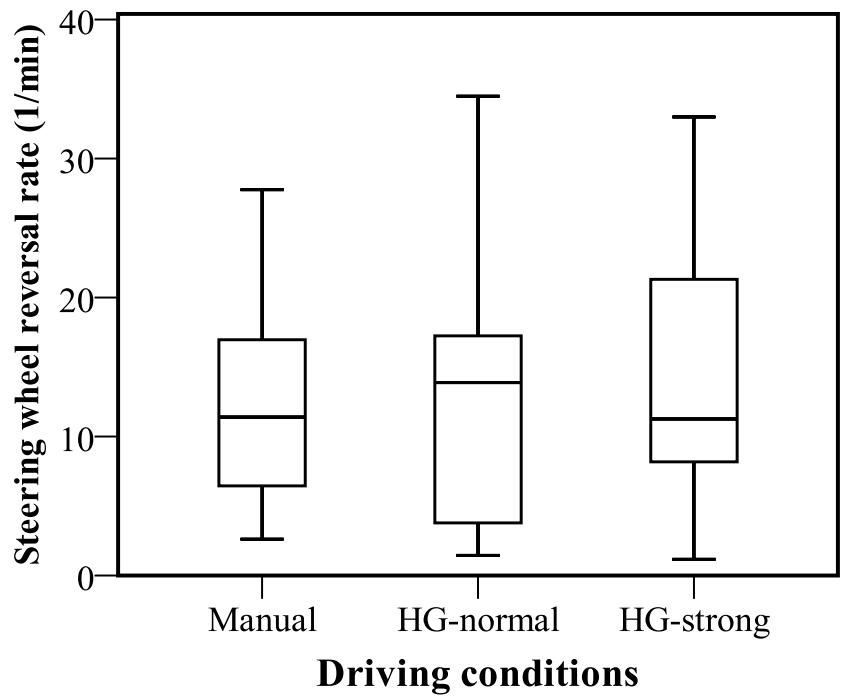

Figure 3.18 Steering wheel reversal rate along curves. A rectangle represents the middle 50% of a set of data. A horizontal line drawn through a rectangle corresponds to the median value of a set of data. An upper bar indicates the maximum value of a set of data, excluding outliers. A lower bar represents the minimum value of a set of data, excluding outliers.

Table 3.5 Post hoc test of steering wheel reversal rate.

|  | Manual | HG-normal | HG-strong |
|---|---|---|---|
| Manual | - | - | - |
| HG-normal | 1.000 | - | - |
| HG-strong | 1.000 | 1.000 | - |

**3.1.4.3 Driver steering effort**

The result of mean value of driver steering torque is shown in Figure 3.19. The mean value of driver torque along curves was significantly different among Manual, HG-normal, and HG-strong ($F(2,26) = 491.988$, $p < 0.001$). Post hoc pairwise comparison results are shown in Table 3.6. From pairwise comparison, it can be seen that HG-strong yielded a significantly lower driver torque than HG-normal and Manual. Moreover, HG-normal yielded a significantly lower driver torque than Manual. The result indicates that the participants' overall steering effort was reduced by haptic guidance, and strong haptic guidance led to more reduction in steering effort compared to normal haptic guidance.

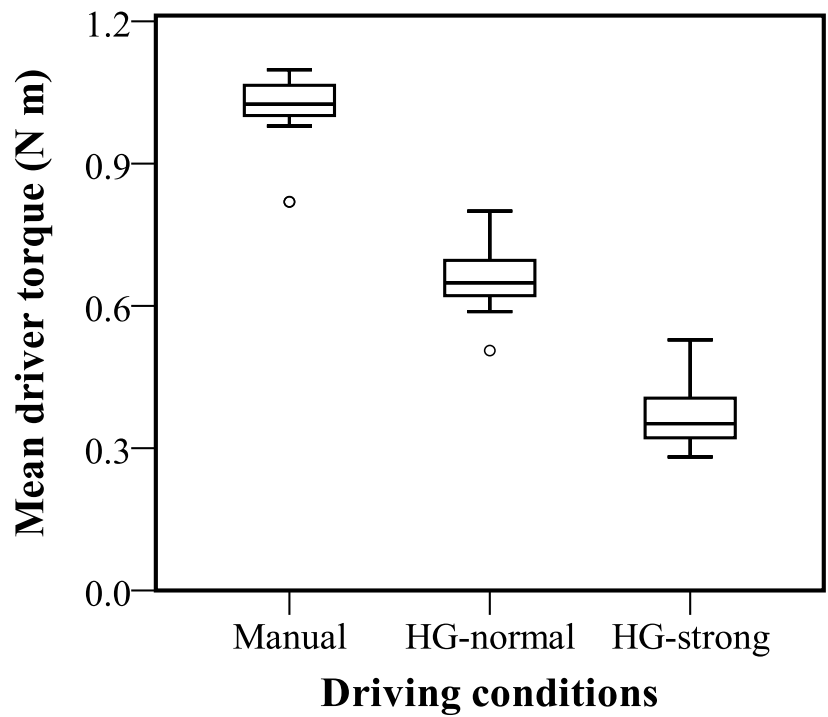

Figure 3.19 Mean value of driver torque along curves. A rectangle represents the middle 50% of a set of data. A horizontal line drawn through a rectangle corresponds to the median value of a set of data. An upper bar indicates the maximum value of a set of data, excluding outliers. A lower bar represents the minimum value of a set of data, excluding outliers. Outliers which are calculated as 1.5-3x the interquartile range, are plotted as individual circles.

Table 3.6 Post hoc test of mean value of driver torque.

|  | Manual | HG-normal | HG-strong |
| --- | --- | --- | --- |
| Manual | - | - | - |
| HG-normal | 0.000*** | - | - |
| HG-strong | 0.000*** | 0.000*** | - |

\*\*\*: $p < 0.001$

**3.1.4.4 Reaction to the critical event**

The average vehicle trajectory with standard deviation during lane change maneuver among all the participants was computed to show the relative position of the ego vehicle to the stopped lead vehicle, as shown in Figure 3.20. In the condition of HG-strong, the vehicle trajectory was near to the stopped lead vehicle compared to the conditions of Manual and HG-normal. HG-normal and Manual yielded a lower collision risk, as their vehicle trajectories were farther away from the stopped lead vehicle. This is reasonable as the HG-strong provided larger resistance torque when the driver performed the lane change maneuver. This result indicates a tendency that a higher degree of haptic guidance resulted in higher risks when a critical event occurred and the driver had to make a maneuver to overrule the assistance system.

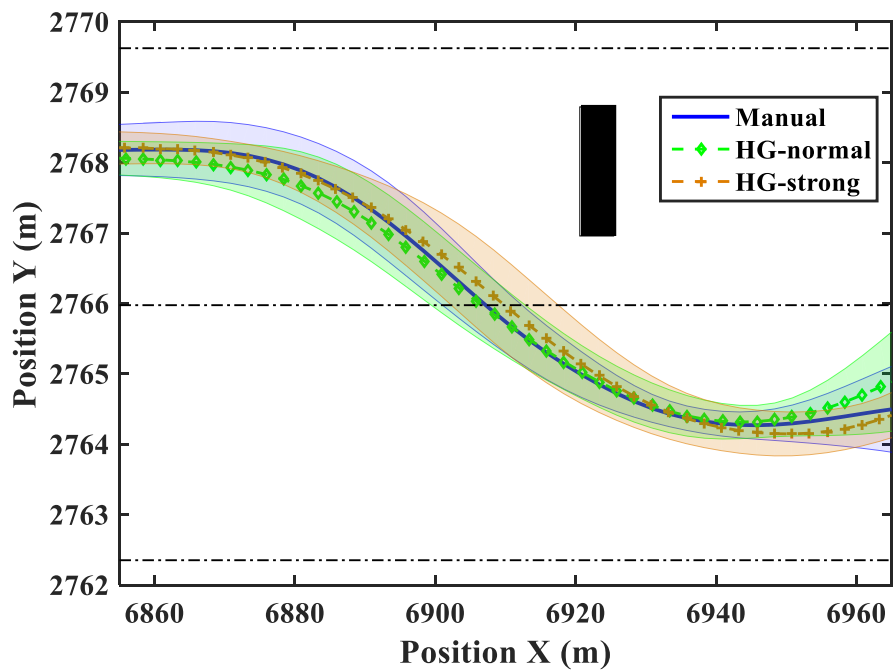

Figure 3.20 Average vehicle trajectory with standard deviation among all participants. Dot lines represent lane boundaries, and the black rectangle represents the stopped lead vehicle.

**3.1.4.5 Subjective evaluation**

Figure 3.21 shows the mean scores on NASA-TLX. From the result, first, it can be observed that there was a tendency that haptic guidance reduced the physical demand and effort, which corresponds to the result of reduction of measure driver torque (see Figure 3.19). On the other hand, we found that HG-strong significantly increased the feeling of frustration (insecure, discouraged, irritated and annoyed feelings). One explanation is that the participants could resist the strong haptic guidance torque sometimes when disagreed with the system, and the disagreement could result in a frustration feeling. It is quite interesting to see that the overall workload was lower in the condition of HG-normal compared to Manual and HG-strong. This can be explained by the observation that HG-normal reduced driver steering torque compared to Manual, and did not induce more frustration feelings than Manual; on the other hand, HG-strong was more effective in reducing driver steering torque, but HG-strong induced more frustration feelings to the driver. This result indicates that the great reduction in driver steering effort caused by strong haptic guidance had a downside effect, which was the feeling of frustration.

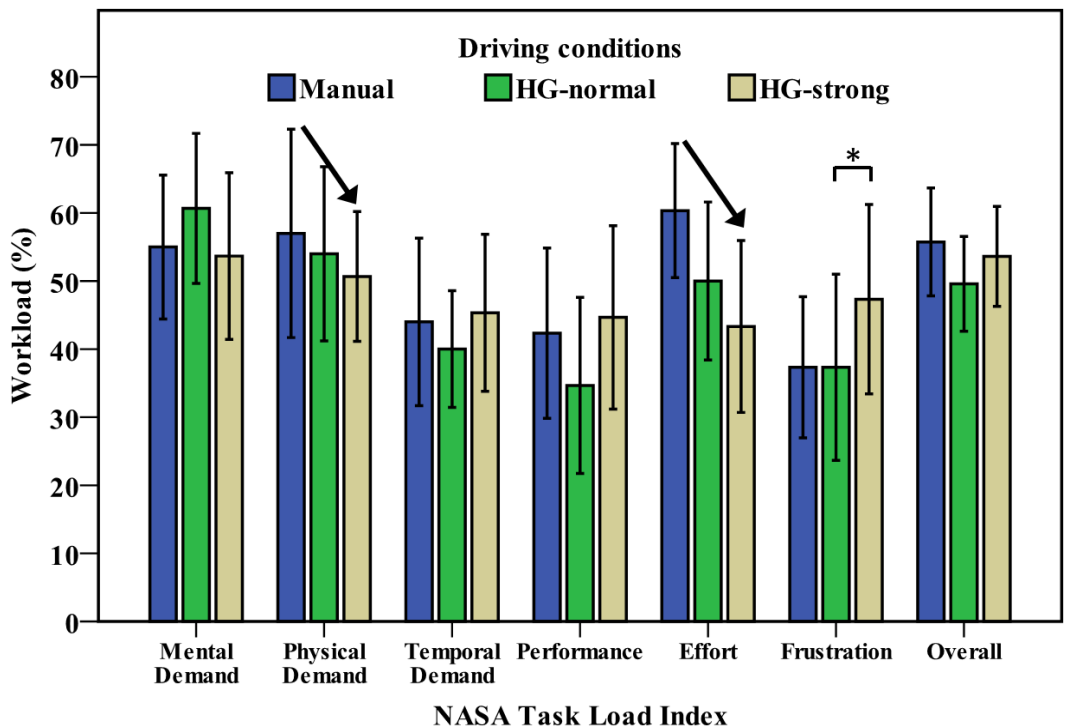

Figure 3.21 Mean scores on NASA-TLX (*p < 0.05). Error bars denote 0.95 confidence intervals.

### 3.1.5 Discussion

In this experiment, different degrees of haptic guidance were provided to the driver by using a haptic guidance steering system, and the driver behavior was investigated in the condition of normal visual information. The degree of haptic guidance plays a crucial role in driver–automation cooperation during a driving task. An optimal degree of haptic guidance is beneficial to achieving high-quality cooperation and mitigating conflicts between driver and automation [27, 78]. This study confirmed this, as HG-normal and HG-strong yielded a significant difference in the driver behavior in terms of lane following performance on straight lanes and driver workload. According to the results of NASA-TLX, driver-automation conflicts caused by strong haptic guidance cannot be ignored.

Although the mean value of driver torque was significantly lower in the condition of HG-strong than in HG-normal, the overall workload in NASA-TLX was somewhat higher. It suggests that the great reduction in overall steering effort caused by HG-strong had a downside effect, which was the participants' feeling of frustration. HG-normal also reduced the overall steering effort, but did not induce more frustration than Manual. The annoying or intrusive feeling caused by commercial lane keeping assistance systems is not rare when the system's control authority is high [92]. Some attempts on designing an adaptive haptic guidance steering system have been conducted to mitigate the annoying or intrusive feeling [38, 93].

When the intent of driver and the haptic guidance system did not match in the case of facing the critical event, the driver was capable of overruling the system to make a lane change by resisting the haptic guidance torque, which kept working for lane following purpose. Regarding the participants' reaction to the critical event, the lane change trajectories indicate that higher degree of haptic guidance would result in higher risks when the automation system failed.

In terms of driver–automation interaction, there is an opinion that drivers' experience on advanced driver assistance systems plays an important role in their behaviors [94]. Considering this, drivers' steering performance, gaze behavior, and subjective feelings may change when they have sufficient experience in driving with the haptic guidance steering system, and this will be an interesting future study.

### 3.1.6 Conclusion

The goal of this experiment is to test the proposed haptic guidance system by investigating its effect on lane following performance, driver workload, and especially on driver visual behavior. This section has detailed the experimental design, data analysis and results of the experiment. A driving simulator experiment was conducted, and the different degrees of haptic guidance were Manual, HG-normal, and HG-strong.

The results of lane following performance indicate that lane following performance increased or remained similar when driving with the proposed haptic guidance system, and meanwhile driver steering effort was reduced. However, the great reduction in driver steering effort caused by strong haptic guidance had a downside effect, which was the feeling of frustration. In addition, the implementation of haptic guidance system did not significantly influence the driver visual attention on the look-ahead point along curves, which indicates that the haptic guidance system was reliable and did not induce the increase of driver visual demand on the look-ahead point.

In the following sections, the effect of haptic guidance on driver behavior in the conditions of degraded visual information will be investigated.

## 3.2 Experimental study II: Effect of haptic guidance on driver behavior when driving with degraded visual information caused by visual occlusion from road ahead

### 3.2.1 Introduction

It has been suggested that the driver model for steering behavior should be based on both current and predictive visual feedback from the road ahead. Visual feedback from the near segment contributes to the position-in-lane task, whereas visual feedback from the far segment takes care of upcoming road curvatures [54]. One previous research has asked participants to steer along a central trajectory while visual feedback from road ahead is restricted by using one or two 1° viewing windows. It has been found that steering performance is equivalent to full visual feedback from the road ahead, when one viewing window displays the far region and another one displays the near region. When only one viewing window is available, the optimal region is midway between the near and far regions [51]. The accurate driving requires that both near and far segments of the road ahead are visible [51, 95].

Driver behavior would be influenced when visual feedback is defective due to environmental obstacles, such as fog [27, 57], rain, and other environmental factors, or during night driving [58]. Additionally, the location of driver support systems strongly impacts the driver's visual attention allocation [96, 97]. As the haptic guidance system is capable of providing reliable haptic guidance when driving with normal visual information, it is hypothesized that the decrement of driver lane following performance caused by degraded visual information could be compensated by the haptic information on the steering wheel provided by the haptic guidance system. Considering that the degraded visual information caused by visual occlusion from road ahead is various from case to case, this experiment addresses the ideal condition by providing the near, mid or far segment of road ahead in the driving scenario.

Therefore, the goal of this experiment is to evaluate the effectiveness of the haptic guidance system on improving lane following performance in the case of degraded visual information caused by visual occlusion from road ahead. In addition, different degrees of haptic guidance will be compared in order to find an appropriate degree in the different conditions of visual occlusion from road ahead. It is hypothesized that relatively higher degree of haptic guidance would be better when visual information is less reliable.

### 3.2.2 Experimental design

A driving simulator experiment was conducted with 12 participants. The experiment was approved by the Office for Life Science Research Ethics and Safety, the University of Tokyo (No. 14-113). The details of experimental design, including participants, apparatus, experiment conditions, experiment scenario and protocol, are presented below.

#### 3.2.2.1 Participants

Twelve healthy males were recruited to participate in the experiment. Their age ranged from 22 to 31 (mean = 24.3 years old, SD = 2.4). All of them had a valid Japanese driver's license for at least 1 year (mean = 4.3 years, SD = 2.6), and their average driving frequency was once per week. Each participant received monetary compensation for the involvement in the experiment.

### 3.2.2.2 Apparatus

The experiment was conducted in a high-fidelity driving simulator (Mitsubishi Precision Co., Ltd., Japan), which was the same one as used in experimental study I, as shown in Figure 3.1 in Section 3.1.2.

### 3.2.2.3 Experiment conditions

Each participant drove twelve trials designed by combining three degrees of haptic guidance (HG): HG none, HG normal, and HG strong, with four scenarios of visual feedback (VF): VF whole, VF near, VF mid, and VF far. The order of twelve trials was partially counterbalanced within the participants based on a Latin square design to mitigate the order effect, as shown in Table 3.7.

Table 3.7 Partially counterbalanced design of trial order by Latin squares.

| Participants | 1st trial | 2nd trial | 3rd trial | 4th trial | 5th trial | 6th trial | 7th trial | 8th trial | 9th trial | 10th trial | 11th trial | 12th trial |
|---|---|---|---|---|---|---|---|---|---|---|---|---|
| No.1 | 1 | 2 | 12 | 3 | 11 | 4 | 10 | 5 | 9 | 6 | 8 | 7 |
| No.2 | 2 | 3 | 1 | 4 | 12 | 5 | 11 | 6 | 10 | 7 | 9 | 8 |
| No.3 | 3 | 4 | 2 | 5 | 1 | 6 | 12 | 7 | 11 | 8 | 10 | 9 |
| No.4 | 4 | 5 | 3 | 6 | 2 | 7 | 1 | 8 | 12 | 9 | 11 | 10 |
| No.5 | 5 | 6 | 4 | 7 | 3 | 8 | 2 | 9 | 1 | 10 | 12 | 11 |
| No.6 | 6 | 7 | 5 | 8 | 4 | 9 | 3 | 10 | 2 | 11 | 1 | 12 |
| No.7 | 7 | 8 | 6 | 9 | 5 | 10 | 4 | 11 | 3 | 12 | 2 | 1 |
| No.8 | 8 | 9 | 7 | 10 | 6 | 11 | 5 | 12 | 4 | 1 | 3 | 2 |
| No.9 | 9 | 10 | 8 | 11 | 7 | 12 | 6 | 1 | 5 | 2 | 4 | 3 |
| No.10 | 10 | 11 | 9 | 12 | 8 | 1 | 7 | 2 | 6 | 3 | 5 | 4 |
| No.11 | 11 | 12 | 10 | 1 | 9 | 2 | 8 | 3 | 7 | 4 | 6 | 5 |
| No.12 | 12 | 1 | 11 | 2 | 10 | 3 | 9 | 4 | 8 | 5 | 7 | 6 |

1: VF whole & HG none; 2: VF near & HG normal; 3: VF mid & HG strong;
4: VF far & HG none; 5: VF whole & HG normal; 6: VF near & HG strong;
7: VF mid & HG none; 8: VF far & HG normal; 9: VF whole & HG strong;
10: VF near & HG none; 11: VF mid & HG normal; 12: VF far & HG strong.

Different degrees of haptic guidance were determined by the value of $K_1$ (see Equation 2.3 in Section 2.5.2): The haptic guidance with a double feedback gain (HG strong, $K_1 = 0.5$) was half of the gain set for automated lane tracking ($K_1 = 1.0$), and the haptic guidance with a normal feedback gain (HG normal, $K_1 = 0.25$) was half of the gain set for HG strong. The values of $a'_1, a'_2, a'_3,$ and $a'_4$ were decided as 3.2, 0.08, 20.0, and 0.50, respectively, through a trial-and-error method. The method was generally divided into two steps. The first step was to find rough solution of automated driving with aggressive tuning. The second step was to find desired solution with small tweaks in order to achieve good performance.

The road is segmented by distance and time, as shown in Figure 3.22. When a particular segment of the road was visible, the rest of the road remained black, as shown in Figure 3.23. In the experiment, there were four scenarios of visual feedback: whole, near, mid, and far segments. The visual feedback from the near segment provided road information from 0 s to 0.5 s ahead, mid segment from 0.75 s to 1.25 s ahead, and far segment from 1.5 s to infinity ahead [51, 52].

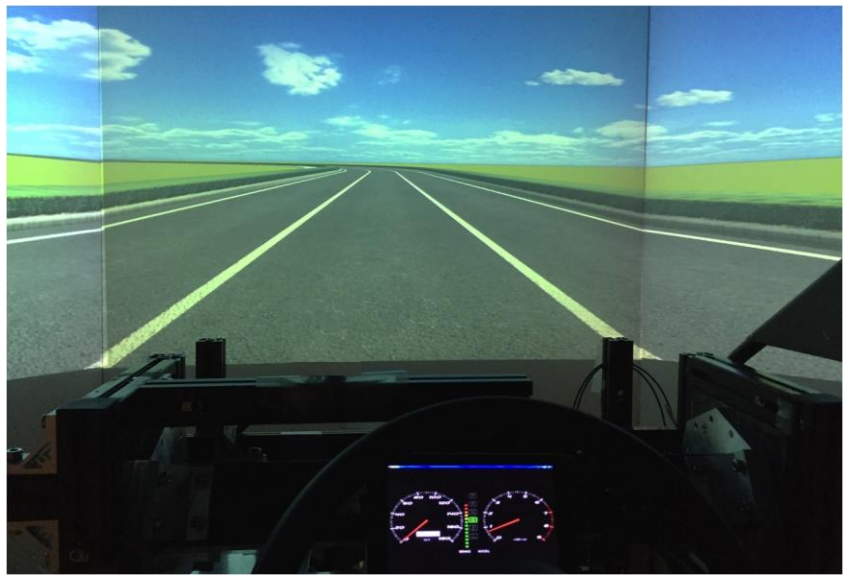

(a)

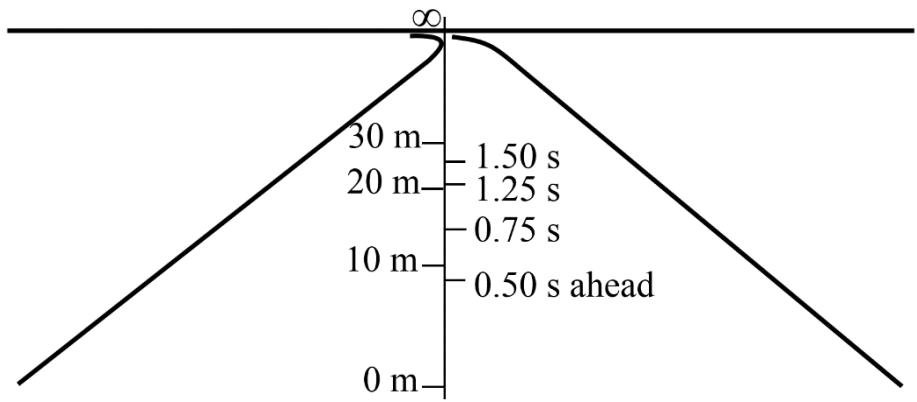

(b)

Figure 3.22 Visual feedback from road ahead. (a) Visual condition of whole segment presented in the driving simulator, (b) Sketch of segmented road. The conversion between the ahead distance and time is based on a driving speed of 60 km/h.

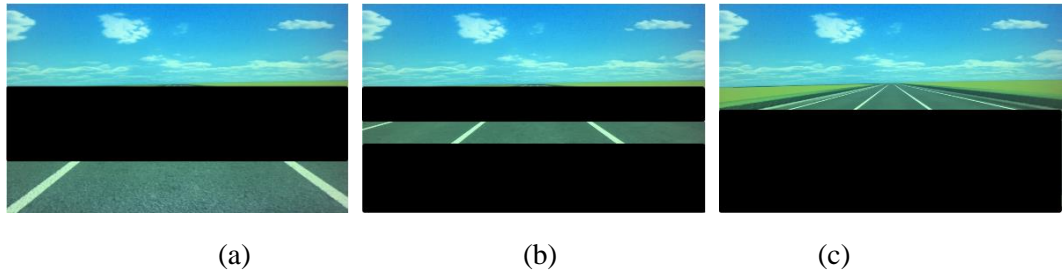

(a) (b) (c)

Figure 3.23 Driving scene with visual feedback from different segments of road ahead. (a) VF near, (b) VF mid, (c) VF far.

### 3.2.2.4 Experiment scenario

The driving environment was a three-lane rural road with a lane width of 3 m (see Figure 3.22 (a)), and the participants were required to drive in the middle lane. The driving speed was fixed at 60 km/h by a PID controller, and participants were asked not to operate the accelerator pedal and brake pedal during the drive unless it was necessary. The driving speed was fixed because the steering maneuver and speed control are related, and removing the speed variable makes the steering performance assessment less difficult.

The driving course consisted of twelve curves, and each curve was followed by one straight road of 100 or 200 m length, as shown in Figure 3.24. The length of each curve was 157 m, and the curve radius was one of 100, 150, and 200 m. The total length of the driving course was 3.9 km. The participants were unable to anticipate the upcoming path due to the various radiuses of the curves and various lengths of the straight roads.

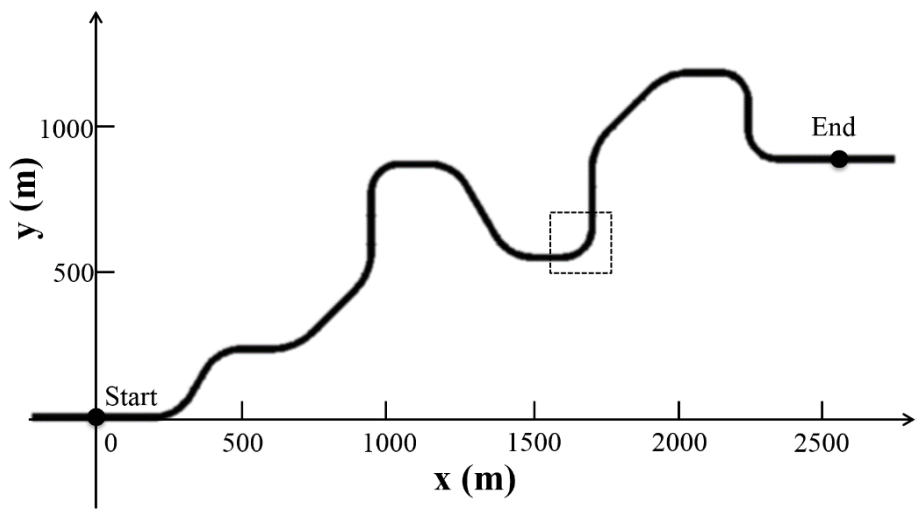

Figure 3.24 Layout of the driving course.

**3.2.2.5 Experiment protocol**

The participants were required to have a good night's sleep for 7 h before the experiment day. Caffeine and alcohol consumption was not allowed during the 12 h prior to the experiment.

On the experiment day, first, the participants signed a consent form, which explained the procedure of the experiment. The participants were naive to the purpose of the experiment. The participants then filled out a questionnaire about their personal demographic information and driving experience. After that, they got into the driving simulator and adjusted their seat to achieve a normal driving position. The participants were required to perform a practice task before the experimental session. In order to mitigate the learning effect, the practice task helped the participants familiarize themselves with the driving simulator, as well as the three degrees of haptic guidance and four scenarios of visual feedback. The participants were required to grab the steering wheel in a 'ten-and-two' position during the driving task. Additionally, they were asked to follow Japanese traffic rules and to drive in the center of the lane as accurately as possible. In the experimental session, each participant drove twelve trials designed by combining three degrees of haptic guidance: HG none, HG normal, and HG strong, with four scenarios of visual feedback: VF whole, VF near, VF mid, and VF far. The order of twelve trials was counterbalanced within the participants based on a Latin square design. After each trial, each participant was requested to fill out a questionnaire. The experiment took approximately 2 h per participant.

**3.2.3 Data analysis**

Initial driving performance data were collected in the driving simulator. Further signal processing steps were conducted in the MATLAB. In addition, statistical analysis on the post-processed data is explained.

**3.2.3.1 Lane following performance measurement**

In this experiment, the lane following performance on straight lanes was measured by standard deviation of lane position. The lane following performance on curves was measured by time-to-lane crossing and steering wheel reversal rate.

- Standard deviation of lane position (SDLP)

The algorithm to calculated SDLP on straight lanes has been presented in Section 3.1.3. Decreased SDLP indicates higher lane following performance or decreased lane departure risk.

- Time-to-lane crossing (TLC)

The algorithm to calculated TLC on curves has been presented in Section 3.1.3. A higher value of TLC indicates higher lane following performance or decreased lane departure risk.

- Steering wheel reversal rate (SWRR)

The algorithm to calculated SWRR on curves has been presented in Section 3.1.3. A lower value of SWRR indicates higher lane following performance or decreased lane departure risk.

**3.2.3.2 Starting moment of turning maneuver**

As shown in Figure 3.6 in Section 3.1.3, one curve negotiation process has been divided into five sections: the straight lane before a curve, approaching the curve, along the curve, leaving the curve, and the straight lane after the curve. In terms of approaching and leaving a curve, there is a starting moment of turning maneuver when the driver approaches a curve from a straight lane or leaving a curve to a straight lane. When approaching a curve, the relative starting moment of turning maneuver to geometrical turning point can reflect driver's steering response, and the same applies to leaving a curve. The geometrical turning point is the junction of a straight lane and a curve, as shown in Figure 3.6 in Section 3.1.3. The starting moment of turning maneuver can be either earlier or later than the geometrical turning point. When calculating the starting moment, a specific steering wheel angle should be determined as a threshold to indicate the starting of turning maneuver [98]. In this study, the threshold was set to 3°, as shown in Figure 3.25.

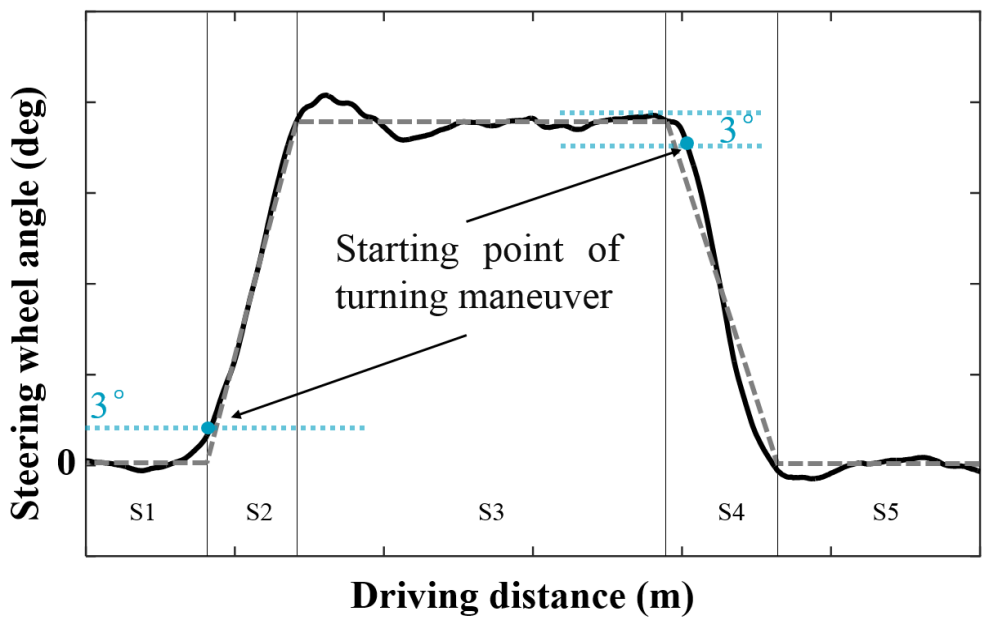

Figure 3.25 Steering wheel angle during curve negotiation. Blue dot represents the starting point of turning maneuver. The curve negotiation is divided into five sections by vertical grey lines: straight lane before the curve (S1), approaching the curve (S2), along the curve (S3), leaving the curve (S4), and straight lane after the curve (S5).

**3.2.3.3 Subjective evaluation**

A subjective evaluation is typically based on subjective feedback from the participants as opposed to objective, measured feedback that has been introduced above.

In this experiment, in addition to objective, measured lane following performance, the participants were asked to provide their self-evaluations on lane following performance in a self-evaluation questionnaire. The evaluation was on a scale of 1 – 5 (1, very low; 2, low; 3, medium; 4, high; 5, very high).

Additionally, after all the driving trials, the participants were asked to choose their preferred degree of haptic guidance for each visual feedback condition.

**3.2.3.4 Statistical analysis**

Repeated measures analysis of variance (ANOVA) has been explained in Section 3.1.3. To investigate the interaction effect between visual feedback and haptic guidance on driver behavior, the data were analyzed using two-way repeated measures ANOVA. In terms of two-way repeated measures ANOVA, "two-way" means there were two factors in this experiment: haptic guidance and visual feedback. Repeated measures means the same group of subjects serves in more than one treatment. In this experiment, the same group of subjects participated all the driving conditions.

The significance level was set to 0.05, and Mauchly's test was conducted prior to the repeated-measures ANOVA, like the steps used in the experimental study I. Post hoc Bonferroni pair-wise comparisons were used to identify the main effects, and significance level was also set to 0.05.

**3.2.4 Results**

Results are presented separately for lane following performance, starting moment of turning maneuver, and subjective evaluation. In terms of starting moment of turning maneuver, SWRR, and SDLP, the main effects were significant for visual feedback, haptic guidance, and their interaction. In terms of TLC and self-evaluation on lane keeping performance, the main effects were significant for visual feedback and haptic guidance, but not significant for their interaction.

**3.2.4.1 Lane following performance**

Lane following performance was measured by calculating SDLP on straight lanes, and calculating TLC and SWRR on curves. In addition, vehicle trajectory when approaching a curve and steering wheel angle when negotiating a curve were presented to give direct illustration of lane following performance in different driving conditions.

- Vehicle trajectory when approaching a curve

The relative position between vehicle trajectory and lane boundary indicates the lane following performance. Figure 3.26 shows the average vehicle trajectory of the participants when approaching a curve (see dotted line rectangle in Figure 3.24). With HG none, the vehicle trajectory for VF mid was similar to VF whole. For VF far, the drivers tended to start the turning maneuver earlier compared to VF whole, and the vehicle trajectory was close to the left lane boundary. By contrast, for VF near, the drivers started the turning maneuver later than for VF whole, and the vehicle trajectory was close to the right lane boundary. With HG strong, the vehicle trajectory was close to the centerline of the lane compared to HG none. HG normal was less effective compared to HG strong. It is indicated that the lane following performance when approaching the curve was worse in the condition of VF near compared to VF mid and VF whole; the performance decrement in the condition of VF near was compensated by the haptic guidance system, and strong haptic guidance was more effective than normal haptic guidance.

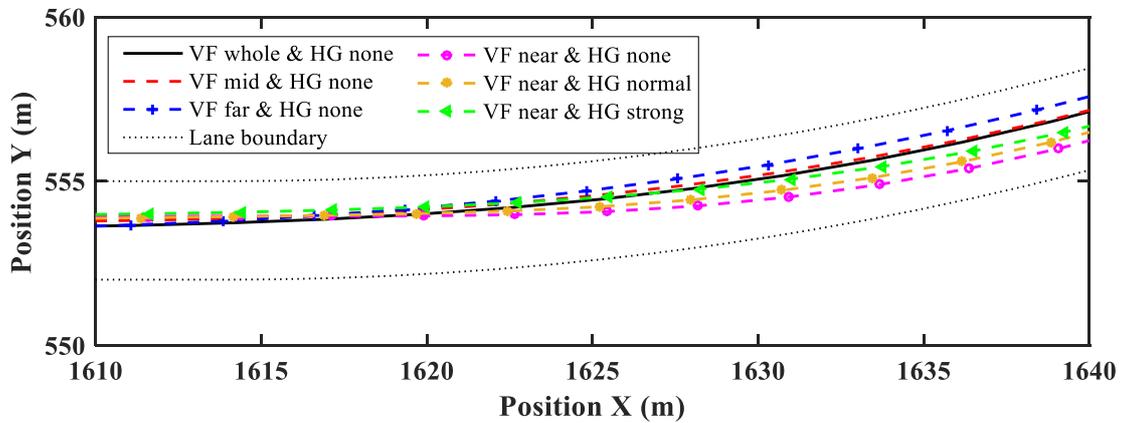

Figure 3.26 Average vehicle trajectory of twelve participants and lane boundaries when approaching a curve.

- Standard deviation of lane position (SDLP)

Figure 3.27 shows the SDLP when driving on straight lanes for each condition. There was a significant main effect of visual feedback ($F(6,66) = 32.561$, $p < 0.001$) as well as haptic guidance ($F(6,66) = 22.560$, $p < 0.001$) on SDLP. Additionally, there was a significant interaction effect between visual feedback and haptic guidance ($F(6,66) = 5.006$, $p < 0.001$). Table 3.8 presents the results of post hoc pair-wise comparisons on SDLP. In the conditions of VF whole, VF near, and VF far, the SDLP with HG strong was significantly lower than with HG normal and lower with HG strong than with HG none. The results indicate that lane following performance on straight lanes was worse without the haptic guidance in the conditions of VF near and VF far compared to VF whole and VF mid. Moreover, the performance decrement in the conditions of VF near and VF far was significantly compensated by the implementation of haptic guidance, and strong haptic guidance was more effective than normal haptic guidance.

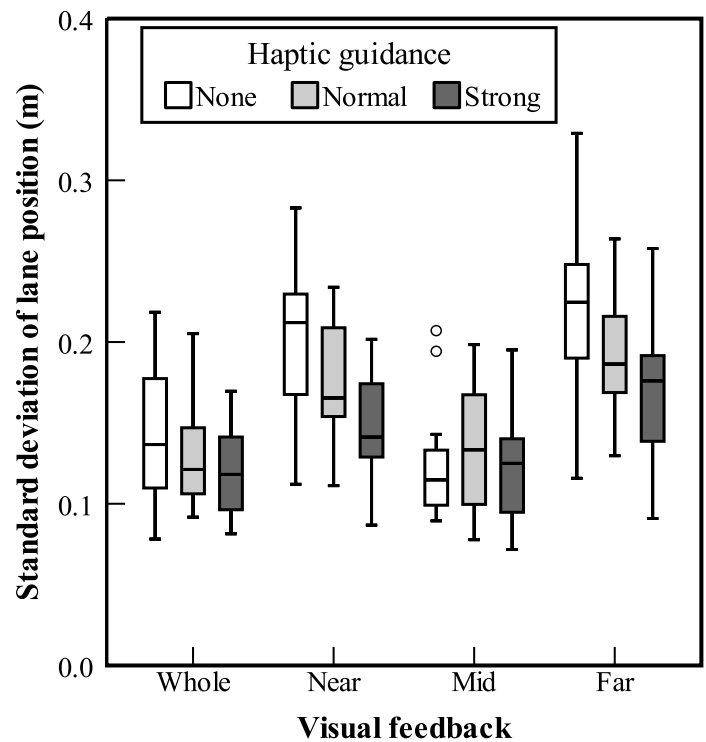

Figure 3.27 Standard deviation of lane position on straight lanes. A rectangle represents the middle 50% of a set of data. A horizontal line drawn through a rectangle corresponds to the median value of a set of data. An upper bar indicates the maximum value of a set of data, excluding outliers. A lower bar represents the minimum value of a set of data, excluding outliers. Outliers which are calculated as 1.5-3x the interquartile range, are plotted as individual circles.

Table 3.8 Post hoc test of standard deviation of lane position.
(a) Difference between VF

|  |  | Whole | Near | Mid | Far |
|---|---|---|---|---|---|
| None | Whole | - | - | - | - |
|  | Near | 0.000*** | - | - | - |
|  | Mid | 1.000 | 0.001** | - | - |
|  | Far | 0.000*** | 0.534 | 0.000*** | - |
| Normal | Whole | - | - | - | - |
|  | Near | 0.002** | - | - | - |
|  | Mid | 1.000 | 0.020* | - | - |
|  | Far | 0.000*** | 0.151 | 0.000*** | - |
| Strong | Whole | - | - | - | - |
|  | Near | 0.020* | - | - | - |
|  | Mid | 1.000 | 0.153 | - | - |
|  | Far | 0.000*** | 0.235 | 0.006** | - |

(b) Difference between HG

| | | None | Normal | Strong |
|---|---|---|---|---|
| Whole | None | - | - | - |
| | Normal | 0.256 | - | - |
| | Strong | 0.020* | 0.020* | - |
| Near | None | - | - | - |
| | Normal | 0.055 | - | - |
| | Strong | 0.000*** | 0.021* | - |
| Mid | None | - | - | - |
| | Normal | 0.777 | - | - |
| | Strong | 1.000 | 0.806 | - |
| Far | None | - | - | - |
| | Normal | 0.102 | - | - |
| | Strong | 0.001** | 0.005** | - |

***: $p < 0.001$, **: $p < 0.01$, *: $p < 0.05$.

- Time-to-lane crossing (TLC)

Figure 3.28 shows the TLC when driving along curves for each condition. There was a significant main effect of visual feedback ($F(6,66) = 17.489$, $p < 0.001$) as well as haptic guidance ($F(6,66) = 10.878$, $p = 0.001$) on TLC. Additionally, there was no significant interaction effect between visual feedback and haptic guidance, but a tendency existed ($F(6,66) = 2.006$, $p = 0.077$). Table 3.9 presents the results of post hoc pair-wise comparisons on TLC. From pair-wise comparisons, the TLC for VF whole was significantly higher than for VF near, higher for VF whole than for VF far, and higher for VF mid than for VF near. Moreover, the TLC with HG strong was significantly higher than with HG normal and higher with HG strong than with HG none. The results indicate that lane following performance along curves was worse without the haptic guidance in the conditions of VF near and VF far compared to VF whole and VF mid. The performance decrement in the condition of VF near was significantly compensated by the implementation of haptic guidance, and strong haptic guidance was more effective than normal haptic guidance. However, no significant improvement by haptic guidance was found in the condition of VF far.

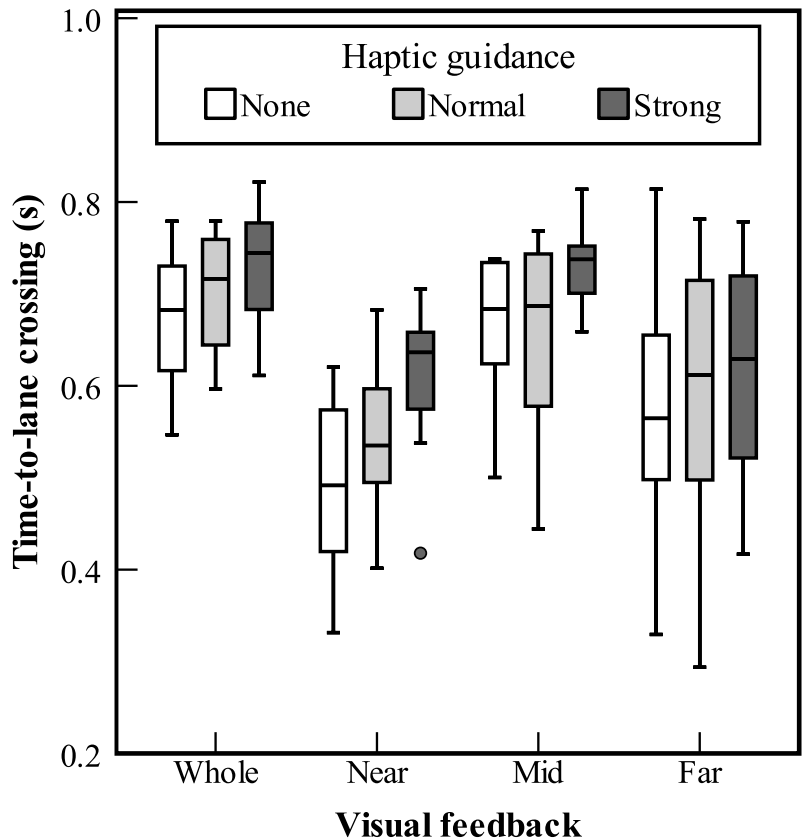

Figure 3.28 Time-to-lane crossing along curves. A rectangle represents the middle 50% of a set of data. A horizontal line drawn through a rectangle corresponds to the median value of a set of data. An upper bar indicates the maximum value of a set of data, excluding outliers. A lower bar represents the minimum value of a set of data, excluding outliers. Outliers which are calculated as 1.5-3x the interquartile range, are plotted as individual circles.

Table 3.9 Post hoc test of time-to-lane crossing.
(a) Difference between VF

|  | Whole | Near | Mid | Far |
| --- | --- | --- | --- | --- |
| Whole | - | - | - | - |
| Near | 0.000*** | - | - | - |
| Mid | 1.000 | 0.002** | - | - |
| Far | 0.012* | 0.758 | 0.085 | - |

(b) Difference between HG

|  | None | Normal | Strong |
| --- | --- | --- | --- |
| None | - | - | - |
| Normal | 0.194 | - | - |
| Strong | 0.007** | 0.047* | - |

***: $p < 0.001$, **: $p < 0.01$, *: $p < 0.05$.

- Steering wheel angle along a curve negotiation

Figure 3.29 shows the average steering wheel angle of the participants during a curve negotiation in different driving conditions. With HG none, the steering behavior for VF mid was similar to VF whole. For VF far, the drivers tended to smoothly start and finish their turning maneuver when approaching and leaving the curve. By contrast, for VF near, the drivers sharply turned the steering wheel due to lack of predictive visual feedback, and there was an obvious peak of steering wheel angle of more than 50°. With HG strong, the peak of steering wheel angle was lower, and the turning maneuver was smoother compared to HG none. HG normal was less effective compared to HG strong. This result is in accordance with the lane following performance measured by TLC, as the lane following performance along curves was worse without the haptic guidance in the conditions of VF near compared to VF whole and VF mid. The performance decrement in the condition of VF near was compensated by the implementation of haptic guidance, and strong haptic guidance was more effective than normal haptic guidance.

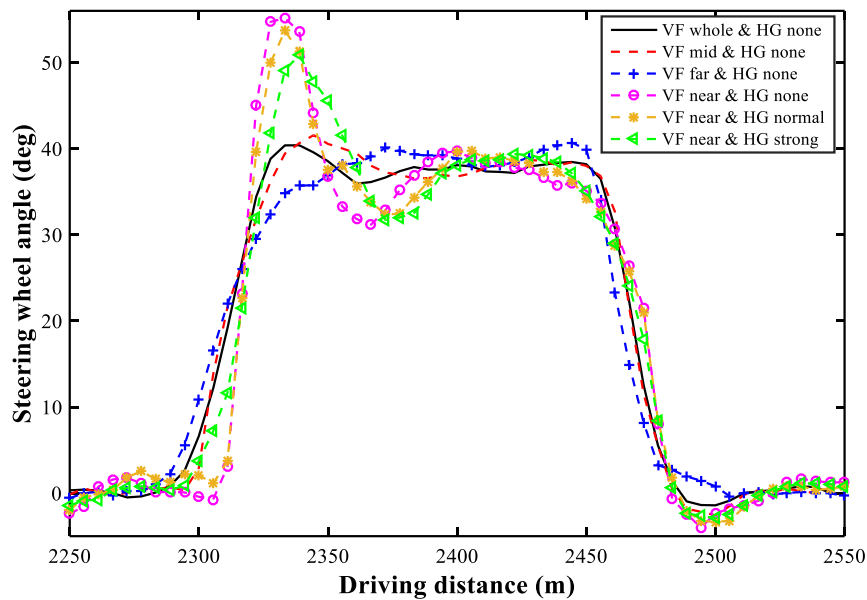

Figure 3.29 Average steering wheel angle of twelve participants when negotiating a curve.

- Steering wheel reversal rate (SWRR)

Figure 3.30 shows the SWRR when driving on curves for each condition. There was a significant main effect of visual feedback ($F(6,66) = 71.872$, $p < 0.001$) as well as of haptic guidance ($F(6,66) = 6.096$, $p = 0.008$) on SWRR. Additionally, there was a significant interaction effect between visual feedback and haptic guidance ($F(6,66) = 4.619$, $p = 0.001$). Table 3.10 presents the results of post hoc pair-wise comparisons on SWRR. For VF near, the SWRR with HG strong was significantly lower than that with HG none. This result is in accordance with the steering wheel angle along a curve negotiation as shown in Figure 3.29.

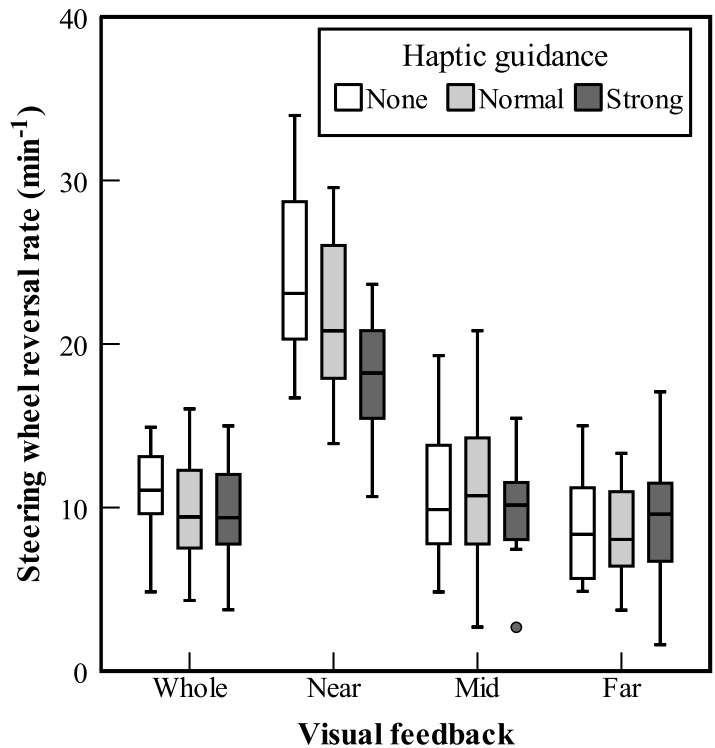

Figure 3.30 Steering wheel reversal rate along curves. A rectangle represents the middle 50% of a set of data. A horizontal line drawn through a rectangle corresponds to the median value of a set of data. An upper bar indicates the maximum value of a set of data, excluding outliers. A lower bar represents the minimum value of a set of data, excluding outliers. Outliers which are calculated as 1.5-3x the interquartile range, are plotted as individual circles.

Table 3.10 Post hoc test of steering wheel reversal rate.
(a) Difference between VF

|   |   | Whole | Near | Mid | Far |
|---|---|---|---|---|---|
| None | Whole | - | - | - | - |
|  | Near | 0.000*** | - | - | - |
|  | Mid | 1.000 | 0.000*** | - | - |
|  | Far | 0.074 | 0.000*** | 0.976 | - |
| Normal | Whole | - | - | - | - |
|  | Near | 0.000*** | - | - | - |
|  | Mid | 1.000 | 0.001** | - | - |
|  | Far | 1.000 | 0.000*** | 0.384 | - |
| Strong | Whole | - | - | - | - |
|  | Near | 0.000*** | - | - | - |
|  | Mid | 1.000 | 0.000*** | - | - |
|  | Far | 1.000 | 0.000*** | 1.000 | - |

(b) Difference between HG

|  |  | None | Normal | Strong |
|---|---|---|---|---|
| Whole | None | - | - | - |
|  | Normal | 0.695 | - | - |
|  | Strong | 0.295 | 1.000 | - |
| Near | None | - | - | - |
|  | Normal | 0.125 | - | - |
|  | Strong | 0.001** | 0.080 | - |
| Mid | None | - | - | - |
|  | Normal | 1.000 | - | - |
|  | Strong | 1.000 | 0.850 | - |
| Far | None | - | - | - |
|  | Normal | 1.000 | - | - |
|  | Strong | 1.000 | 1.000 | - |

***: $p < 0.001$, **: $p < 0.01$, *: $p < 0.05$.

### 3.2.4.2 Starting moment of turning maneuver

There is a starting moment of turning maneuver when the driver approaches a curve from a straight lane or leaving a curve to a straight lane. When approaching a curve, the relative starting moment of turning maneuver to geometrical turning point can reflect driver's steering response, and the same applies to leaving a curve.

- Approaching curves

Figure 3.31 shows the starting moment of turning maneuver when approaching curves for each driving condition. There was a significant main effect of visual feedback ($F(6,66) = 257.65$, $p < 0.001$) as well as of haptic guidance ($F(6,66) = 27.131$, $p < 0.001$). Additionally, there was a significant interaction effect between visual feedback and haptic guidance ($F(6,66) = 12.658$, $p < 0.001$). Table 3.11 presents the results of post hoc pair-wise comparisons. For the VF near, the starting moment of turning maneuver occurred significantly earlier with HG strong than with HG normal, earlier with HG strong than with HG none, and earlier with HG normal than with HG none. Moreover, for the VF far, the starting moment of turning maneuver occurred significantly later with HG strong than with HG none. This result indicates that the starting moment of turning maneuver when approaching curves was significantly affected by the interaction between haptic guidance and visual feedback. In the condition of VF near, the effect of haptic guidance on assisting drivers' turning maneuver was significant and strong haptic guidance was more effective than normal haptic guidance. It suggests that the driver tended to integrate the feedback of visual and haptic information to achieve better performance.

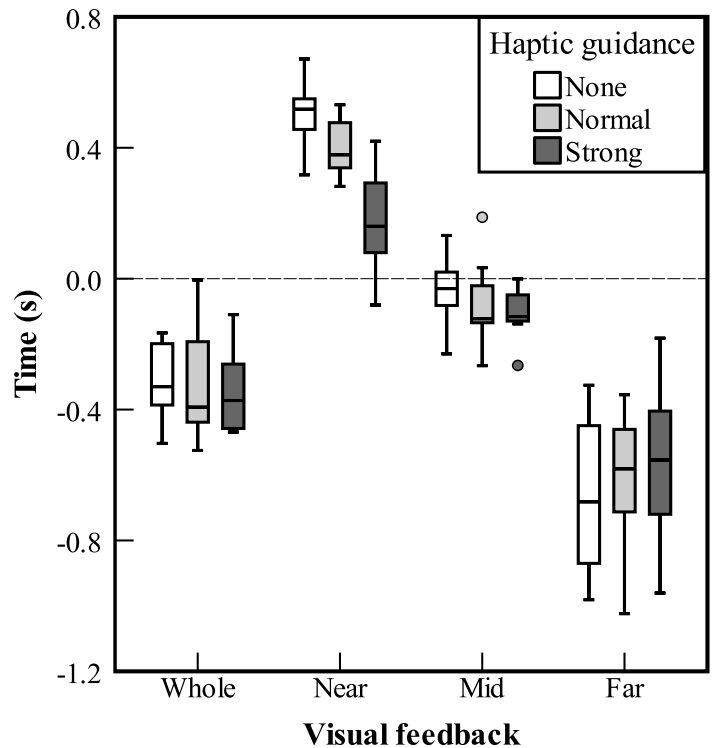

Figure 3.31 Starting moment of turning maneuver when approaching curves. The reference point is the junction of straight lane and curve. A rectangle represents the middle 50% of a set of data. A horizontal line drawn through a rectangle corresponds to the median value of a set of data. An upper bar indicates the maximum value of a set of data, excluding outliers. A lower bar represents the minimum value of a set of data, excluding outliers. Mild outliers are indicated by circles.

Table 3.11 Post hoc test of starting moment of turning maneuver when approaching curves.
(a) Difference between VF

|  |  | Whole | Near | Mid | Far |
|---|---|---|---|---|---|
| None | Whole | - | - | - | - |
|  | Near | 0.000*** | - | - | - |
|  | Mid | 0.000*** | 0.000*** | - | - |
|  | Far | 0.000*** | 0.000*** | 0.000*** | - |
| Normal | Whole | - | - | - | - |
|  | Near | 0.000*** | - | - | - |
|  | Mid | 0.003** | 0.000*** | - | - |
|  | Far | 0.000*** | 0.000*** | 0.000*** | - |
| Strong | Whole | - | - | - | - |
|  | Near | 0.000*** | - | - | - |
|  | Mid | 0.000*** | 0.000*** | - | - |
|  | Far | 0.004** | 0.000*** | 0.000*** | - |

(b) Difference between HG

|  |  | None | Normal | Strong |
|---|---|---|---|---|
| Whole | None | - | - | - |
|  | Normal | 1.000 | - | - |
|  | Strong | 0.840 | 1.000 | - |
| Near | None | - | - | - |
|  | Normal | 0.014 | - | - |
|  | Strong | 0.000*** | 0.000*** | - |
| Mid | None | - | - | - |
|  | Normal | 0.805 | - | - |
|  | Strong | 0.138 | 1.000 | - |
| Far | None | - | - | - |
|  | Normal | 0.426 | - | - |
|  | Strong | 0.018* | 0.797 | - |

***: $p < 0.001$, **: $p < 0.01$, *: $p < 0.05$.

- Leaving curves

Figure 3.32 shows the starting moment of turning maneuver when leaving curves for each condition. There was a significant main effect of visual feedback ($F(6,66) = 25.056$, $p < 0.001$) as well as of haptic guidance ($F(6,66) = 19.085$, $p < 0.001$). Additionally, there was a significant interaction effect between visual feedback and haptic guidance ($F(6,66) = 5.596$, $p < 0.001$). Table 3.12 presents the results of post hoc pair-wise comparisons on starting moment of turning maneuver when leaving curves. For the VF near, the starting moment of turning maneuver occurred significantly earlier with HG strong than with HG normal, and earlier with HG strong than with HG none.

This result indicates that the starting moment of turning maneuver when leaving curves was significantly affected by the interaction between haptic guidance and visual feedback. In the condition of VF near, the effect of strong haptic guidance on assisting drivers' turning maneuver was significant. For the other three conditions of visual feedback, there was no significant effect of haptic guidance on turning maneuver. In the condition of VF near, the starting moment of turning maneuver was late compared to the geometrical turning point when approaching curves; however, it is interesting that the starting moment of turning maneuver was not quite late compared to the geometrical turning point when leaving curves, and the starting moment of turning maneuver was even earlier than the geometrical turning point when driving with strong haptic guidance. One explanation would be that when approaching curves the drivers were not aware of the appropriate steering angle, but when leaving curves they could start the turning maneuver as soon as they felt the strong guidance.

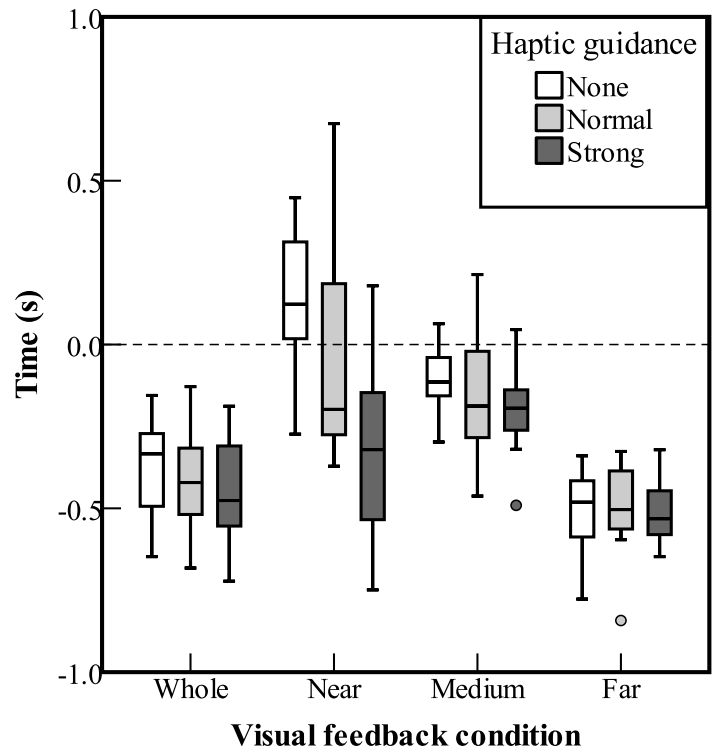

Figure 3.32 Starting moment of turning maneuver when leaving curves. The reference point is the junction of straight lane and curve. A rectangle represents the middle 50% of a set of data. A horizontal line drawn through a rectangle corresponds to the median value of a set of data. An upper bar indicates the maximum value of a set of data, excluding outliers. A lower bar represents the minimum value of a set of data, excluding outliers. Mild outliers are indicated by circles.

Table 3.12 Post hoc test of starting moment of turning maneuver when leaving curves.
(a) Difference between VF

|  |  | Whole | Near | Mid | Far |
|---|---|---|---|---|---|
| None | Whole | - | - | - | - |
|  | Near | 0.002** | - | - | - |
|  | Mid | 0.001** | 0.040* | - | - |
|  | Far | 0.058 | 0.000*** | 0.000*** | - |
| Normal | Whole | - | - | - | - |
|  | Near | 0.009** | - | - | - |
|  | Mid | 0.003** | 1.000 | - | - |
|  | Far | 0.910 | 0.005** | 0.001** | - |
| Strong | Whole | - | - | - | - |
|  | Near | 1.000 | - | - | - |
|  | Mid | 0.068 | 1.000 | - | - |
|  | Far | 1.000 | 0.387 | 0.000*** | - |

(b) Difference between HG

|  |  | None | Normal | Strong |
|---|---|---|---|---|
| Whole | None | - | - | - |
|  | Normal | 1.000 | - | - |
|  | Strong | 0.959 | 1.000 | - |
| Near | None | - | - | - |
|  | Normal | 0.288 | - | - |
|  | Strong | 0.000*** | 0.003** | - |
| Mid | None | - | - | - |
|  | Normal | 1.000 | - | - |
|  | Strong | 0.174 | 1.000 | - |
| Far | None | - | - | - |
|  | Normal | 1.000 | - | - |
|  | Strong | 1.000 | 1.000 | - |

***: $p < 0.001$, **: $p < 0.01$, *: $p < 0.05$.

### 3.2.4.3 Subjective evaluation

- Self-evaluation on lane following performance

Figure 3.33 shows the self-evaluation score on lane keeping performance for each condition. There was a significant main effect of visual feedback ($F(6,66) = 32.774$, $p < 0.001$) as well as haptic guidance ($F(6,66) = 3.811$, $p = 0.038$) on evaluation score. Additionally, there was no significant interaction effect between visual feedback and haptic guidance ($F(6,66) = 0.635$, $p = 0.701$). Table 3.13 presents the results of post hoc pair-wise comparisons. From pair-wise comparisons, it can be observed that the self-evaluation score on lane keeping performance for VF whole was significantly higher than for VF near, higher for VF whole than for VF mid, higher for VF whole than for VF far, higher for VF mid than for VF near, and higher for VF far than for VF near. In general, the result of self-evaluation on lane following performance is in accordance with the measured lane following performance by SDLP on straight lanes and TLC along curves, as the lane following performance was best in the condition of VF whole, worst in the condition of VF near, and somewhere between them in the condition of VF far and VF mid. In the conditions of VF whole, VF near, and VF mid, there was a tendency that haptic guidance improved the lane following performance and strong haptic guidance was more effective than normal haptic guidance.

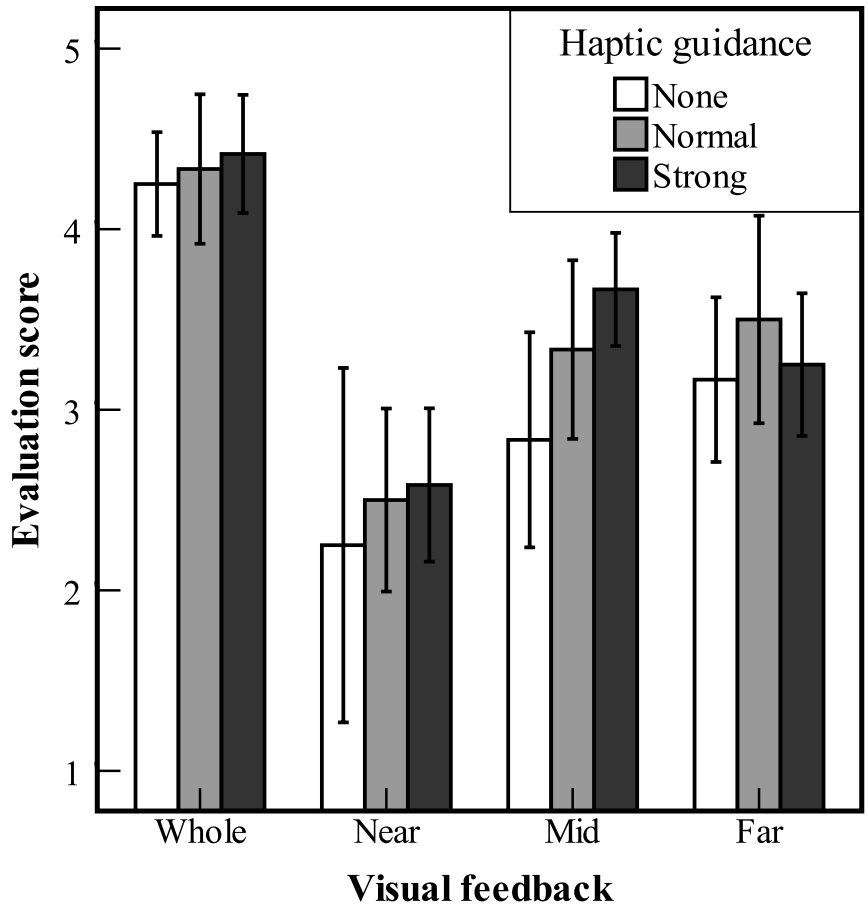

Figure 3.33 Self-evaluation on lane keeping performance. Error bars denote 0.95 confidence intervals.

Table 3.13 Post hoc test of self-evaluation on lane keeping performance.
(a) Difference between VF

|       | Whole    | Near     | Mid   | Far |
|-------|----------|----------|-------|-----|
| Whole | -        | -        | -     | -   |
| Near  | 0.000*** | -        | -     | -   |
| Mid   | 0.000*** | 0.003**  | -     | -   |
| Far   | 0.001**  | 0.020*   | 1.000 | -   |

(b) Difference between HG

|        | None  | Normal | Strong |
|--------|-------|--------|--------|
| None   | -     | -      | -      |
| Normal | 0.300 | -      | -      |
| Strong | 0.091 | 1.000  | -      |

***: $p < 0.001$, **: $p < 0.01$, *: $p < 0.05$.

- Subjective preference

According to the subjective evaluation, the participants had different preferences on the degree of haptic guidance. The result in Figure 3.34 shows that in the condition of VF near, most participants prefer strong haptic guidance, and the other two participants prefer normal haptic guidance. For the other three conditions of visual feedback, most participants prefer normal haptic guidance. Under the condition of visual feedback from whole segment, there were three participants who prefer to drive without haptic guidance. This result indicates that when visual feedback was reliable, participants preferred lower degree of haptic guidance, as higher degree of haptic guidance could induce a frustration feeling to the driver which has been discussed in Sections 3.1.4 and 3.1.5.

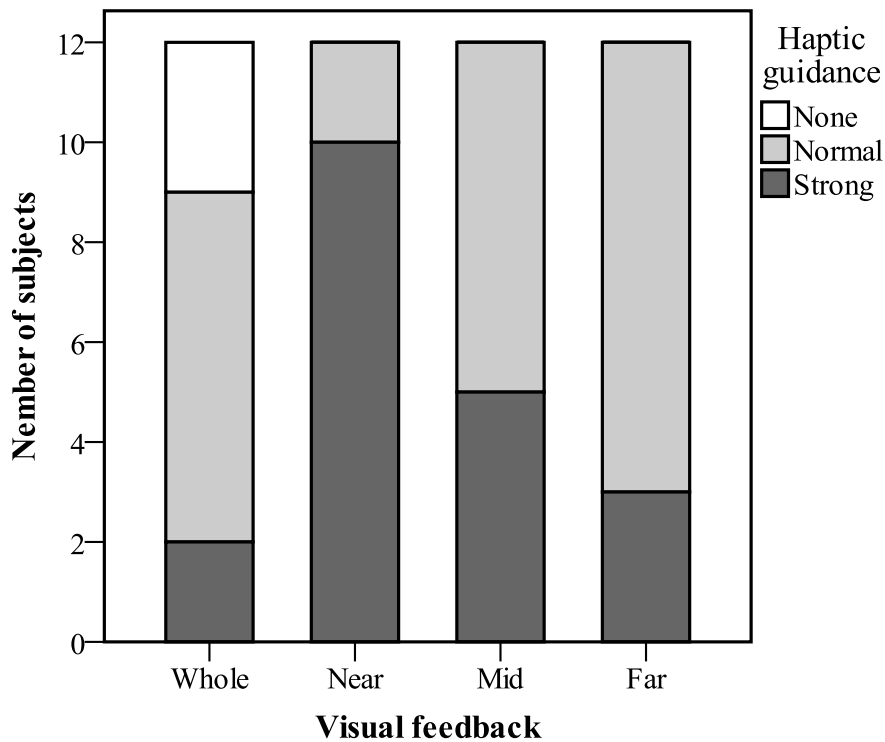

Figure 3.34 Subjective preference on the degree of haptic guidance in different visual feedback conditions.

### 3.2.5 Discussion

The driver steering behavior was significantly influenced by degraded visual information caused by the visual occlusion from road ahead. For VF near, the starting moment of turning maneuver occurred later than the geometrical turning point due to the lack of advance road information. In addition, SWRR was higher than the other three VF conditions, which suggests that the steering behavior for VF near was unstable. Our findings are in accordance with the viewpoint that available tangent point, a point on the lane edge of the inside of the curve where the line of sight is tangential to the lane edge, would lead to a smoother and more stable steering maneuver [98]. According to the results of SDLP and TLC, the lane following performance for VF mid was similar to VF whole, where both near and far segments of the road were partially provided [51]. Compared to VF whole and VF mid, the lane departure risk was higher for VF near and VF far. Visual feedback from the near segment helps the drivers attain their vehicle position, and the far segment leads to the prediction of upcoming roads [54]. Our observations confirm that having either a near or far region alone is not sufficient for accurate steering, and an optimal region should be midway between the near and far regions [51]. In addition, we observed that the starting moment of turning maneuver occurred significantly earlier in the condition of VF whole compared to VF mid. It suggests that VF mid influenced the drivers' steering strategies, although the lane following performance in the condition of VF mid was as good as it in the condition of VF whole. In another previous research on driving with the degraded visual feedback, the authors used an opacity mask (20%, 40%, 60%, 80%, or 100% opaque) to cover either the top half (far region of a road ahead as well as sky) or bottom half of the display (near region of a road ahead). They found a large systematic increase in steering wheel velocity from a 60% upper-field mask, to an 80% upper-field mask, to a 100% upper-field mask, whereas the same increases in opacity of a lower-field mask did not markedly impair steering wheel velocity [99]. In our study, we further observed that for HG none, SWRR was significantly higher in the condition of VF near than that of VF whole, whereas no significant difference in SWRR was found between VF far and VF whole. Additionally, we found that HG strong significantly reduced SWRR in the condition of VF near.

In terms of the influence of haptic guidance on driver behavior, HG strong resulted in similar or better measured driving performance compared to HG normal in the condition of VF whole. The similar results were found in [100], in which continuous haptic guidance with double feedback gain (high-level of haptic guidance) resulted in the best lane keeping performance. However, according to [27], the best driving performance was achieved with relatively low-level of haptic guidance. It should be noticed that the way of designing the haptic guidance steering system (haptic guidance model shown in Figure 2.9 in Section 2.5.2) could be crucial for the driving performance, and may lead to the above different results. Regarding the degraded visual feedback conditions, our observations confirm that the driving performance decrement due to unavailable visual feedback from far segment can be compensated by the implementation of haptic guidance [27], as SDLP and SWRR were significantly reduced by HG strong. We further observed that the starting moment of turning maneuver occurred closer to the geometrical turning point owing to the use of haptic guidance in the condition of VF near. For VF far, compared to HG none, the starting moment of turning maneuver occurred later, resulting from HG strong. It has been reported that human perceptual judgments rely upon the flexible weighted combination of information with weaker or less reliable signals being down-weighted [63, 101]. By our observations on the starting moment of turning maneuver, it was validated that the drivers relied upon the haptic guidance by actively following it, when the visual feedback was degraded. It is also important to note that for VF near, the starting moment occurred later than the geometrical point even with HG strong. It indicates that the HG compensated for the lack of VF instead of replacing it [25]. The drivers still needed visual feedback to estimate the radius of the upcoming curves to decide the steering angle. The results of SWRR on curves indicate that a strong haptic guidance helped stabilize the driver steering movement for VF near. The similar results have been found by measuring the torque reversal rate of driver under low visibility conditions [57], as steering control activity is lower for driving with the haptic guidance system compared to manual driving. In terms of lane following performance on straight lanes, we found that HG strong had a significant effect on decreasing SDLP for VF whole, VF near, and VF far, whereas HG normal did not have any significant effect. It was unexpected that no effect of haptic guidance on SDLP was observed for VF mid, and this observation needs to be further addressed in the future study. The results of TLC indicate that lane departure risk on curves was significantly decreased by haptic guidance, and HG strong was more effective than HG normal.

Based on the analysis of the self-evaluation, VF whole had the best lane following performance while VF near had the worst. These corresponded well with the results of SDLP and TLC. Additionally, the lane following performance by self-evaluation for VF far was similar to that of VF mid, and was better than that of VF near. However, the results of SDLP and TLC show that the lane following performance of VF far was significantly lower than that of VF mid. It seems that the participants were not able to accurately evaluate their lane following performance when VF from only the far segment was provided. In addition, the subjective preference results have shown that most participants preferred HG strong for VF near. However, HG strong was not always the preferred choice, as most participants preferred HG normal in the conditions of VF whole, VF mid and VF far. This is not surprising because the participants could resist the strong haptic guidance sometimes when in disagreement with the system's target trajectory [100]. In addition, this disagreement could result in an intrusive feeling to the drivers [38], although the strong haptic guidance improved the driving performance including the increased TLC and decreased SDLP. It is assumed that the disagreements could be less for long-term driving with HG strong.

In terms of drivers' visual behavior, it should be noticed that the way drivers control their gaze directions would significantly influence their steering behavior when visual feedback is degraded [101]. Therefore, it would be an interesting topic for future study, to investigate drivers' visual behavior under degraded visual feedback conditions while implementing different degrees of haptic guidance.

### 3.2.6 Conclusion

The goal of this experiment is to evaluate the effectiveness of haptic guidance system on improving lane following performance in the condition of degraded visual information caused by visual occlusion from road ahead, and also to investigate the effect of different degree of haptic guidance on driver behavior. A driving simulator experiment was conducted, and experimental conditions were designed by combining three degrees of haptic guidance, namely HG none, HG normal, and HG strong, with four scenarios of visual feedback, namely VF whole, VF near, VF mid, and VF far. The influence of haptic guidance associated with degrade visual feedback on the driver behavior was investigated based on lane following performance and subjective evaluation.

The obtained results show that the decrement of lane following performance caused by visual occlusion from far segment of road ahead could be compensated by haptic feedback, and strong haptic guidance was more effective than normal haptic guidance. The decrement of lane following performance on straight lanes caused by visual occlusion from near segment of road ahead could be compensated by haptic feedback, but on curves the effectiveness was not significant. Moreover, the result of starting moment of turning maneuver proved evidence that the driver tended to integrate the feedback of visual and haptic information to achieve a better driving performance. Subjective evaluation on lane following performance showed a similar tendency with the objectively measured lane following performance of SDLP and TLC. Subjective preference result shows that most participants preferred strong haptic guidance in the condition of VF near. However, normal haptic guidance was the more preferred choice in the other three conditions, where visual feedback was more reliable compared to VF near.

Therefore, this experimental study suggests that the decrement of driving performance caused by visual occlusion from road ahead, especially in the condition of VF near, could be compensated by haptic information provided by the haptic guidance system. The next experimental study will address another kind of degraded visual information that is caused by declined visual attention under fatigue driving, and will evaluate the effectiveness of the haptic guidance system on improving driver lane following performance.

## 3.3 Experimental study III: Effect of haptic guidance on driver behavior when driving with degraded visual information caused by declined visual attention under fatigue

### 3.3.1 Introduction

Experimental study II addressed the effect of haptic guidance on driver behavior when driving with degraded visual information caused by visual occlusion from road ahead, and the scenario corresponds to the characteristic of driver visual model with two look-ahead points (a near point and a far point), as shown in Figure 2.4 in Section 2.3.2. Another crucial parameter in the visual system of driver model is the processing time delay. A longer processing time delay would cause the feedback of visual information less reliable [102]. It has been found that reduced attentiveness induces longer processing time delay [50]. In such case, it is hypothesized that the haptic guidance system is capable of compensating the decrement of lane following performance caused by declined visual attention, by means of providing reliable haptic information to drivers.

Fatigue is a well-known human factor that causes traffic collisions all over the world. The National Highway Traffic Safety Administration (NHTSA) conservatively estimates that 100,000 police-reported crashes are the direct result of driver fatigue each year, which results in an estimated 1,550 deaths, 71,000 injuries, and 12.5 billion dollars in monetary losses [103]. Driver fatigue can be subcategorized into sleep-related and task-related fatigue according to the underlying factors contributing to the fatigue state. Task-related fatigue is caused by the driving task and the driving environment. Task-related driver fatigue is likely to occur after a prolonged drive on a monotonous road [104].

Driver fatigue is a detriment to driver safety and, therefore, technology designed to assess and counteract driver fatigue is necessary. Previous studies have used a number of measurements to assess driver fatigue, such as electrocardiogram (ECG) [105], electroencephalography (EEG) [106], and eye movement [107], [108]. Heart rate variability (HRV), calculated based on ECG, has been widely used to assess levels of driver fatigue with high accuracy [109, 110]. As one type of eye movement measurements, percentage of eye closure (PERCLOS) has been regarded as a useful index in detecting driver fatigue by the NHTSA [111].

Throughout the years, various types of countermeasures for driver fatigue have been evaluated. Three low-cost engineering treatments, including variable message signs, chevrons, and rumble strips, were implemented in a driving simulator study to assess their effects on alleviating the symptoms of driver fatigue [107]. Compared with visual signals, audio signals were easier to detect for inattentive drivers, and the 1,750 Hz audio signal achieved the best performance in the warning process [112]. Vibrations with frequency varying from 100 to 300 Hz were produced on the seat belt to warn fatigued drivers and, furthermore, the warning required manual deactivation rather than automatic deactivation [113].

Although the above methods adequately dealt with driver fatigue, they still have concerning limitations. When warning signals are designed, habituation, or the decreasing likelihood that the driver will notice the warning as the number of subsequent encounters increases, must be considered [114]. Thus, simple audio and vibration signals tend to gradually lose their effect as they are repeated. In addition, it is difficult for audio signals to grab the attention of drivers who are engaging in auditory non-driving tasks, such as listening to music [115]. Sometimes, tactile stimulus can also be easily overlooked due to engine vibrations, an uneven roadway, or thick clothing. Intense vibration stimulus, however, might cause annoyance, discomfort, or pain to drivers [116].

Based on our study [77], the haptic guidance system has been designed to support the driver by generating torques on a human-machine interface. For example, haptic guidance could assist drivers to perform the appropriate actions for curve negotiation by producing both the recommended direction and magnitude of the suitable steering operation [14]. Up to now, the studies have evaluated the applications of haptic guidance to assist in lane following on straight lanes and along curves, as well as obstacle avoidance, mainly for the drivers in the normal condition rather than the fatigue-related condition of which the visual information becomes less reliable.

It is hypothesized that the haptic guidance system is capable of compensating the decrement of lane following performance caused by declined visual attention under fatigue driving. In addition, it is also expected that the continuous active torque on the steering wheel, as a stimulus, would increase the attention of fatigued driver. Therefore, the goal of this experiment is to investigate the effect of haptic guidance on fatigue-related driver behavior.

### 3.3.2 Experimental design

A driving simulator experiment was conducted with 12 participants. The experiment was approved by the Office for Life Science Research Ethics and Safety, the University of Tokyo (No. 14-113). The details of experimental design, including participants, apparatus, experiment conditions, experiment scenario and protocol, are presented below.

#### 3.3.2.1 Participants

Twelve healthy males were recruited to participate in the experiment. Their age ranged from 22 to 41 (mean = 25.8 years old, SD = 5.3). All had a valid Japanese driver's license for at least 2 years (mean = 4.9 years, SD = 2.3), and their average driving frequency was once per week. Each participant received monetary compensation for the involvement in the experiment.

#### 3.3.2.2 Apparatus

The experiment was conducted in a high-fidelity driving simulator (Mitsubishi Precision Co., Ltd., Japan), which is the same as the one used in experimental study I, as shown in Figure 3.1 in Section 3.1.2.

#### 3.3.2.3 Experiment conditions

The experiment consisted of two conditions, a treatment session that implemented the haptic guidance steering system and a control session without the system. The same driving course was used for both control and treatment sessions. Each participant took part in both control and treatment sessions. The control session and treatment session were conducted at the same time on two different days and counterbalanced among the participants in order to mitigate the order effect; to be specific, six of twelve participants took part in the control session on the first day and the other six participants took part in the treatment session on the first day. The counterbalanced design of trial order is shown in Table 3.14.

Table 3.14 Counterbalanced design of trial order.

| Participants | 1st trial | 2nd trial |
|---|---|---|
| No.1 | 2 | 1 |
| No.2 | 1 | 2 |
| No.3 | 1 | 2 |
| No.4 | 2 | 1 |
| No.5 | 1 | 2 |
| No.6 | 2 | 1 |
| No.7 | 2 | 1 |
| No.8 | 1 | 2 |
| No.9 | 1 | 2 |
| No.10 | 2 | 1 |
| No.11 | 2 | 1 |
| No.12 | 1 | 2 |

1: Control session; 2: Treatment session

The haptic guidance torque was determined by Equation 2.3 in Section 2.5.2. The values of $K_1$, $a_1$, $a_2$, $a_3$, and $a_4$ were decided as 1.0, 0.16, 0.004, 1.8, and 0.045, respectively, through a trial-and-error method. The method was generally divided into two steps. The first step was to find rough solution with aggressive tuning. The second step was to find desired solution with small tweaks in order to achieve good performance.

**3.3.2.4 Experiment scenario**

The driving environment was a two-lane rural road with 3.15 m lane widths and 0.6 m shoulder widths, as shown in Figure 3.35. Lane markings were solid lines and dashed lines. Several cyclists and oncoming vehicles appeared at various intervals. The eye-gaze tracking system (Smart Eye Pro, Smart Eye AB, Sweden) was employed to measure the eye movement behaviors of the participants in the experiment.

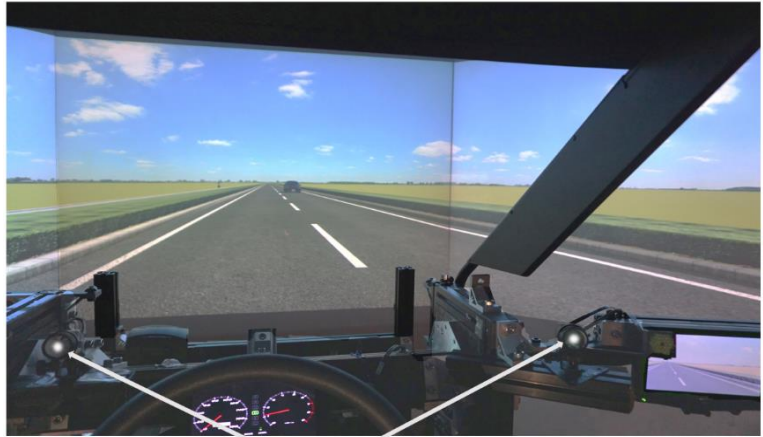

Eye tracking cameras

Figure 3.35 Driving environment and Smart Eye Pro system.

The driving course was composed of six parts and the total length was 24 km, as shown in Figure 3.36. All parts, excluding part 3, were identical and each consisted of prolonged straight roads and four curves with radius of 100 m. Part 3 was a 4 km straight road without any curves, where driver fatigue was extremely induced. In the treatment session, as shown in Figure 3.36 (b), haptic guidance was activated during part 5. Meanwhile, the drivers were still required to provide their steering torque to cooperate with the haptic guidance torque, for the haptic guidance torque was not capable of automated driving on the straight roads as well as the curves. Part 4 and part 5 were considered as 'before' and 'during' sections of the haptic system implementation. Then, the system was inactive during part 6. As shown in Figure 3.36 (a), a control session without haptic guidance was conducted as a comparison.

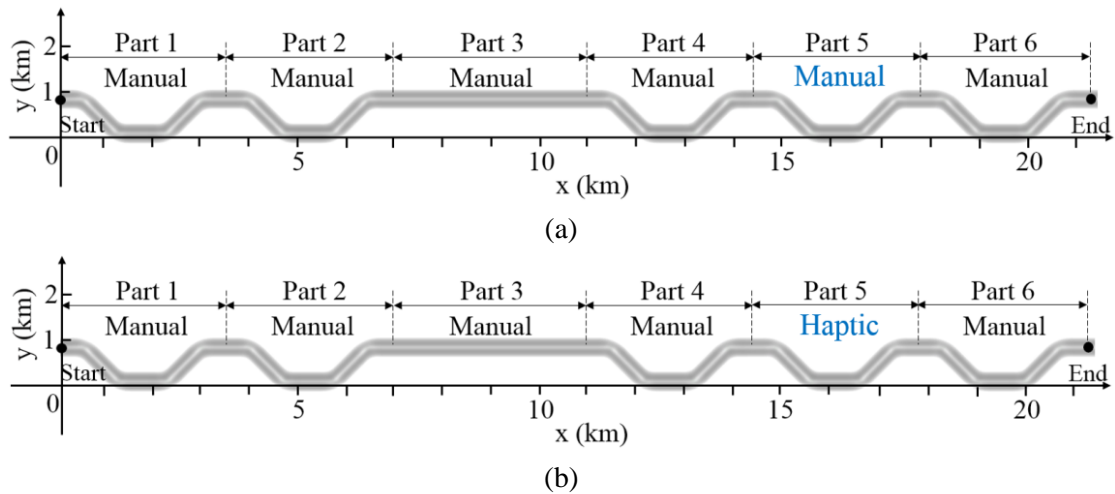

(a)

(b)

Figure 3.36 Driving course in the experiment. (a) Control session, (b) Treatment session.

A relatively low speed contributed to inducing driver fatigue during a prolonged drive on a monotonous road and, otherwise, drivers might feel stressed at a high driving speed. The driving speed was fixed at 40 km/h by program to reduce the workload of the driving task. The drivers thus did not need to operate the accelerator pedal during the whole driving procedure. Driving speed was fixed, also because steering performance and speed choice were related and removing this variable eased driving performance assessment.

**3.3.2.5 Experiment protocol**

One day before the experiment, participants were requested to have a good night's sleep of 7 hours. In addition, caffeine and alcohol consumption was not allowed during the 12 hours prior to the experiment.

On the experiment day, first, the participants signed a consent form, which explained the procedure of the experiment. The participants were naive to the purpose of the experiment. The participants then filled out a questionnaire about their personal demographic information and driving experience. After that, they got into the driving simulator and adjusted their seat to achieve a normal driving position. Then, the Smart Eye Pro system was set up. Afterward, the participants were required to grab the steering wheel in a "ten-and-two" position and to start a practice drive to familiarize themselves with the driving simulator and road track in order to mitigate the learning effect. The participants were asked to follow Japanese traffic rules and to drive in the center of the lane as accurately as possible. Then, the participants were allowed 5 minutes of rest in the driving simulator. After the rest, the participants started the experimental driving session, which lasted for approximately 36 minutes. At the end of experiment, the participants got out of the driving simulator and completed a questionnaire.

**3.3.3 Data analysis**

Initial eye movement data were collected in the Smart Eye Pro system, and initial driving performance data were collected in the driving simulator. Further signal processing steps were conducted in the MATLAB, and are presented in the following parts. In addition, statistical analysis on the post-processed data is explained.

**3.3.3.1 Visual behavior measurement**

Percentage of eye closure (PERCLOS) is defined as the percentage of eyelid closure over the pupil over time. The algorithm used to compute PERCLOS was "PERCLOS P80" [117], which calculates the proportion of time that the eyes are 80% or more closed over a one-minute interval, as shown in Figure 3.37. PERCLOS has been established as a generally useful and reliable index of lapses in visual attention by NHTSA [111].

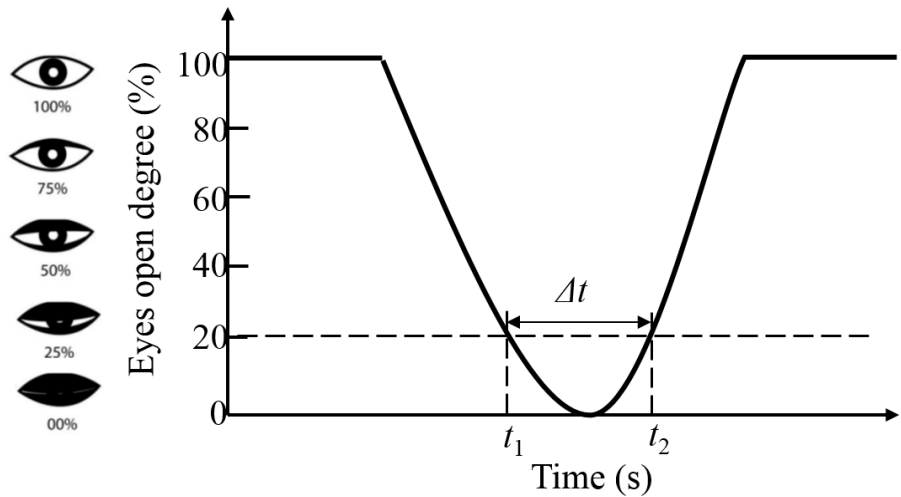

Figure 3.37 Eyes open degree for calculating PERCLOS, adapted from [118].

An example of eyelid opening within 10-s portion from one participant is shown in Figure 3.38. The eyelid opening was measured by the Smart Eye Pro system. It can be observed that the measured eyelid opening did not go all the way down to 0 mm, and it is a normal phenomenon according to[119]. It can be seen that the PERCLOS is larger in Figure 3.38 (b) compared to Figure 3.38 (a), which indicates that the PERCLOS increased over time from Part 1 to Part 5.

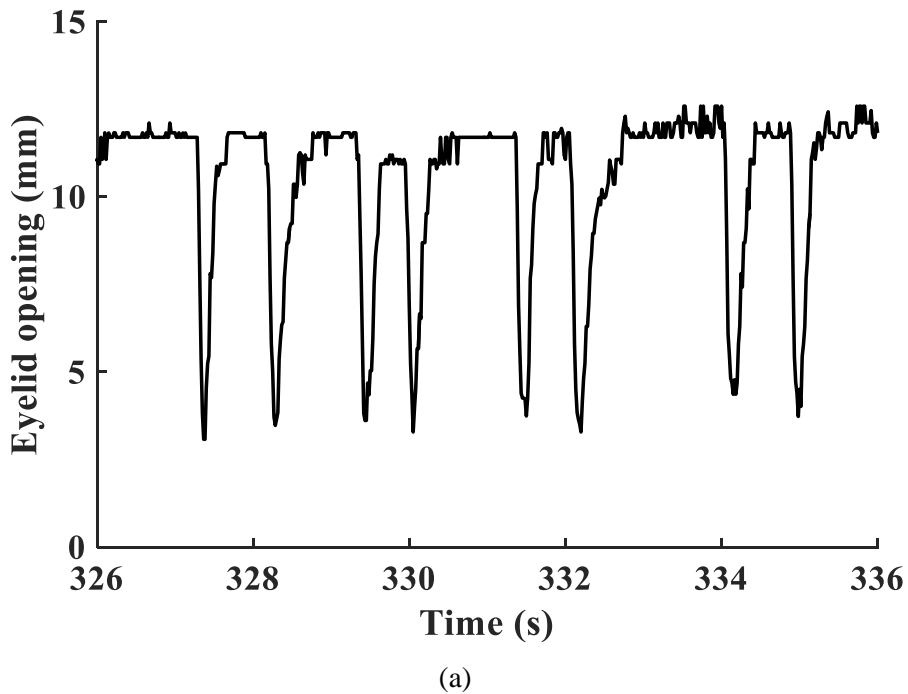

(a)

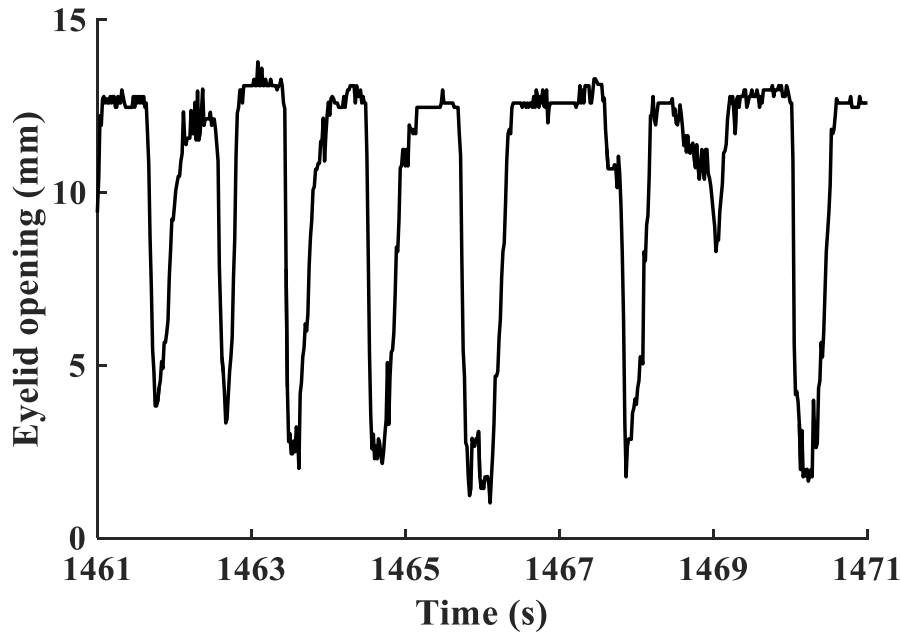

(b)

Figure 3.38 An example of eyelid opening within 10-s portion from one participant, which is measured by the Smart Eye Pro system. (a) measured within Part 1 of the control session (b) measured within Part 5 of the control session.

### 3.3.3.2 Lane following performance measurement

In this experiment, the lane following performance on straight lanes was measured by standard deviation of lane position and mean absolute lateral error.

- Standard deviation of lane position (SDLP)

The algorithm to calculated SDLP on straight lanes has been presented in Section 3.1.3. Decreased SDLP indicates higher lane following performance or decreased lane departure risk.

- Mean absolute lateral error (MALE)

Mean absolute lateral error, in relation to the centerline of the lane, was calculated to describe the lane following accuracy. Lower value of MALE indicates higher lane following accuracy or decreased lane departure risk.

### 3.3.3.3 Subjective evaluation

In this experiment, the perceived workload of the driving task was evaluated by subjective feedback. The perceived workload of the driving task was rated by using the NASA-TLX [90]. The participants were asked to use the NASA-TLX to assess their workload at the end of each driving trial in the experiment. Each item of the index was investigated separately to obtain each scale score. The details of the six items have been introduced in Section 3.1.3.

### 3.3.3.4 Statistical analysis

Repeated measures analysis of variance (ANOVA) was used for statistical analysis, and it has been explained in Section 3.1.3.

In this experiment, data were first analyzed using repeated measures ANOVA to investigate the effect of time (6 parts) on driver behaviors. Post hoc Bonferroni pair-wise comparisons were used to determine the differences between parts. The significance level was set to 0.05.

In addition, paired *t*-tests were performed to understand the effect of haptic guidance on the behavior of fatigued drivers, and the significance level was also set to 0.05. Paired *t*-test was used because the same group of subjects participated both driving conditions.

Regarding the *t*-test, first, the difference in SDLP between part 4 and part 5 in the treatment session was analyzed by paired *t*-test. Part 4 and part 5 represented the 'before' and 'during' sections of the haptic guidance implementation, respectively. Second, a paired *t*-test was used to compare SDLP between part 4 and part 5 in the control session. Afterward, the variation of SDLP between part 4 and part 5 in the treatment session was compared with that in the control session, also by using a paired *t*-test. The variation of SDLP between part 4 and part 5 was expressed as

$$SDLP_{\text{var}} = \left(\frac{SDLP_{\text{part 5}}}{SDLP_{\text{part 4}}} - 1\right) \times 100\% \;, \tag{3.4}$$

and $SDLP_{\text{var}}$ was a relative value; $SDLP_{\text{part4}}$ and $SDLP_{\text{part5}}$ were the mean values of SDLP across part 4 and part 5, respectively. The difference in $SDLP_{\text{var}}$ between the control and treatment sessions demonstrated the effect of haptic guidance on SDLP.

The above method for analyzing the results of SDLP was also employed for the results of MALE and PERCLOS.

### 3.3.4 Results

Results are presented separately for driver visual behavior, lane following performance, and subjective evaluation. The results show that the haptic guidance system was effective in improving lane following performance in the condition of declined visual attention under fatigue driving.

### 3.3.4.1 Driver visual behavior

Figure 3.39 illustrates the percentage of eye closure (PERCLOS) from part 1 to part 6 for both sessions. There was a significant time-on-task effect on PERCLOS from part 1 to part 6 ($p = 0.001$). Pair-wise comparisons revealed that PERCLOS was higher in part 3 than in part 1 ($p =$ 0.025) and higher in part 4 than in part 1 ($p = 0.030$). It is indicated that there was an increasing trend of PERCLOS, which resulted from declined visual attention under fatigue of the drivers. Additionally, there was no significant difference in PERCLOS between both sessions for the first four parts ($p = 0.738$). It indicates that there was no order effect for the two sessions.

In the control session, there was no significant difference in PERCLOS between parts 4 and 5. In the treatment session, PERCLOS was significantly lower in part 5 than in part 4 ($t = 2.970$, $p = 0.013$). Additionally, the difference in PERCLOS between parts 4 and 5 was significantly lower in the treatment session compared to the control session ($t = 2.457$, $p = 0.032$), as illustrated in Figure 3.40. The results indicate that applying haptic guidance had a significant effect on PERCLOS. It suggests that the implementation of the haptic guidance system in part 5 decreased PERCLOS; additionally, driver visual attention was still lower compared to part 1 as shown in Figure 3.39.

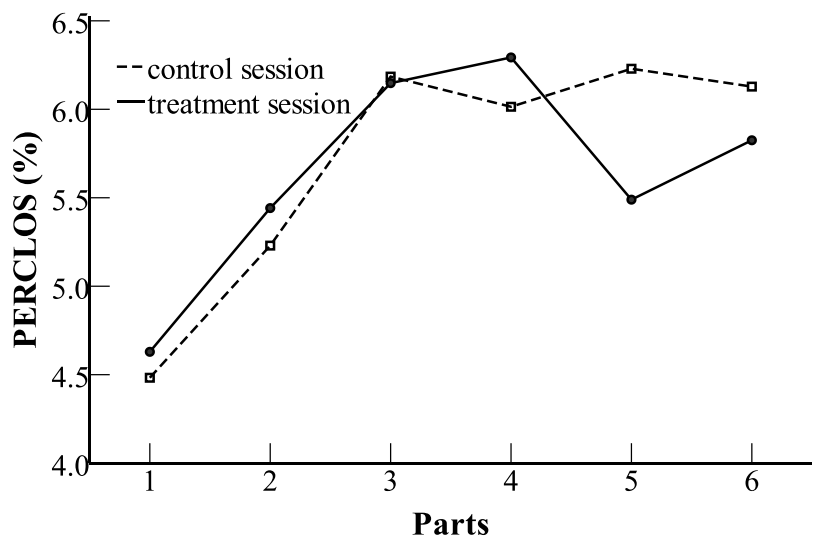

Figure 3.39 PERCLOS throughout the driving course in both sessions.

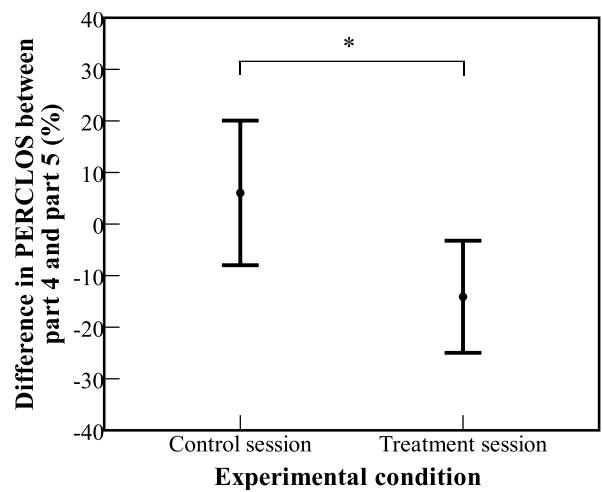

Figure 3.40 Difference in PERCLOS between part 4 and part 5 in both sessions. (*$p < 0.05$). Error bars denote 0.95 confidence intervals.

**3.3.4.2 Lane following performance**

The results of measured lane following performance consist of SDLP and MALE.

- Standard deviation of lane position (SDLP)

Figure 3.41 illustrates the mean value of standard deviation of lane position (SDLP) from part 1 to part 6 for both sessions. There was a significant time-on-task effect on SDLP from part 1 to part 6 ($p = 0.004$). Pair-wise comparisons revealed that mean SDLP was higher in part 4 than in part 1 ($p = 0.018$), higher in part 4 than in part 2 ($p = 0.014$), and higher in part 6 than in part 1 ($p = 0.027$). It is indicated that there was an increasing trend of mean SDLP, which resulted from declined visual attention under fatigue of the drivers. Additionally, there was no significant difference in SDLP between both sessions for the first four parts ($p = 0.987$). It indicates that there was no order effect for the two sessions.

In the control session, there was no significant difference in mean SDLP between parts 4 and 5. In the treatment session, mean SDLP was significantly lower in part 5 than in part 4 ($t = 4.699$, $p = 0.001$). Moreover, the difference in mean SDLP between parts 4 and 5 was significantly lower in the treatment session compared to the control session ($t = 3.033$, $p = 0.011$), as illustrated in Figure 3.42. The results show that the effect of haptic guidance on SDLP was significant. It is indicated that the decrement of lane following performance on straight lanes caused by declined visual attention under fatigue driving could be compensated by the haptic guidance system.

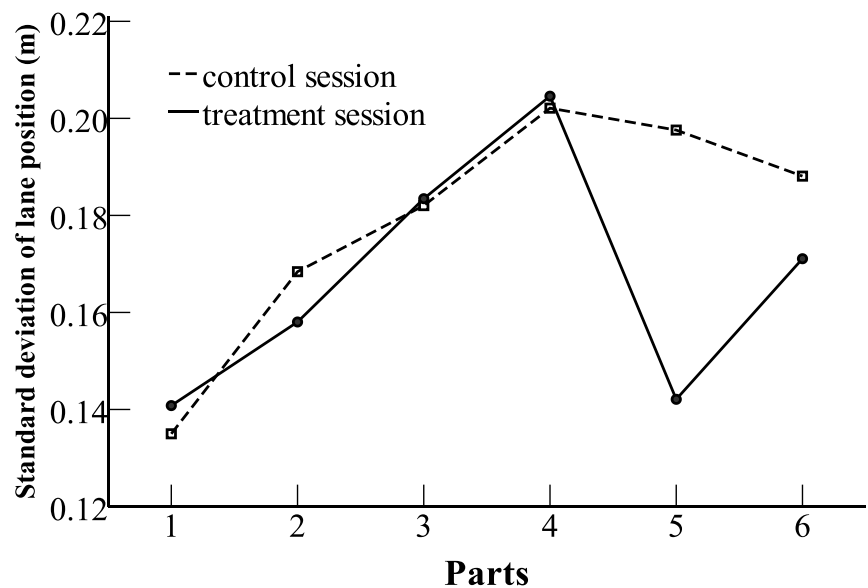

Figure 3.41 Standard deviation of lane position throughout the driving course in both sessions.

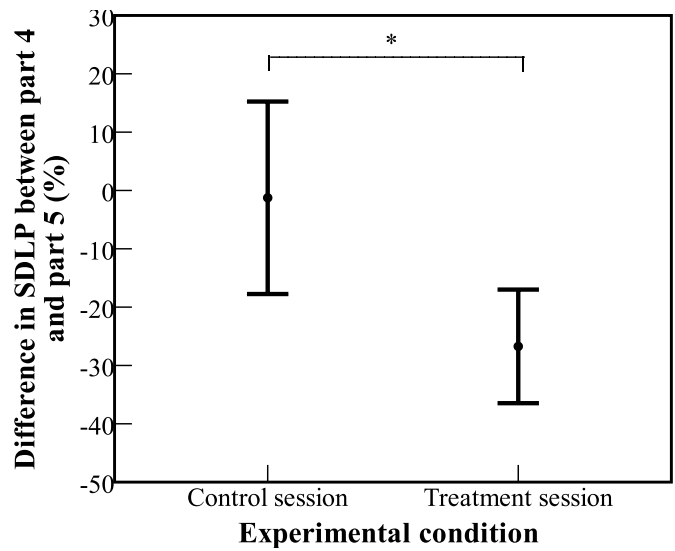

Figure 3.42 Difference in SDLP between part 4 and part 5 in both sessions. (*p < 0.05). Error bars denote 0.95 confidence intervals.

- Mean absolute lateral error (MALE)

Figure 3.43 illustrates the mean value of absolute lateral error from part 1 to part 6 for both sessions. There was a significant time-on-task effect on mean absolute lateral error from part 1 to part 6 ($p = 0.009$). With pair-wise comparisons, no significant difference between parts was found. Additionally, there was no significant difference in mean absolute lateral error between both sessions for the first four parts ($p = 0.470$). It indicates that there was no order effect for the two sessions.

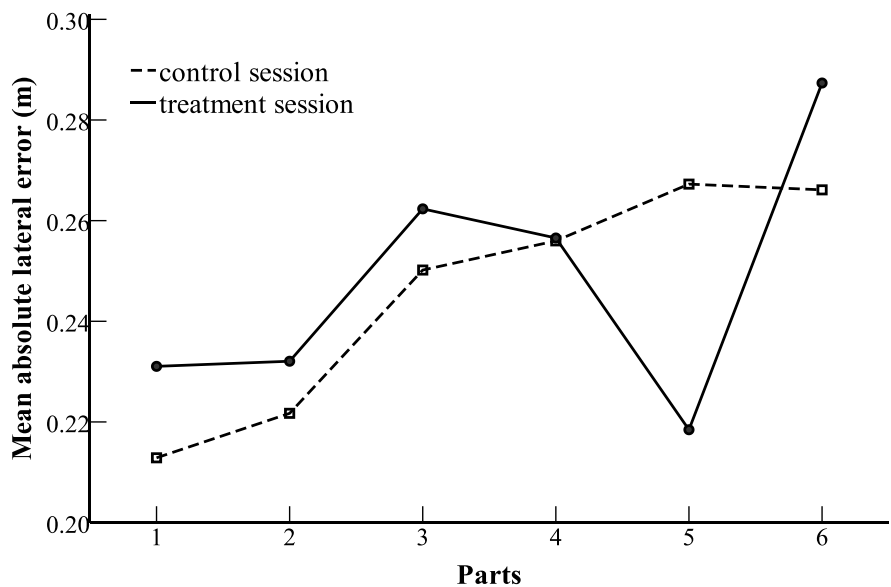

Figure 3.43 Mean absolute lateral error throughout the driving course in both sessions.

Mean absolute lateral error was significantly lower in part 5 than in part 4 ($t = 2.643$, $p = 0.023$) in the treatment session, and no significant difference was found in the control session. Furthermore, the difference in mean absolute lateral error between parts 4 and 5 was significantly lower in the treatment session than in the control session ($t = 2.593$, $p = 0.025$), as illustrated in Figure 3.44. The results show that haptic guidance had a significant effect on mean absolute lateral error. Similar to the result of SDLP, it is also indicated that the decrement of lane following performance on straight lanes caused by declined visual attention under fatigue driving could be compensated by the haptic guidance system.

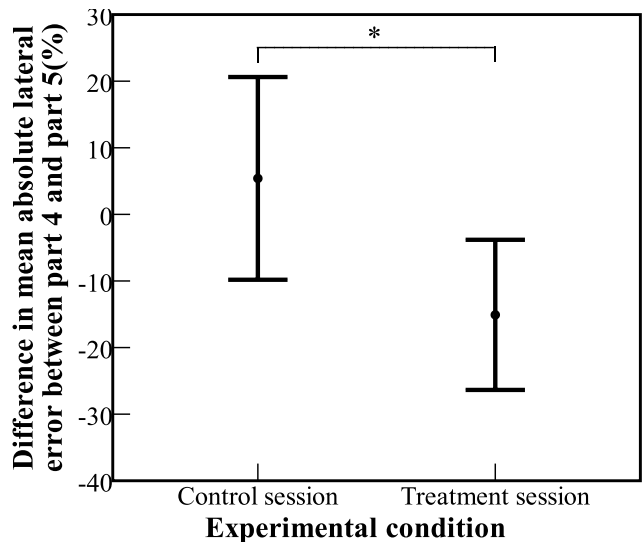

Figure 3.44 Difference in mean absolute lateral error between part 4 and part 5 in both sessions. (*p < 0.05). Error bars denote 0.95 confidence intervals.

### 3.3.4.3 Subjective evaluation

The result of NASA-TLX is shown in Figure 3.45. As for Physical Demand, a paired *t*-test revealed a significant difference between the control session and the treatment session ($t = -2.816$, $p = 0.017$). The physical demand was significantly higher in the treatment session (mean = 47.9, SD = 22.4) than in the control session (mean = 33.75, SD = 21.25). As for the other items, there was no significant difference between the control session and the treatment session. One explanation of the observation about the difference in physical demand would be that the drivers operated the steering wheel more frequently when positively relied on the haptic guidance. As a result, the lane following performance was improved by the haptic guidance as indicated by the results of SDLP and MALE.

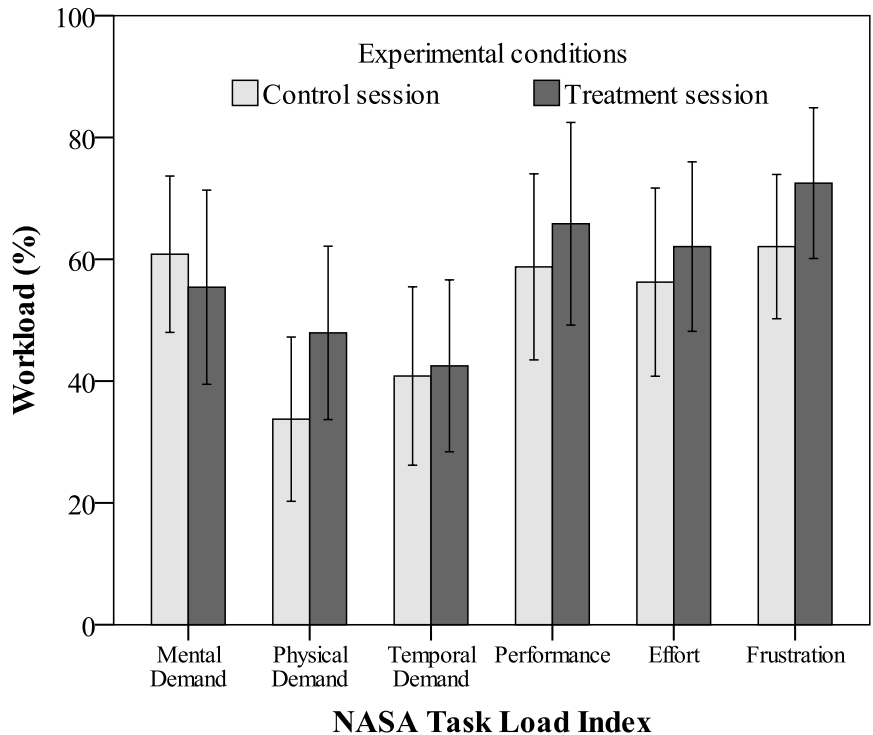

Figure 3.45 Mean scores on NASA-TLX. Error bars denote 0.95 confidence intervals.

**3.3.5 Discussion**

Related to the results of lane following performance, the decrease of SDLP and of MALE, when haptic guidance was activated, indicates that the risk of lane departure was reduced. These results were expected because the haptic system continuously guided the drivers by exerting an assistive torque, and the drivers reacted by operating the steering wheel with relying on the haptic guidance. The mean value of SDLP was found decreased from part 4 to part 6 in the control session, though the tendency was not significant ($p = 0.593$). One explanation would be that the steering performance of fatigued drivers could fluctuate. As for MALE, no significant difference was found between part 1 and part 4 according to the pair-wise comparison; even the effect of driver fatigue on mean absolute lateral error was found throughout the driving scenario. One would expect the mean absolute lateral error to be higher in part 4 than in part 1 due to driver fatigue, like SDLP. One explanation would be that drivers were not able to perfectly follow the centerline of a lane due to its invisibility in the driving scenario. In other words, even when the drivers were alert, there might exist a constant absolute lateral error.

According to the results of PERCLOS, it is indicated that the visual attention of drivers was raised when haptic guidance was activated after a prolonged driving; additionally, driver visual attention was still lower than the normal state at the beginning of the driving task. Moreover, the physical demand of the driving task was significantly affected by the haptic guidance. It can be argued that the drivers had to operate the steering wheel more frequently when positively cooperating with the haptic guidance. The mental demand of the driving task, however, was not significantly affected by the haptic guidance. In both sessions, the mental demand of drivers was at a low level due to the monotonous road and minimal traffic.

It should be noticed that it is better for a driver who gets fatigued to stop the vehicle for a break than to continue driving; however, fatigued drivers might tend to insist driving by ignoring the suggestion given by a warning system, as these drivers underestimated the impact of driver fatigue [120]. In this case, the haptic guidance steering system would be beneficial to fatigued drivers by arousing them and assisting them to continue driving.

A limitation of this experiment is that only a short-term implementation of the haptic guidance system was considered. The haptic guidance system was active for a duration of 6 minutes, and it is a relatively shorter period compared to the whole driving course. It has been argued that the effect of long-term use of haptic guidance on driver behavior might be different from short-term use [7]. The drivers might get used to the haptic guidance system after a long-term driving, and the driver behavior is unclear in such situation. Considering this, an experiment applying long-term use of haptic guidance system as a treatment session and applying the same period of manual driving as a control session, would be an interesting future study.

### 3.3.6 Conclusion

The goal of this experiment is to investigate the effect of haptic guidance on fatigue-related driver behavior, and to evaluate the effectiveness of the haptic guidance system on improving lane following performance of fatigued driver. The experiment was conducted with 12 participants in a high-fidelity driving simulator. A treatment session was arranged with the haptic guidance steering system, and a control session was conducted as a comparison.

The results of SDLP and MALE indicate that the decrement of lane following performance on straight lanes caused by declined visual attention under fatigue could be compensated by haptic information provided by the guidance system. On the other hand, the active torque on the steering wheel stimulated the fatigued driver, and according to the result of PERCLOS, the driver visual attention was increased. It can be concluded that the implementation of the haptic guidance system, which continuously provided reliable haptic information (guidance torque) on the steering wheel to stimulate and guide the drivers, was effective in improving drivers' lane following performance in the condition of declined visual attention under fatigue driving.

To conclude the three experimental studies in this chapter, the design and evaluation of the haptic guidance system were performance by driving simulator experimental studies, in cases of normal and degraded visual information including visual occlusion from road ahead and declined visual attention under fatigue driving. The results indicate that the proposed haptic guidance system was capable of providing reliable haptic information, and the driver tended to integrate visual and haptic information to achieve better lane following performance. Still, the objective evaluation of the effect of haptic guidance on driver behavior required the development of a suitable driver model. Moreover, numerical analysis is useful to investigate the effect of the haptic guidance system on lane following performance, which will be addressed in the next chapter.

# Chapter 4

# Behavior Modeling and Numerical Analysis

# 4 Behavior Modeling and Numerical Analysis

## 4.1 Introduction

Chapter 3 explained the design and evaluation of the haptic guidance system, by driving simulator experimental studies, in cases of normal and degraded visual information including visual occlusions and declined visual attention under fatigue. The results indicate that the proposed haptic guidance system was capable of providing reliable haptic information, and the driver tended to integrate visual and haptic information to achieve better lane following performance. Therefore, the decrement of lane following performance caused by degraded visual information was compensated by the haptic information. However, it is unknown how the driver integrates visual and haptic information. In addition, the objective evaluation of the effect of haptic guidance on driver behavior requires the development of a suitable driver model with consideration of both visual and haptic information The goal of this chapter is therefore to better understand the driver behavior obtained in the experiments through behavior modeling and numerical analysis, and furthermore, to predict driver behavior for investigating the effect of haptic guidance on lane following performance in cases of degraded visual information.

To attain the goal, this chapter focuses on driver behavior modeling and numerical analysis. The proposed driver model is explained in Section 2.3 and 2.4. To identify the driver model parameters, measured data from experimental study I are used. After that, model validation test is performed in a SIMULINK model. Finally, numerical analysis is conducted to predict driver behavior in the conditions of normal and degraded visual information, and to investigate the effect of haptic guidance system on the driver's lane following performance.

## 4.2 Model parameters identification

The goal of this section is to show that the proposed model in Section 2.3 and 2.4 is appropriate to match the driver behavior with integrated feedback of visual and haptic information. The measured data in experimental study I were used for model identification in the conditions of Manual, HG-normal, and HG-strong. The Prediction Error Method (PEM) was used in the parameters identification.

### 4.2.1 Summary of used experimental data

The measured data included vehicle trajectory, haptic guidance torque, steering wheel angle, and driver input torque. The vehicle trajectory and haptic guidance torque were calculated by the host computer connected to the driving simulator. The steering wheel angle was measured by the angular sensor, and the driver input torque was measured by torque sensor, shown in Figure 3.2 in Section 3.1.2.

For the reason that the participants needed time to adapt their driving behavior to the haptic guidance steering system, the data recorded between P1 and P2 on the driving course were used for driver model identification for each participant, as shown in Figure 4.1.

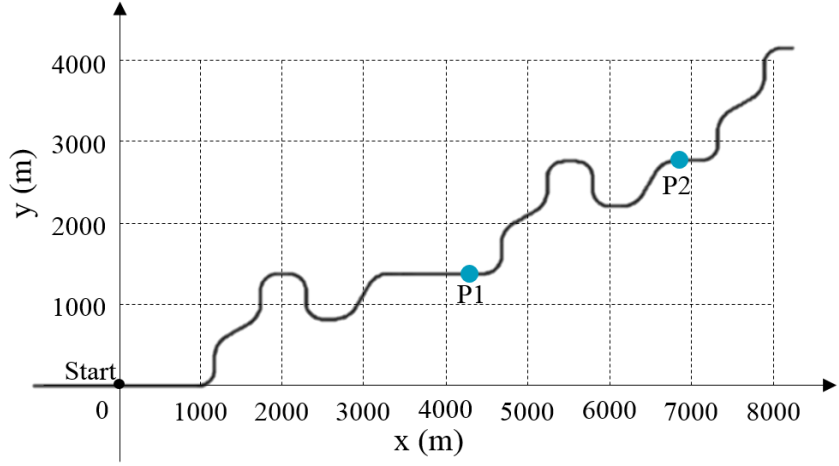

Figure 4.1 Driving course in the experiment.

### 4.2.2 Identification approach

The linear driver model with integrated feedback of visual and haptic information is shown in Figure 2.13 in Section 2.8. By using lateral error $e_y$, yaw error $e_\theta$, steering wheel angle $\varphi$ and haptic guidance torque $T_h$ as inputs, and driver input torque $T_d$ and target steering wheel angle $\varphi'$ as outputs, the driver model is represented as a structured state-space realization, as follow:

$$\begin{aligned}\dot{x}(t) &= Ax(t) + Bu(t) \\ y(t) &= Cx(t) + Du(t) + e(t) \\ x(0) &= x_0\end{aligned} \quad (4.1)$$

$$\begin{bmatrix}\dot{x}_1\\\dot{x}_2\\\dot{x}_3\end{bmatrix} = \begin{bmatrix}0 & 0 & 0\\a_2\dfrac{2}{t_p} & -\dfrac{2}{t_p} & 0\\-a_2\dfrac{K_d+K_{nms}}{t_{nms}} & \dfrac{2(K_d+K_{nms})}{t_{nms}} & -\dfrac{1}{t_{nms}}\end{bmatrix}\begin{bmatrix}x_1\\x_2\\x_3\end{bmatrix}$$

$$+ \begin{bmatrix}1 & 0 & 0 & 0\\a_1\dfrac{2}{t_p} & a_4\dfrac{2}{t_p} & 0 & 0\\-a_1\dfrac{K_d+K_{nms}}{t_{nms}} & -a_4\dfrac{K_d+K_{nms}}{t_{nms}} & -\dfrac{K_{nms}}{t_{nms}} & -\dfrac{K_{hf}}{t_{nms}}\end{bmatrix}\begin{bmatrix}e_y\\e_\theta\\\varphi\\T_h\end{bmatrix}$$

$$\begin{bmatrix}T_d\\\varphi'\end{bmatrix} = \begin{bmatrix}0 & 0 & 1\\-a_2 & 2 & 0\end{bmatrix}\begin{bmatrix}x_1\\x_2\\x_3\end{bmatrix} + \begin{bmatrix}0 & 0 & 0 & 0\\-a_1 & -a_3 & 0 & 0\end{bmatrix}\begin{bmatrix}e_y\\e_\theta\\\varphi\\T_h\end{bmatrix} \quad (4.2)$$

It should be noticed that $a_3$ is not presented in the above state-space representation, as it is negligible in the condition of normal visual information [49, 54]. In addition, an analysis indicates the low identifiability of the model when considering the output of driver torque only. It may result in an identified model with a higher fitting, but the vehicle does not follow the target trajectory adequately in the simulation study. For this reason, the target steering wheel angle is considered as an additional output of the driver model. The parameters of $a_1$, $a_2$, and $a_4$, consequently, are decided according to the fitness of steering wheel angle compared to target steering wheel angle.

A first-order Pade expansion to approximate the time delay $t_p$ with a rational transfer function is performed, as follows:

$$e^{-t_p s} = \frac{1 - 0.5 t_p s}{1 + 0.5 t_p s} \tag{4.3}$$

The discretized state-space realization is given by:

$$\begin{aligned} x(k+1) &= Ax(k) + Bu(k) \\ y(k) &= Cx(k) + Du(k) + e(k) \\ x(0) &= x_0 \end{aligned} \tag{4.4}$$

This state-space realization was used for identifying the parameters of the driver model by Grey Box Identification Toolbox of Matlab. The parameters of the driver model were identified by using Prediction Error Method (PEM) implemented in the Grey Box Identification Toolbox. Minimum percentage difference between the current value of the loss function and its expected improvement after the iteration was set as 0.01%, which means that the iteration stopped when the expected improvement was less than 0.01%. The estimate of the expected loss function improvement at the next iteration was based on the Gauss-Newton vector computed for the current parameter value. Table 4.1 shows the default value and variation interval of the identified driver model parameters, some of which ($a_1$, $a_2$, $a_4$, $K_d$, and $K_{hf}$) were obtained through a trial-and-error process, and some of which ($t_p$, $K_{nms}$, and $t_{nms}$) according to the references [47, 56].

Table 4.1 Driver model parameters.

|  | **Default value** | **Variation interval** |
|---|---|---|
| $a_1$ | 0.1 | [0-0.5] |
| $a_2$ | 0.01 | [0-0.1] |
| $a_4$ | 3.7 | [3-5] |
| $t_p$ | 0.1 | [0.01-0.3] |
| $K_d$ | 3 | [1-5] |
| $K_{hf}$ | 0.5 | [0-1] |
| $K_{nms}$ | 1 |  |
| $t_{nms}$ | 0.1 |  |

**4.2.3 Identification results**

The driver input torque fitting result from a typical subject in the condition of HG-normal shows the comparison of driver input torque between actual human driver and the identified model; the fitness was 73.7%, as shown in Figure 4.2.

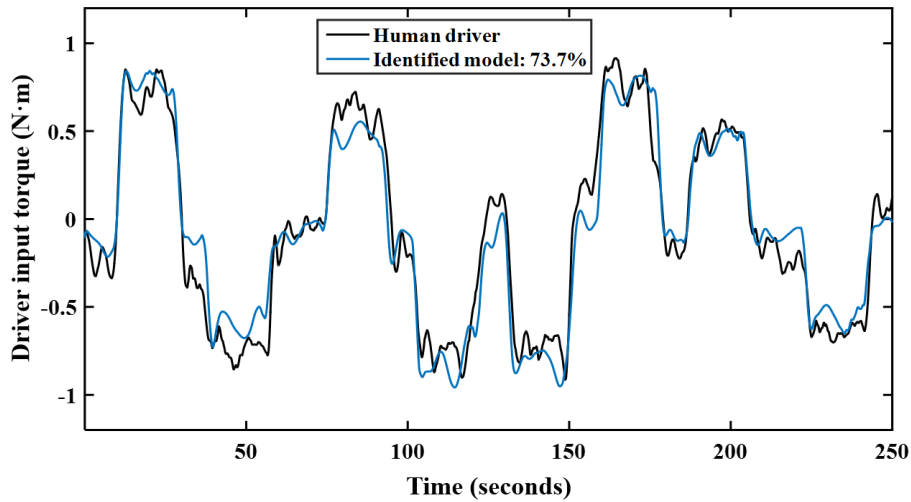

Figure 4.2 An example of driver input torque fitting in the condition of HG-normal.

As for the results from all the participants, we found that the proposed driver model matched driver input torque with a fitness of 76% on average in the condition of manual, a fitness of 69% on average in the condition of HG-normal, and a fitness of 57% on average in the condition of HG-strong. The decreased fitness in the conditions of HG-normal and HG-strong resulted from the complicated driver behavior of interaction with the haptic guidance system.

According to Figure 2.7 in Section 2.4.3, the driver provides integrated feedback of visual and haptic information through a neuromuscular system. $K_d$, which is the steering angle to torque gain, represents that the neuromuscular system provides a steering torque proportional to the target steering wheel angle. $K_{hf}$, which is the neuromuscular reaction gain for haptic feedback, represents driver interaction and reliance on the haptic guidance system. Regarding the identification results of $K_d$ and $K_{hf}$, promising findings were drawn among all the participants, which are discussed below.

Figure 4.3 shows the result of comparison in $K_d$ between the driving conditions of Manual, HG-normal and HG-strong. The value of $K_d$ was significantly different among Manual, HG-normal, and HG-strong ($F(2,26) = 11.476, p < 0.001$). Post hoc pairwise comparison results are shown in Table 4.2. It can be observed that the value of $K_d$ was significantly higher in the condition of Manual compared to HG-normal and HG-strong. It indicates that the drivers tended to reduce the angle to torque gain when interacting with the haptic guidance steering system. In addition, there was no significant difference between HG-normal and HG-strong.

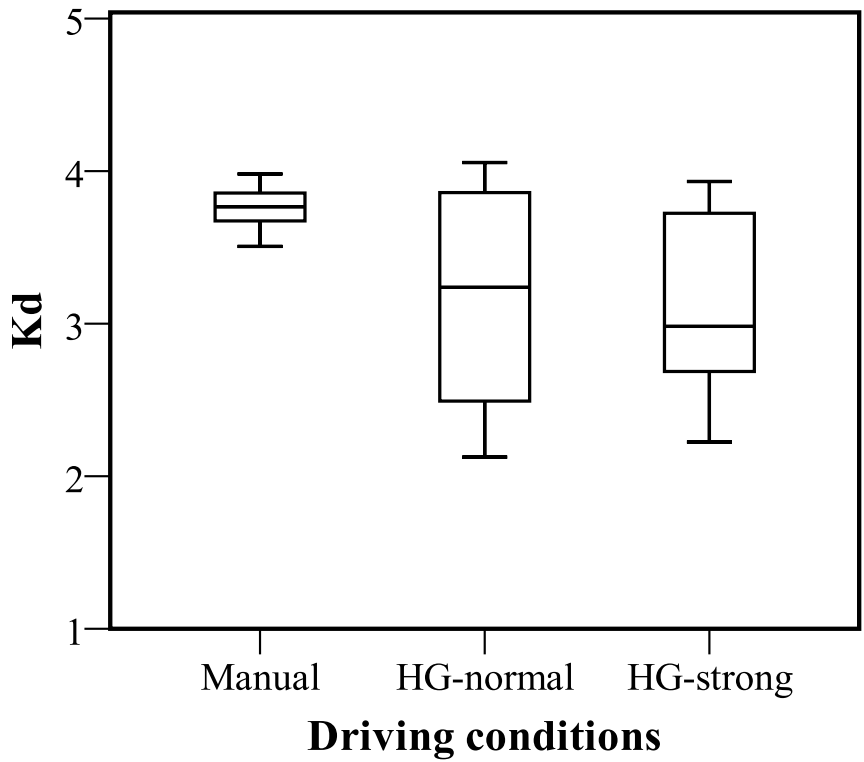

Figure 4.3 Comparison in $K_d$ between the driving conditions of Manual, HG-normal and HG-strong. A rectangle represents the middle 50% of a set of data. A horizontal line drawn through a rectangle corresponds to the median value of a set of data. An upper bar indicates the maximum value of a set of data, excluding outliers. A lower bar represents the minimum value of a set of data, excluding outliers.

Table 4.2 Post hoc test of $K_d$.

|  | Manual | HG-normal | HG-strong |
|---|---|---|---|
| Manual | - | - | - |
| HG-normal | 0.009** | - | - |
| HG-strong | 0.000*** | 1.000 | - |

***: $p < 0.001$, **: $p < 0.01$, *: $p < 0.05$

The result of comparison in $K_{hf}$ between the driving conditions of HG-normal and HG-strong is shown in Figure 4.4. The value of $K_{hf}$ was not significantly different between HG-normal and HG-strong ($t = -1.749$, $p = 0.104$), although a tendency existed, and the median value was higher in the condition of HG-strong compared to HG-normal. It indicates that the participants relied on the strong haptic guidance in a relatively lower degree compared to the normal haptic guidance. This is reasonable, as higher degree of driver reliance on the strong haptic guidance would result in higher risks compared to the normal haptic guidance if the automation system fails.

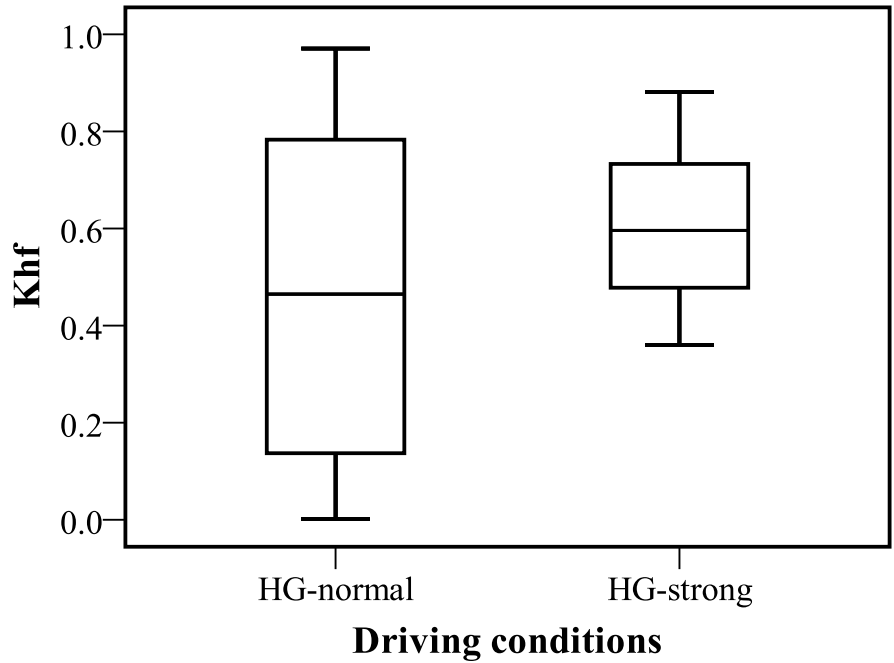

Figure 4.4 Comparison in $K_{hf}$ between the driving conditions of HG-normal and HG-strong. A rectangle represents the middle 50% of a set of data. A horizontal line drawn through a rectangle corresponds to the median value of a set of data. An upper bar indicates the maximum value of a set of data, excluding outliers. A lower bar represents the minimum value of a set of data, excluding outliers.

### 4.2.4 Model validation test

The model validation test was performed using SIMULINK in Matlab. The parameters of vehicle and steering systems (see Section 2.6 and 2.7) were identified using the measured data in the driving simulator experimental study I. Figure 4.5 shows the simulated output of the vehicle trajectory against the measurement vehicle trajectory from one typical participant in the condition of HG-normal. It can be observed that the simulated vehicle trajectory well followed the target trajectory within the lane along the whole driving course. In addition, the actual driver and the identified model showed similar profiles when negotiating a curve as shown in Figure 4.6. The absolute mean trajectory error along the whole driving course between the measured one and the simulated one was calculated, and was only 0.155 m, which also proves evidence that the simulated vehicle trajectory matched the measured vehicle trajectory.

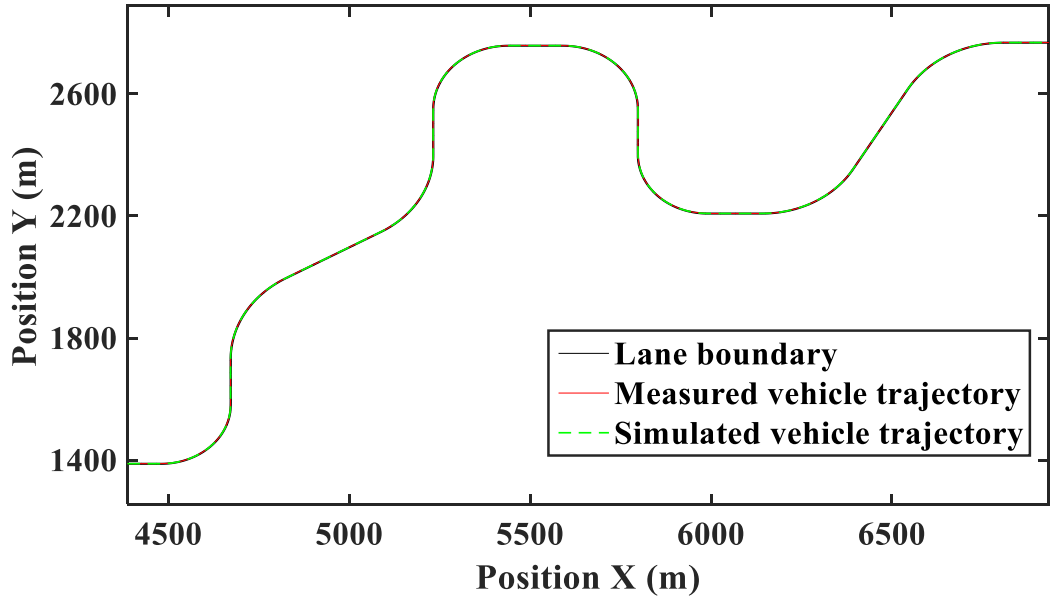

Figure 4.5 An example of comparison in vehicle trajectory between measured and simulated results along the driving course.

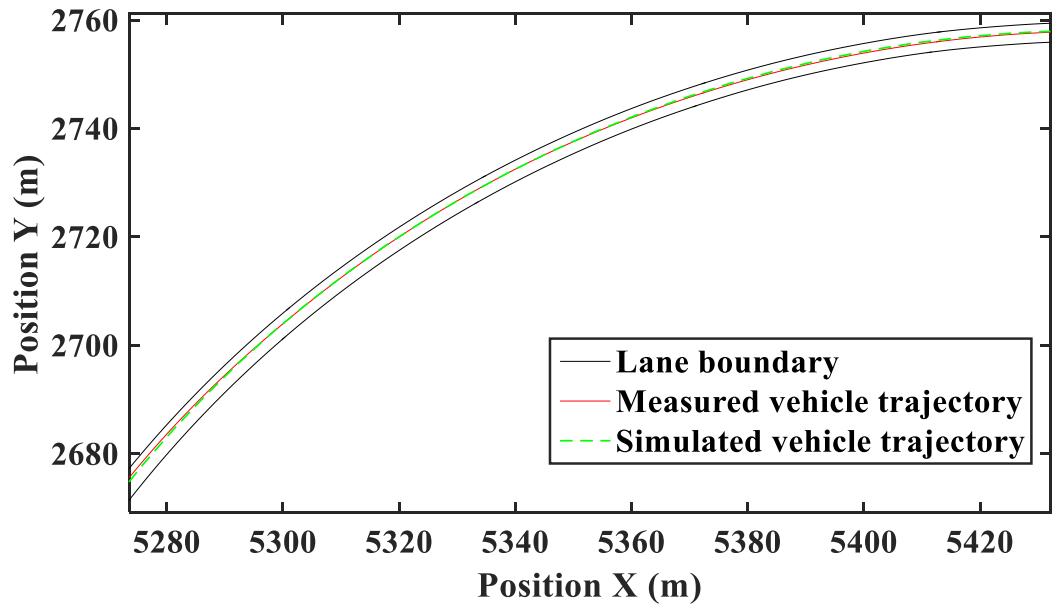

Figure 4.6 An example of comparison in vehicle trajectory between measured and simulated results along a curve negotiation.

**4.3 Numerical simulation of driver behavior**

The goal of this section is to provide evidence to support the proposed driver model and to evaluate the effect of haptic guidance system on lane following performance by numerical analysis. At first, the numerical analysis addressed the lane following performance of manual driving in cases of normal and degraded visual information, which includes low visibility and declined visual attention. Then, the effect of haptic guidance system on lane following performance was investigated.

### 4.3.1 Manual driving case

The manual driving case includes three driving conditions: normal visual information, low visibility, and declined visual attention. After presenting the driver model parameters for the three driving conditions, parameters of vehicle and steering systems used in the numerical analysis are illustrated and a driving course is described.

#### 4.3.1.1 Driving conditions

The model of driver visual system has been explained in Figure 2.4 in Section 2.3.2. Three driving conditions were addressed: normal visual information, low visibility, and declined visual attention. The driver model parameters in the three conditions are shown in Table 4.3, which were obtained by identification results, some ($t_f$, $t_n$, $t_p$) were according to references [51, 52, 54], and $a_3$ was obtained by a trial-and-error process.

Table 4.3 Driver model parameters in manual driving.

| Parameters | Normal | Low visibility | Declined visual attention |
|---|---|---|---|
| $a_1$ | 0.1 | 0.1 | 0.1 |
| $a_2$ | 0.05 | 0.05 | 0.05 |
| $a_3$ | - | 0.3 | - |
| $a_4$ | 3.7 | - | 3.7 |
| $t_f$ | 1.0 | - | 1.0 |
| $t_n$ | 0.3 | 0.3 | 0.3 |
| $t_p$ | 0.1 | 0.1 | 0.5 |

#### 4.3.1.2 Numerical values of system parameters

The simulations of three driving conditions were performed with SIMULINK in Matlab by using the numerical values of the parameters within the driver neuromuscular system, vehicle and steering systems, as shown in Table 4.4, which were obtained by identification and some ($t_{nms}$, $K_{nms}$) were according to references [46, 47]. The longitudinal speed of the vehicle was set to be 60 km/h corresponding to the driving speed in the experimental studies I and II.

Table 4.4 Parameters from driver neuromuscular system, vehicle and steering systems.

| Symbol | Value | Unit |
|---|---|---|
| $K_d$ | 3.8 | - |
| $t_p$ | 0.1 | s |
| $t_{nms}$ | 0.1 | s |
| $K_{nms}$ | 1.0 | - |
| $v$ | 60 | km/h |
| $m$ | 1100 | kg |
| $I$ | 2940 | kg m² |
| $l_f$ | 1 | m |
| $l_r$ | 1.635 | m |
| $K_f$ | 53300 | N/rad |
| $K_r$ | 117000 | N/rad |
| $J_s$ | 0.11 | kg m² |
| $B_s$ | 0.57 | N m s/rad |
| $K_t$ | 1/17 | - |
| $E_t$ | 0.026 | - |
| $K_s$ | 48510 | N m/rad |

### 4.3.1.3 Driving course

One curve negotiation basically consists of five sections: a straight lane before a curve; approaching the curve; along the curve; leaving the curve; and a straight lane after the curve, as shown in Figure 3.6 in Section 3.1.3. In the numerical simulation, a simple driving course was used as shown in Figure 4.7. The radius of the curve was 200 m, and the straight lane before the curve was 1000 m. The lane width was 3.6 m, and the centerline of lane was set as target trajectory.

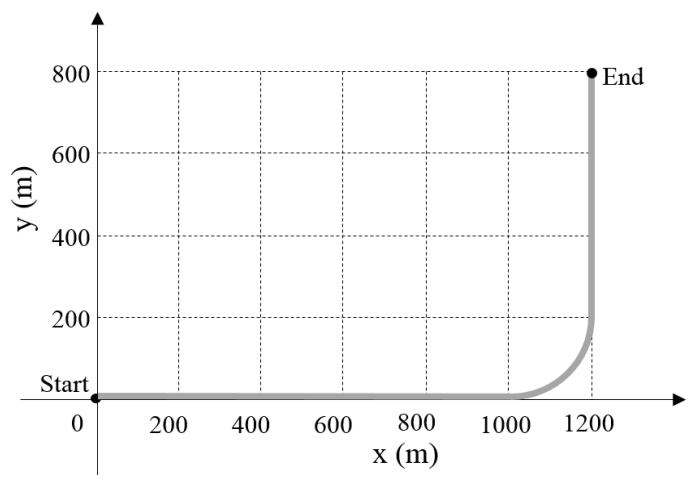

Figure 4.7 The simple driving course used in the numerical simulation.

An example of the simulation result on steering wheel angle of manual driving in the condition of normal visual information is shown in Figure 4.8 (a). To see the effect of haptic guidance and degraded visual information on driver lane following performance on the straight lane, a pulse signal was added on the steering wheel to imitate an inappropriate operation by the driver. The duration of the pulse signal was 2 seconds, and the magnitude was 1 N·m. An example of the simulation result on steering wheel angle when the pulse signal was added is shown in Figure 4.8 (b). A sudden increase of steering wheel angle can be observed, and it is followed by a fluctuated response.

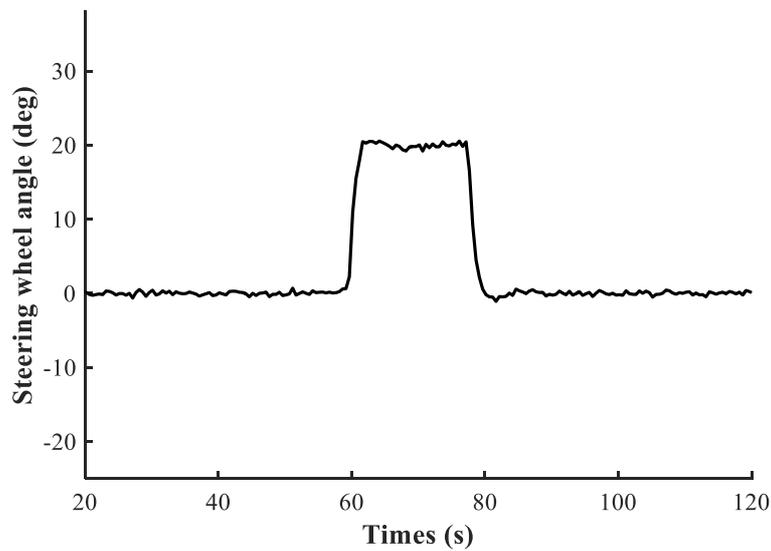

(a)

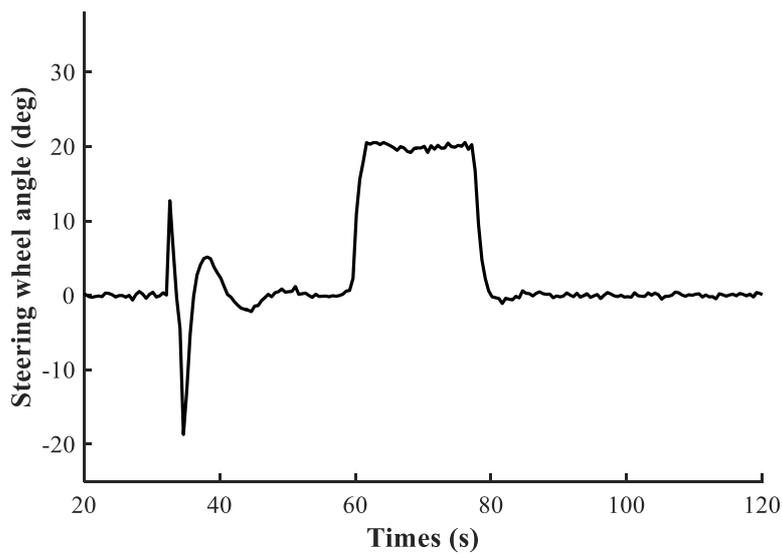

(b)

Figure 4.8 Steering wheel angle of manual driving in the condition of normal visual information. (a) Normal condition, (b) With a pulse signal on the straight lane before the curve.

### 4.3.2 Assisted driving case

The simulations of assisted driving with the haptic guidance system were performed with SIMULINK in Matlab. The numerical values of the parameters within driver neuromuscular system and haptic guidance system are shown in Table 4.5, which were obtained from identification results ($K_d$, $K_{hf}$), trial-and-error process and some ($t_{nms}$, $K_{nms}$) were according to references [47].

Table 4.5 Parameters of driver neuromuscular system and haptic guidance system.

| Symbol | Value | Unit |
|---|---|---|
| $K_d$ | 3.2 | - |
| $K_{hf}$ | 0.5 | - |
| $t_{nms}$ | 0.1 | s |
| $K_{nms}$ | 1.0 | - |
| $t'_n$ | 0.3 | s |
| $t'_f$ | 0.7 | s |
| $a'_1$ | 1.9 | - |
| $a'_2$ | 0.05 | - |
| $a'_3$ | 38 | - |
| $a'_4$ | 1.9 | - |
| $K_1$ | 0.25 | - |

### 4.4 Numerical simulation results

The numerical simulation results address driver lane following performance in the manual driving case and the assisted driving case. In each case, three driving conditions are addressed: normal visual information, low visibility, and declined visual attention.

### 4.4.1 Manual driving case

Figure 4.9 shows the comparison in steering wheel angle between the conditions of normal visibility and low visibility. It can be observed that there is an obvious peak of steering wheel angle when approaching and leaving the curve in the condition of low visibility, and the response is more fluctuated compared to the condition of normal visibility. The similar results were obtained from the measured steering wheel angle in experimental study II. According to the experimental results, driving condition of VF near yielded more fluctuated response of steering wheel angle compared to the condition of VF whole. In terms of the perturbation on the straight lane, both conditions of normal visibility and low visibility yielded a quick response to maintain the vehicle position.

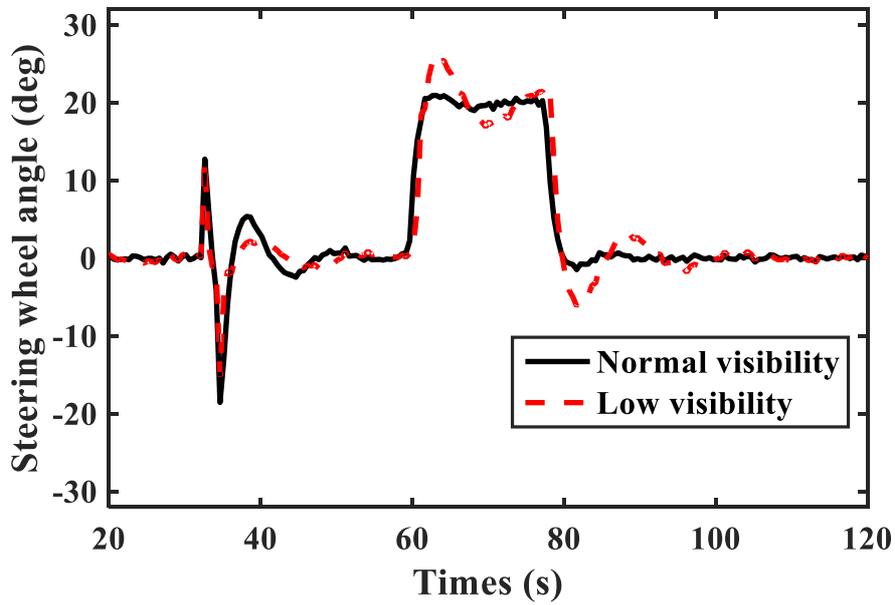

Figure 4.9 Comparison in steering wheel angle between the conditions of normal visibility and low visibility.

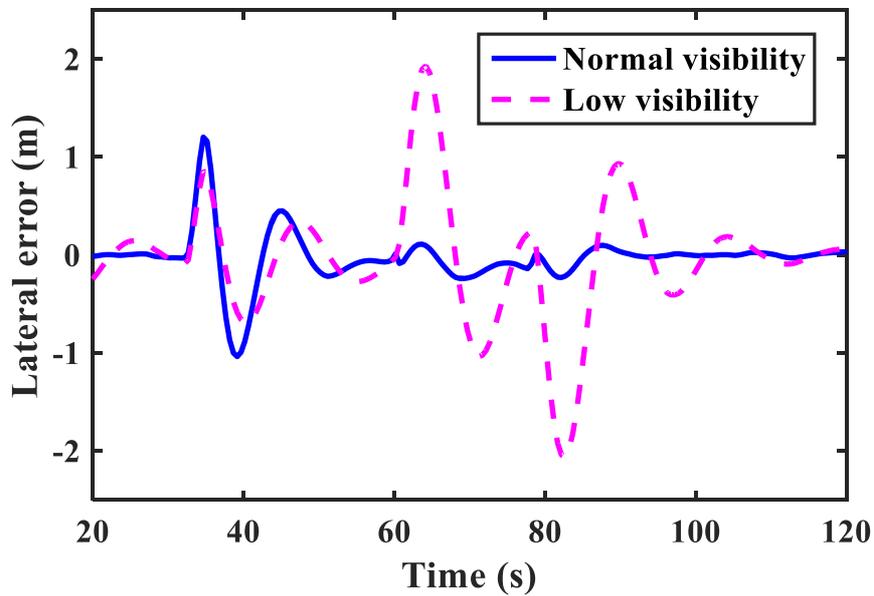

Figure 4.10 Comparison in lateral error between the conditions of normal visibility and low visibility.

Figure 4.10 shows the comparison in lateral error between the conditions of normal visibility and low visibility. It can be observed that, along the curve, the lateral error is larger and the response is more fluctuated in the condition of low visibility compared to normal visibility. The similar results were obtained from the measured lane following performance in experimental study II. According to the experimental results, driving condition of VF near yielded lower TLC and higher SWRR along curves compared to the condition of VF whole. In terms of the perturbation on the straight lane, both conditions of normal visibility and low visibility yielded a quick response and the vehicle maintains within the lane.

Figure 4.11 shows the comparison in steering wheel angle between the conditions of normal visual attention and declined visual attention. It can be observed that there is an obvious peak of steering wheel angle when approaching and leaving the curve in the condition of declined visual attention, and the response is more fluctuated compared to the condition of normal visual attention. In terms of the perturbation on the straight lane, the condition of declined visual attention also yielded more fluctuated response of steering wheel angle compared to normal visual attention.

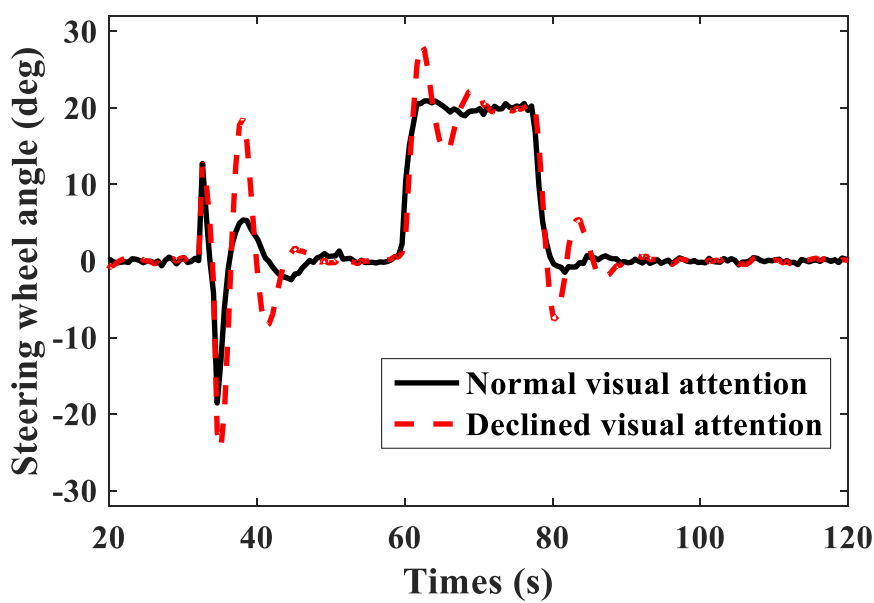

Figure 4.11 Comparison in steering wheel angle between the conditions of normal visual attention and declined visual attention.

Figure 4.12 shows the comparison in lateral error between the conditions of normal visual attention and declined visual attention. It can be observed that, along the curve, the lateral error is larger and response is more fluctuated in the condition of declined visual attention compared to normal visual attention. In terms of the perturbation on the straight lane, the condition of declined visual attention also yielded more fluctuated response and larger later error compared to the condition of normal visual attention. It is indicated that declined visual attention leads to a higher lane departure risk.

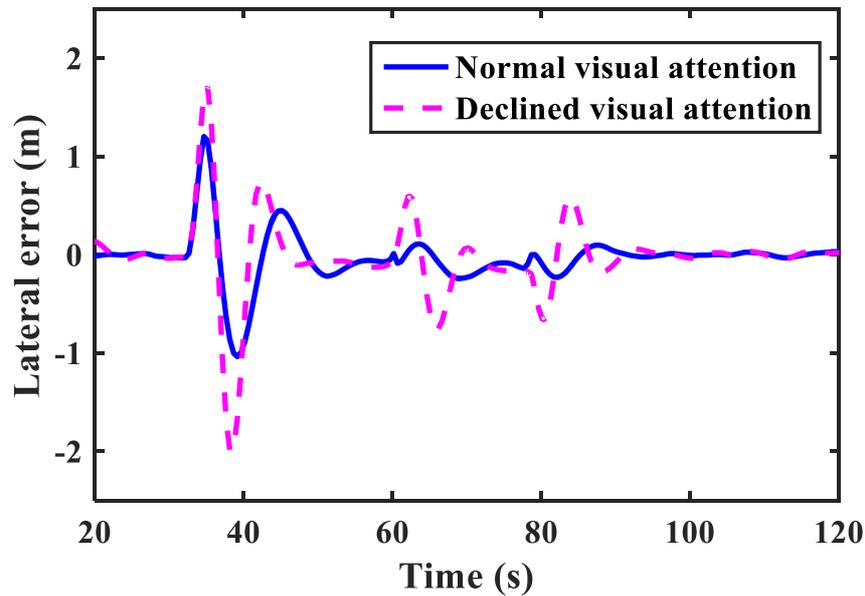

Figure 4.12 Comparison in lateral error between the conditions of normal visual attention and declined visual attention.

### 4.4.2 Assisted driving case

- Condition of normal visual information

Figure 4.13 shows the comparison in steering wheel angle between the manual driving and assisted driving with the haptic guidance system in the condition of normal visual information, and Figure 4.14 shows the comparison in lateral error. It can be observed that the effect of haptic guidance system on lane following performance is not evident along the curve, as the performance in manual driving is already satisfying. The similar observations on the lane following performance measured by TLC and SWRR were obtained in the experimental study I. In addition, the numerical result shows that the response to the perturbation on the straight lane is less fluctuated and lateral error is smaller in the condition of assisted driving than of manual driving. The reason would be that derivative action is important to make a quick response action; by relying on the haptic guidance system, the lack of derivative action by human driver was compensated by the proportional-derivative control of the haptic guidance system.

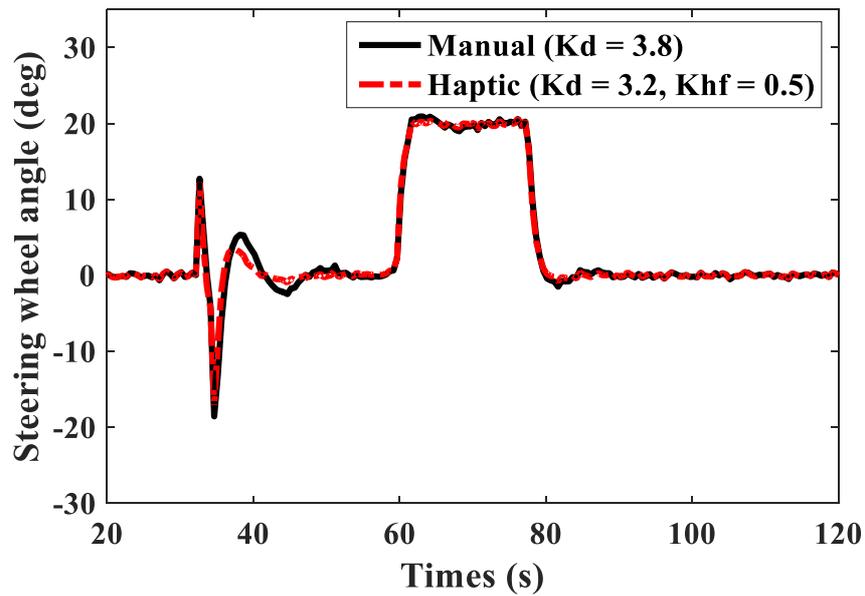

Figure 4.13 Comparison in steering wheel angle between the conditions of manual driving and assisted driving with the haptic guidance system.

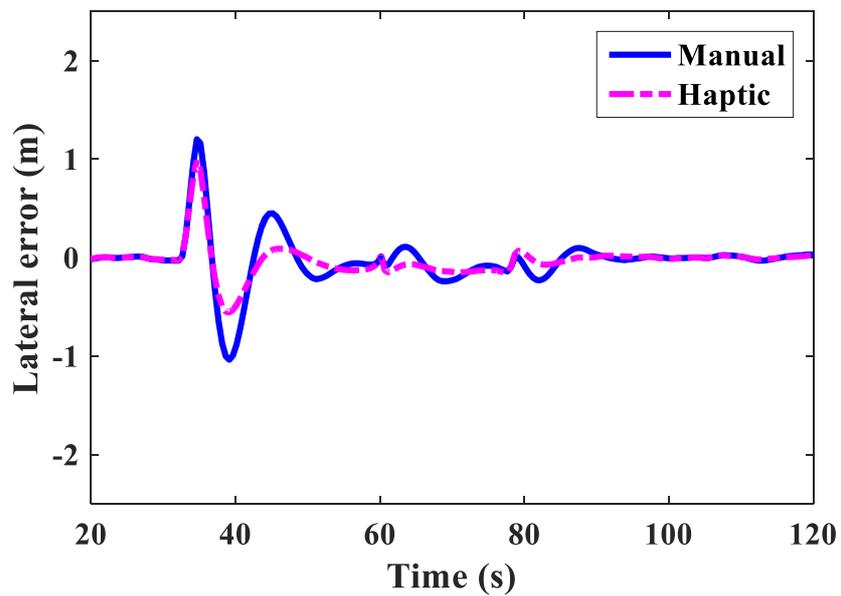

Figure 4.14 Comparison in lateral error between the conditions of manual driving and assisted driving with the haptic guidance system.

- Condition of low visibility

Figure 4.15 shows the comparison in steering wheel angle between the manual driving and assisted driving with the haptic guidance system in the condition of low visibility, and Figure 4.16 shows the comparison in lateral error. Low visibility induces fluctuated steering maneuver and larger later error. It can be observed that the effect of the haptic guidance system on improving lane following performance is evident, as the haptic guidance system leads to a smoother steering response and reduced lateral error. It suggests that the haptic guidance system is effective in compensating the decrement of lane following performance caused by low visibility.

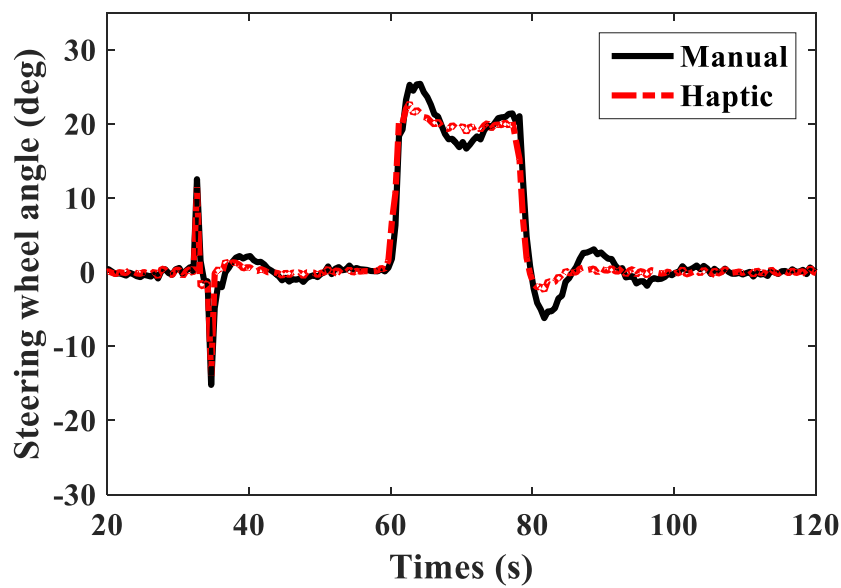

Figure 4.15 Comparison in steering wheel angle between the conditions of manual driving and assisted driving with the haptic guidance system.

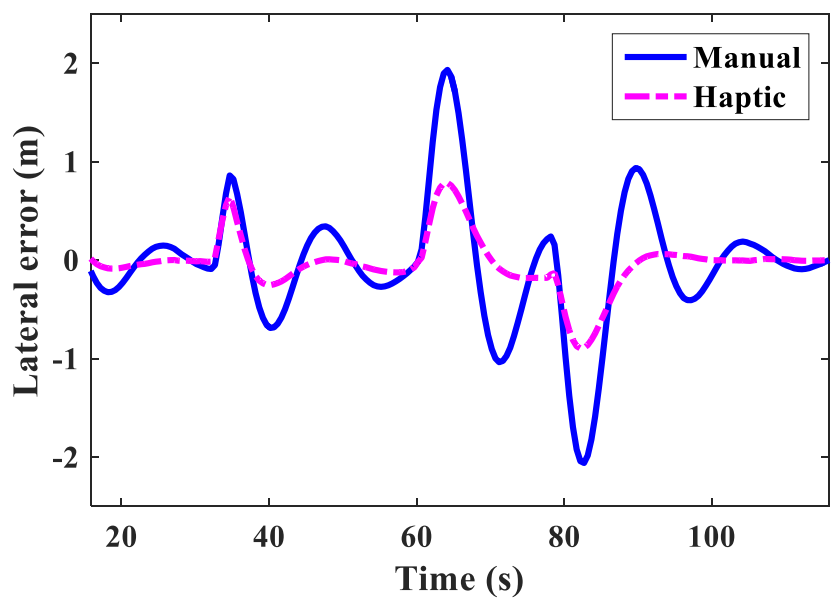

Figure 4.16 Comparison in lateral error between the conditions of manual driving and assisted driving with the haptic guidance system.

- Condition of declined visual attention

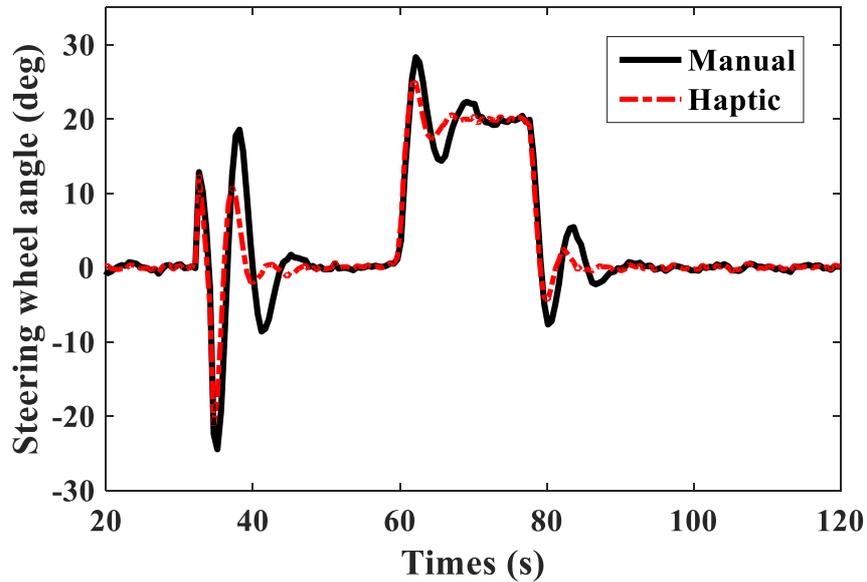

Figure 4.17 Comparison in steering wheel angle between the conditions of manual driving and assisted driving with the haptic guidance system.

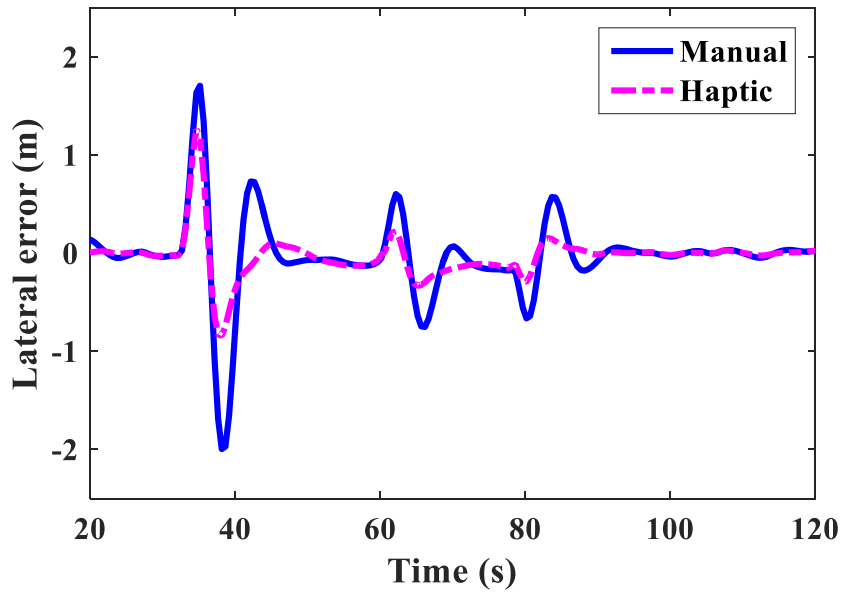

Figure 4.18 Comparison in lateral error between the conditions of manual driving and assisted driving with the haptic guidance system.

Figure 4.17 shows the comparison in steering wheel angle between the manual driving and assisted driving with the haptic guidance system in the condition of declined visual attention, and Figure 4.18 shows the comparison in lateral error. Declined visual attention induces fluctuated steering maneuver and larger later error along the driver course. It can be observed that the effect of haptic guidance system on improving lane following performance is evident, as the haptic guidance system leads to a smoother steering response and reduced lateral error. It suggests that the haptic guidance system is effective in compensating the decrement of lane following performance caused by declined visual attention.

## 4.5 Discussion

From the identification results of model fitness, we found that the proposed driver model matched driver input torque with a fitness of 76% on average among the participants in the condition of Manual, and a fitness of 69% on average in the condition of HG-normal, which is shown to be acceptable after considering the individual variability in steering behavior. The results of lane following performance in experimental study I indicate that the individual variability in steering behavior was quite large. Individual differences such as driving experience, age, or physical factors like body size could explain the individual variability in steering behavior. In order to improve the model fitness, future study could address more advanced nonlinear driver parameter estimation method, for example nonlinear Kalman filter, which has been used to understand the steering behavior of different types of drivers [56].

However, the fitness of driver input torque was 57% on average in the condition of HG-strong, which is lower than the conditions of Manual and HG-normal. According to the results of experimental study I, one explanation would be that the driver behavior became more complicated when driving with strong haptic guidance, especially when the participants felt frustration due to the conflicts between the system and the driver. In the proposed driver model, driver's interaction with the haptic guidance system was represented by the parameter $K_{hf}$ and it is assumed that $K_{hf}$ remains constant along the driving course, but the actual driver behavior could be more complicated. Thus, in order to improve the fitness of driver input torque, future study would focus on modeling driver's interaction with the haptic guidance system with some additional parameters, for example time response of haptic feedback.

One of the issues concerning the driver model is the driver feedback of visual information during a lane following task. There is still controversy over how visual information is used by drivers [121], especially in the case of driving with visual occlusions from road ahead [51]. In the numerical simulation, it was assumed that lateral error at the near point was used for lane following task in the condition of low visibility, with a PID controller and a processing time delay in order to yield stable control. The parameter of constant gain for derivative action $a_3$ is determined by a trial-and-error process. It was assumed that driver could increase $a_3$ to deal with low visibility, especially when negotiating a curve. Numerical simulations with fixed $a_1$ and $a_2$ and with low value of $a_3$ indicate a bad lane following performance (lane boundary crossing along the curve). The increase of $a_3$ yielded a better lane following performance, and $a_3$ equaling to 0.3 kept the vehicle within the lane along the curve. In future study, more values of $a_3$ (such as 0.4, 0.5 and so on) would be addressed in the numerical simulations. However, increase of derivative action could give much workload to the human driver. Accordingly, future study would also address the parameter identification of driver visual system in the condition of low visibility in order to find the range of $a_3$.

In terms of the condition of declined visual attention under fatigue, it is a limitation to simply represent declined visual attention with a higher value of processing time delay. It has been found that the processing time delay increases when the driver visual attention declined due to fatigue driving. However, the actual driver behavior will be more complicated. According to [56], a processing time delay for a normal driver would be between 0.01 to 0.3 s. A higher value of 0.5 s, was used in the simulation study to represent the driver with declined visual attention. Considering the individual difference of driver behavior, a set of values (0.3 s, 0.4 s, 0.5 s, …) will be addressed in the future study. Moreover, the driver visual attention was increased by the stimulus of haptic information provided by the guidance system in the experimental study III, and this will also be addressed in the numerical simulation in the future study.

In addition to the driver's own state, the driver behavior will also be influenced by the environmental factors in real driving situations. For example, other vehicles occur at various intervals, and the driver would have to deal with the intent of other drivers. Traffic signs on the road could attract the driver's attention. Thus, the current driver model is a simplified one that basically deals with a lane following task in a monotonous driving environment. When it comes to a real-life driving task, the driver model should address more environmental factors, which will be a future study.

Due to the above-mentioned limitations, it is early to draw a broad conclusion because more validation process of the proposed model is still needed. However, within the content of this chapter, it is shown that the proposed driver model has the potential to predict driver behavior and to become a useful tool to evaluate the effectiveness of the haptic guidance system.

## 4.6 Conclusion

The objective evaluation of the effect of haptic guidance on driver behavior requires the development of a driver model with consideration of both visual and haptic information. The goal of this chapter is to understand the driver behavior obtained in the experiments, and furthermore, to predict driver behavior for investigating the effect of haptic guidance on lane following performance. To achieve the goal, this chapter has detailed the identification and validation of the proposed driver model, and numerical analysis of driver behavior influenced by degraded visual information and the haptic guidance system.

The results of model identification and validation indicate that the proposed driver model has the potential to predict driver behavior when driving with the haptic guidance system. In addition, the identified parameter of steering angle to torque gain, $K_d$, was significantly higher in the condition of Manual compared to HG-normal and HG-strong. It indicates that the participants reduced the steering angle to torque gain when interacting with the haptic guidance system. The results of numerical simulations indicate that the lane following performance was decreased in the conditions of degraded visual information, including low visibility and declined visual attention under fatigue, and the haptic guidance system was effective in improving the lane following performance.

Finally, the observations in this chapter suggest the potential of using the proposed model to predict driver behavior, and to conduct numerical simulations for further designing and evaluating the haptic guidance system in future work.

# Chapter 5

# General Discussions

# 5 General Discussions

This thesis focuses on analysis and modeling of driver behavior based on integrated feedback of visual and haptic information for shared steering control. This section generally discusses the results obtained in the driving simulator experiments and numerical analysis. Firstly, the effect of degraded visual information on driver behavior will be discussed. Then, the discussion on the effect of haptic guidance on driver behavior in cases of normal and degraded visual information will be presented.

**5.1 Effect of degraded visual information on driver behavior**

Drivers always suffer varying degrees of performance decrements in cases of degraded visual information during a lane following task. For example, in real-life driving, driver behavior would be influenced when visual feedback is defective due to environmental obstacles, such as fog, rain, or during night driving [57, 58]. In addition, declined visual attention leads to decreased speed of neural transmission or longer visual processing time [102], which would also influence driver behavior. This thesis firstly attempted to reflect these two kinds of degraded visual information by introducing a two-point driver visual model. To investigate the effect of degraded visual information, including visual occlusions from road ahead and declined visual attention, on driver behavior, experimental studies were conducted. Moreover, numerical simulations based on the two-point visual model were performed to test the model for predicting driver behavior with degraded visual information.

Experimental study II investigated the driver behavior when driving with degraded visual information caused by the visual occlusion from road ahead. In the experiment, when a particular segment of the road was visible, the rest of the road remained black. There were four conditions of visual feedback (VF): VF whole, VF near, VF mid, and VF far. The results indicate that the driver steering behavior was significantly influenced by visual feedback. For VF near, the starting moment of turning maneuver occurred later than the geometrical turning point due to the lack of advance road information. In addition, SWRR was higher than the other three visual feedback conditions, which suggests that the steering behavior for VF near was unstable. According to the results of SDLP and TLC shown in Figure 3.27 and Figure 3.28 in Section 3.2.4, the lane following performance for VF mid was similar to VF whole, where both near and far segments of the road were partially provided [51]. Compared to VF whole and VF mid, the lane departure risk was higher for VF near and VF far. This can be explained by the fact that visual feedback from the near segment helps the drivers attain their vehicle position, and the far segment leads to the prediction of upcoming roads [54].

Experimental study III investigated the driver behavior when driving with declined visual attention under fatigue. According to the result of PERCLOS shown in Figure 3.39, there was a tendency that the driver visual attention decreased over time. According to the results of SDLP and MALE shown in Figures 3.41 and 3.43, there was a tendency that the lane following performance decreased when the driver visual attention decreased over time.

Therefore, the experimental results have shown that the driver behavior was significantly influenced by the visual occlusion from the road ahead and declined visual attention under fatigue driving. The objective evaluation of the effect of haptic guidance on driver behavior requires the development of a driver model with consideration of both visual and haptic information. Considering this, numerical analysis based on modeling is useful to understand the driver behavior obtained in the experiments. Moreover, to test the effect of haptic guidance on lane following performance in the conditions of degraded visual information, the development of a suitable driver model that can predict driver behavior is also required. In the current state, to gain confidence that the developed model can address the effect of haptic guidance system on driver behavior, two different simulation scenarios of degraded visual information were tested.

One scenario of degraded visual information, namely low visibility, corresponded to the condition of VF near in the experimental study II. It was assumed that the driver used visual information from only a near point of road ahead to perform a lane following task in the condition of low visibility. As a comparison, both a near point and a far point from road ahead were used in the condition of normal visual information (or normal visibility) in the numerical simulation. The simulation results show that there was an obvious peak of steering wheel angle when approaching and leaving the curve in the condition of low visibility, and the response was more fluctuated compared to the condition of normal visibility. The similar results were obtained from the measured steering wheel angle in the experimental study II. In the experiment, the driving condition of VF near yielded more fluctuated response of steering wheel angle compared to the condition of VF whole. Moreover, the simulation results show that, along the curve, the lateral error was larger and response was more fluctuated in the condition of low visibility compared to normal visibility. Similarly, in the experiment, driving condition of VF near yielded lower TLC and higher SWRR along curves compared to the condition of VF whole.

Another scenario of degraded visual information corresponded to the condition of declined visual attention under fatigue driving in the experimental study III. It was assumed that the processing time delay of driver visual model was longer when performing a lane following task in the condition of declined visual attention under fatigue. As a comparison, a shorter processing time delay was applied in the condition of normal visual attention in the numerical simulation. The simulation results show that on the straight lane, the condition of declined visual attention yielded a more fluctuated response of steering wheel angle compared to the condition of normal visual attention, and the lateral error became larger. Similarly, in the experiment, the lane following performance, measured by SDLP and MALE, decreased when the driver visual attention decreased.

It should be noticed that a limitation of the current numerical simulations is that the values of parameters in the simulation study were mainly obtained from references and based on a trial-and-error process, although the similar tendency of driver behavior was observed between the numerical simulation and the experimental results. Therefore, although it is early to draw a broad conclusion because more validation process of the proposed model is still needed, the proposed driver model has the potential to predict driver behavior, and future study will be carried on to further validate the driver model.

## 5.2 Effect of haptic guidance on driver behavior in cases of degraded visual information

In the current state, there is not much agreement on how to design and how to evaluate a haptic guidance system on an as-needed basis in order to improve both driving safety and comfort. To solve this problem, we proposed that the haptic guidance system should be designed to provide supplementary haptic information based on the reliability of visual information perceived by the driver. Thus, this thesis focused on analysis of driver behavior influenced by the haptic guidance system in cases of normal and degraded visual information.

To effectively interpret and interact with the world, humans are capable of combining multiple sensory information. In this thesis, it is hypothesized that the driver could integrate the feedback of visual and haptic information when driving with the haptic guidance system. Thereby, the decrement of lane following performance caused by degraded visual information could be compensated by the haptic information provided by the guidance system, and driver workload could be reduced by the complemented haptic information in the case of normal visual information.

In the experimental study I, the effect of different degrees of haptic guidance on driver behavior was investigated. The results indicate that lane following performance increased or remained similar when driving with the proposed haptic guidance system, and meanwhile driver steering effort was reduced. However, the great reduction in driver steering effort caused by strong haptic guidance had a downside effect, which was the feeling of frustration measured by the NASA-TLX. Moreover, the results of measured driver visual behavior show that the implementation of haptic guidance system did not significantly influence the driver visual attention on the look-ahead point when driving along curves. It indicates that the implementation of haptic guidance system did not induce additional demand of visual attention on the look-ahead point along curves. On the other hand, it also indicates that the driver did not reduce the visual attention on the look-ahead point when the complemented haptic information was provided.

It should be noticed that the strong haptic guidance induced frustration feelings to the drivers, as sometimes the haptic guidance system and the driver intent could not match. However, the strong haptic guidance torque cannot be completely replaced by weaker torque, as it is still necessary in the cases of emergency. To address this issue, nonlinear models, like a quadratic function based model, would have the potential to reduce the frustration feelings or conflict between the driver and the system, while improving the driving safety. Specifically, if the haptic guidance torque is designed as a quadratic function of the lateral error, the haptic guidance torque would be little when the lateral error is small, but would increase quite rapidly when the lateral error becomes larger. Considering this, although the linear model has the advantage of being more intuitive to the drivers, non-linear models would be an interesting future study.

In the experimental study II, the effect of haptic guidance on driver behavior when driving with degraded visual information caused by the visual occlusion from road ahead was investigated. The results show that the decrement of lane following performance caused by the visual occlusion from the far segment of road ahead (or VF near) could be compensated by haptic information, and strong haptic guidance was more effective than normal haptic guidance. In addition, the decrement of lane following performance on straight lanes caused by the visual occlusion from the near segment of road ahead (or VF far) could be compensated by haptic information, but on curves the effectiveness was not significant. Subjective evaluation results show that most participants preferred strong haptic guidance in the condition of VF near. However, normal haptic guidance was the more preferred choice in the other three conditions, where visual information was more reliable compared to the condition of VF near.

In the experimental study III, the effect of haptic guidance on driver behavior when driving with degraded visual information caused by declined visual attention under fatigue was investigated. The results show that the decrement of lane following performance on straight lanes caused by declined visual attention was compensated by haptic information provided by the guidance system. On the other hand, the active torque on the steering wheel stimulated the fatigued driver, and according to the result of PERCLOS, the driver visual attention was increased.

The experimental results have shown that the haptic guidance system had significant effect on the driver behavior, and the decrement of lane following performance caused by degraded visual information was compensated by haptic information provided by the guidance system. The results indicate that the proposed haptic guidance system was capable of providing reliable haptic information, and the driver tended to integrate visual and haptic information to achieve better lane following performance. However, it is unknown how the driver integrated visual and haptic information. In addition, the objective evaluation of the effect of haptic guidance on driver behavior required the development of a suitable driver model with consideration of both visual and haptic information. Therefore, measured data from the experimental study I were used to identify the driver model parameters. After that, model validation test was performed in the SIMULINK model. The identification and validation results indicate that the proposed driver model was shown to be appropriate for identification of driver torque when driving with the haptic guidance system. In addition, the validated model was used to predict driver steering behavior when driving with the haptic guidance system.

The effect of the haptic guidance system on lane following performance in the conditions of normal and degraded visual information was investigated by numerical simulations. In the condition of normal visual information, the effect of haptic guidance system on lane following performance was not evident, as the performance in manual driving was already satisfying. In the condition of low visibility, the effect of haptic guidance system on improving lane following performance was evident, as the haptic guidance system led to a smoother steering response and reduced lateral error on the straight lane and the curve. In the condition of declined visual attention, the effect of haptic guidance system on improving lane following performance was evident, as the haptic guidance system resulted in a smoother steering response and reduced lateral error on the straight lane and the curve. The results indicate that the haptic guidance system was effective in improving the lane following performance in the conditions of low visibility and declined visual attention under fatigue. Still, the actual driver behavior would be much more complicated than the model presented. Thus, the proposed driver model has the potential to predict driver behavior, although it is early to conclude that the driver model is in parallel with the driver behavior observed in the experiment, and future study will be carried on to further validate the driver model when driving with the haptic guidance system.

In real-life driving, the haptic guidance system would not always perform well, as the system could fail to detect the driving path or obstacles due to environmental factors, sensor failures, and so on. In those situations, the haptic guidance system would have to stop providing assistant torque, or even worse, the haptic guidance torque could become a strong disturbance to the driver's intent. Thus, the influence of automation failure should be comprehensively considered when designing the assistance system. We have considered the failure of the haptic guidance system in the case of avoiding a forward collision in the experimental study I. The result indicates that higher degree of haptic guidance led to a higher risk of forward collision when the system failed and the driver had to make a maneuver to overrule the system. In order to eliminate the effect of automation failure on reducing the driving performance, warning signals, like vibrations on the steering wheel, should be provided to the driver. However, sometimes the system failure cannot be successfully detected, so an alternative solution would be setting safety parameters in the design of the system. In this study, the maximum haptic guidance torque was limited to 5 N·m by the program, so that the driver can overrule the haptic guidance system at any time by generating more torque on the steering wheel. Still, other safety parameters, for example, the parameters that address a smooth transition from assisted driving to manual driving in the case of a critical event, need to be considered when designing the haptic guidance system in the future study.

# Chapter 6

# Conclusions

# 6 Conclusions

## 6.1 General conclusions

The current conclusions in this chapter reflect the aim of this thesis, which is to perform an analysis of the effect of haptic information provided by the guidance system on driver behavior in cases of normal and degraded visual information. Through analysis and modeling of driver behavior, we aim to provide theoretical and experimental knowledge of driver behavior when driving with a haptic guidance system in lane following tasks. Reflecting our aim on the accomplished work, we may conclude that it contributed to the design of a haptic guidance system, which is capable of providing reliable haptic information and compensating the performance decrement caused by degraded visual information through effective evaluations, and also contributed to a parameterized driver model that predicts the driver behavior influenced by both visual and haptic information.

This thesis begins with detailed reviews about various advanced driver assistance systems to improve driving safety and to reduce the driver workload, and pays specific attention to driver-automation shared control system. It has been shown that the driver-automation shared control through haptic interface always remains the driver in the control loop, and the driver could receive continuous feedback about the automation functionality; moreover, the driver benefits from the increased performance and reduced workload. However, the haptic guidance system has its drawback. When facing a critical event, confusion may occur to the driver on trusting the system or not. What's more, the haptic guidance system may induce conflicts and increased workload to the driver if the system is misused and the guidance torques cannot be relied on. Thus, it is reasonable to believe that supplementary haptic information should be provided by the haptic guidance system on an as-needed basis. To address this issue, in this thesis, we proposed that the need of haptic information provided by the haptic guidance system could be based on the reliability of visual information perceived by a driver, and the system could be evaluated by its performance compensation on visual perception. Therefore, this thesis focuses on the design and evaluation of a haptic guidance system by analysis and modeling of driver behavior with integrated feedback of visual and haptic information.

In Chapter 2, the general structure of the driver-vehicle-road system was first presented, and it was followed by the description of each subsystem: driver visual system, driver neuromuscular system, haptic guidance system, steering column system, and vehicle dynamics system. The model of driver visual system addresses different cases of degraded visual information, including visual occlusion from the far segment of road ahead and declined visual attention under fatigue driving. An original model of driver neuromuscular system which integrates visual information from road ahead and haptic information from the guidance system, was proposed. Based on the model, the driver could integrate visual and haptic information by tuning the target steering wheel angle to torque gain and neuromuscular reaction gain for haptic feedback. The haptic information was provided by a haptic guidance controller designed based on a two-point visual model, and proportional-derivative control theory was applied to compensate the limitation of driver's derivative control action and to keep the vehicle accurately following the target trajectory.

In Chapter 3, three driving simulator experimental studies were presented. Experimental study I focused on testing driver behavior under different degrees of haptic guidance when driving with normal visual information. The 15 participants drove five trials: no haptic guidance (Manual), haptic guidance with a normal feedback gain (HG-normal), haptic guidance with a strong feedback gain (HG-strong), and two kinds of automated driving. The driver behaviors, including lane following performance, gaze behavior and subjective evaluation, were analyzed. To evaluate the lane following performance, standard deviation of lane position, and time-to-lane crossing were calculated. In terms of driver visual behavior, the heat maps of gaze distribution were plotted to indicate the high and low frequent gaze direction. To statistically study the distribution of gaze direction along curves, the mean value of percent road center was calculated. Subjective evaluation on NASA-TLX was used to rate the perceived workload of the driving task. Finally, the method of statistical analysis, including repeated measures analysis of variance and paired wise comparisons, were presented. From the results, we found that the lane following performance increased or remained similar when driving with the proposed haptic guidance system, and meanwhile driver steering effort was reduced. However, the great reduction in driver steering effort caused by strong haptic guidance had a downside effect, which was the feeling of frustration. In addition, the implementation of the haptic guidance system did not significantly influence the driver visual attention on the look-ahead point along curves, which indicates that the haptic guidance system was reliable and did not induce the increase of driver visual demand.

Experimental study II focused on testing driver behavior influenced by the haptic guidance system when driving with degraded visual information caused by visual occlusion from road ahead. The 12 participants drove twelve trials designed by combining three degrees of haptic guidance (HG): HG none (or manual driving), HG normal, and HG strong, with four scenarios of visual feedback (VF): VF whole, VF near, VF mid, and VF far. From the results, we found that the decrement of lane following performance caused by visual occlusion from road ahead could be compensated by the haptic information provided by the guidance system, and strong haptic guidance was more effective than normal haptic guidance. Subjective evaluation results show that most participants preferred strong haptic guidance in the condition of VF near. However, normal haptic guidance was the more preferred choice in the other three conditions, where visual feedback was more reliable compared to the condition of VF near.

Experimental study III focused on testing driver behavior influenced by the haptic guidance system when driving with declined visual attention under fatigue. The 12 participants drove two trials, a treatment session that implemented the haptic guidance steering system after a prolonged driving on monotonous roads and a control session without the system. From the results, we found that the decrement of lane following performance on straight lanes caused by declined visual attention under fatigue could be compensated by haptic information provided by the guidance system. On the other hand, the active torque on the steering wheel stimulated the fatigued driver, and according to the result of PERCLOS, the driver visual attention was increased.

In Chapter 4, firstly, the driver model parameters were identified based on the measured data from the experimental study I. Secondly, model validation test was performed. After that, numerical analyses of driver behavior with normal and degraded visual information were conducted. The driver model addressed the integrated feedback of visual and haptic information, and the effect of the haptic guidance system on the lane following performance was investigated. The numerical simulation results indicate that the lane following performance was decreased in the conditions of low visibility and declined visual attention under fatigue, and the haptic guidance system was effective in improving the lane following performance. It suggests that the proposed driver model has the potential to predict driver behavior and to become a useful tool to design and evaluate the haptic guidance system.

In Chapter 5, general discussions about the observed results from three experimental studies and the numerical analysis were provided. The discussions were divided into two parts. The first part addressed the effect of degraded visual information, including visual occlusion from road ahead and declined visual attention under fatigue driving, on driver behavior. The second part addressed the effect of the haptic guidance system on driver behavior in cases of degraded visual information.

To sum up, the thesis presents contributions made to the evaluation and design of a haptic guidance system on improving driving performance in cases of normal and degraded visual information, which are based on behavior experiments, modeling and numerical simulations. The main conclusions are drawn as follow:

(a) The effect of shared control on driver behavior in cases of normal and degraded visual information has been successfully evaluated experimentally and numerically. The evaluation results indicate that the proposed haptic guidance system is capable of providing reliable haptic information, and is effective on improving lane following performance in the conditions of visual occlusion from road ahead and declined visual attention under fatigue driving.

(b) The appropriate degree of haptic guidance is highly related to the reliability of visual information perceived by the driver, which suggests that designing the haptic guidance system based on the reliability of visual information would allow for greater driver acceptance.

(c) The parameterized driver model, which considers the integrated feedback of visual and haptic information, is capable of predicting driver behavior under shared control, and has the potential of being used for designing and evaluating the haptic guidance system.

## 6.2 Limitations and recommendations

The limitations of experimental studies and numerical analysis have been indicated in the discussion part in each chapter. In this section, the general limitations and recommendations of the current study are presented.

The design and evaluation of the haptic guidance system in the current study were based on driving simulator experiments. Future study will address real-world driving tests, in which the haptic guidance system would not always be reliable, as it could fail to detect the driving path due to environmental factors or sensor failures. Safety parameters, for example, the parameters that would allow for a smooth transition from assisted driving to manual driving, need to be considered when designing the system. It was argued that, by carefully designing the driving simulator experiment, the above situation caused by sensor failures could also be to some extent simulated. However, the driver's reaction to the failures would be quite different in real-world driving tests, as the driver has to face the risk of injury which could be caused by a lane departure [122]. Moreover, drivers' subjective feedback on the appropriate degree of haptic guidance, which might be based on a cost-benefit analysis of safety and comfort, would also be different, because the safety concerns would become more crucial in real-world driving tests.

Another limitation is that the current driver model consists of a visual system and a neuromuscular system, which are linear and time-invariant. In the current experiment, the driver steering behavior could be approximately constant along the driving trial with a lane following task as the driving simulator experiment was carefully designed. The aspect of driver's cognition, which is crucial for higher-level of driver behavior, such as decision making and planning [123], is helpful to improve the driver model but was not considered in the current model. In future study, advanced non-linear modeling by taking account of driver's cognition, for example the fuzzy modeling, would be explored to address such higher-level of driver behavior.

## 6.3 Outlook

Regarding further applications based on the current findings, the proposed driver model is expected to be used to more effectively design the haptic guidance system through numerical simulations before implementing the system in a driving simulator or a real vehicle. Moreover, the parameterized driver model can also be used to predict driver behavior for designing of individualized haptic guidance systems. It is expected that a well-designed haptic guidance system based on the reliability of visual information would allow for greater driver acceptance not only on lane following assistance, but also on lane change support. Furthermore, in order to optimize the haptic guidance system, further study on evaluation criterions could address the cost-benefit analysis of driving safety and comfort by taking account of the evaluation indexes in this thesis.

Further application fields would include lateral steering assistance systems and haptic teleoperation systems. If the research challenges of system modeling, design, and evaluation are well fixed, it is expected that the haptic guidance system would be a promising way to assist an operator especially when the perceived visual information is degraded.